\documentclass[10pt,twocolumn,letterpaper]{article}

 \usepackage{cvpr}              %
\usepackage{graphicx}
\usepackage{caption}
\usepackage{subcaption}
\usepackage{array}
\usepackage{float}
\usepackage{microtype}
\usepackage{comment}
\usepackage{longtable}
\usepackage{chapterbib}
\usepackage[accsupp]{axessibility}

\definecolor{cvprblue}{rgb}{0.21,0.49,0.74}
\usepackage[pagebackref,breaklinks,colorlinks,allcolors=cvprblue]{hyperref}

\def\paperID{xxxxx} %
\def\confName{CVPR}
\def\confYear{2026}

\title{When Pretty Isn’t Useful: Investigating Why Modern Text-to-Image Models Fail as Reliable Training Data Generators}

\author{
Krzysztof Adamkiewicz$^{1,2\star}$ \and
Brian B. Moser$^{2\star}$ \and
Stanislav Frolov$^{2\star}$ \and \
Tobias Christian Nauen$^{1,2}$ \and
Federico Raue$^{2}$ \and
Andreas Dengel$^{1,2}$  \and \\
$^{1}$RPTU University Kaiserslautern-Landau, Kaiserslautern, Germany \\
$^{2}$German Research Center for Artificial Intelligence (DFKI), Kaiserslautern, Germany \\
$^{\star}$Equal contribution \\
{\tt\small first.last@dfki.de \quad first\_second.last@dfki.de}
}

\begin{document}
\maketitle

\begin{abstract}
Recent text-to-image (T2I) diffusion models produce visually stunning images and demonstrate excellent prompt following. But do they perform well as synthetic vision data generators?
In this work, we revisit the promise of synthetic data as a scalable substitute for real training sets and uncover a surprising performance regression.
We generate large-scale synthetic datasets using state-of-the-art T2I models released between 2022 and 2025, train standard classifiers solely on this synthetic data, and evaluate them on real test data.
Despite observable advances in visual fidelity and prompt adherence, classification accuracy on real test data consistently declines with newer T2I models as training data generators.
Our analysis reveals a hidden trend: These models collapse to a narrow, aesthetic-centric distribution that undermines diversity and real data distribution coverage. 
Overall, our findings challenge a growing assumption in vision research, namely that progress in generative realism implies progress in data realism. We thus highlight an urgent need to rethink the capabilities of modern T2I models as reliable training data generators.
\end{abstract}

\section{Introduction}
Quality, quantity, and distribution of training data are fundamental determinants of performance and generalization in modern deep learning~\cite{kaplan2020scaling,ramanujan2023connection,hestness2017deep,udandarao2024no,chen2023meditron}. 
Yet as model architectures grow in scale and capacity, their appetite for vast, diverse, and well-curated data has rapidly outpaced the availability of real-world datasets~\cite{oquab2023dinov2,ravi2024sam2,grattafiori2024llama}. 
Collecting, annotating, and balancing such data has become an increasingly prohibitive bottleneck, particularly in domains constrained by privacy, limited sample availability, or strong domain shifts~\cite{gonzales2023synthetic,barisic2022sim2air,chen2023deep,koetzier2024generating,xu2023innovative}. 
As a result, data scarcity, not model capacity, is often the limiting factor in scaling.

Recent progress in text-to-image (T2I) diffusion has been seen as a promising way to overcome this bottleneck. 
Modern systems such as Stable Diffusion~\cite{rombach2022sd15,podell2023sdxl,esser2024scaling}, Flux~\cite{blackforestlabs2024flux1dev}, and Qwen-Image~\cite{wu2025qwen} can now generate detailed and diverse images from text alone. 
This capability has fueled interest in using synthetic data as a scalable, privacy-preserving, and domain-flexible alternative to real datasets. 
However, despite these advances in visual quality, it remains unclear whether such progress also improves the \emph{usefulness} of synthetic images for training downstream models.

\begin{figure}[t]
    \centering
    \includegraphics[width=\linewidth]{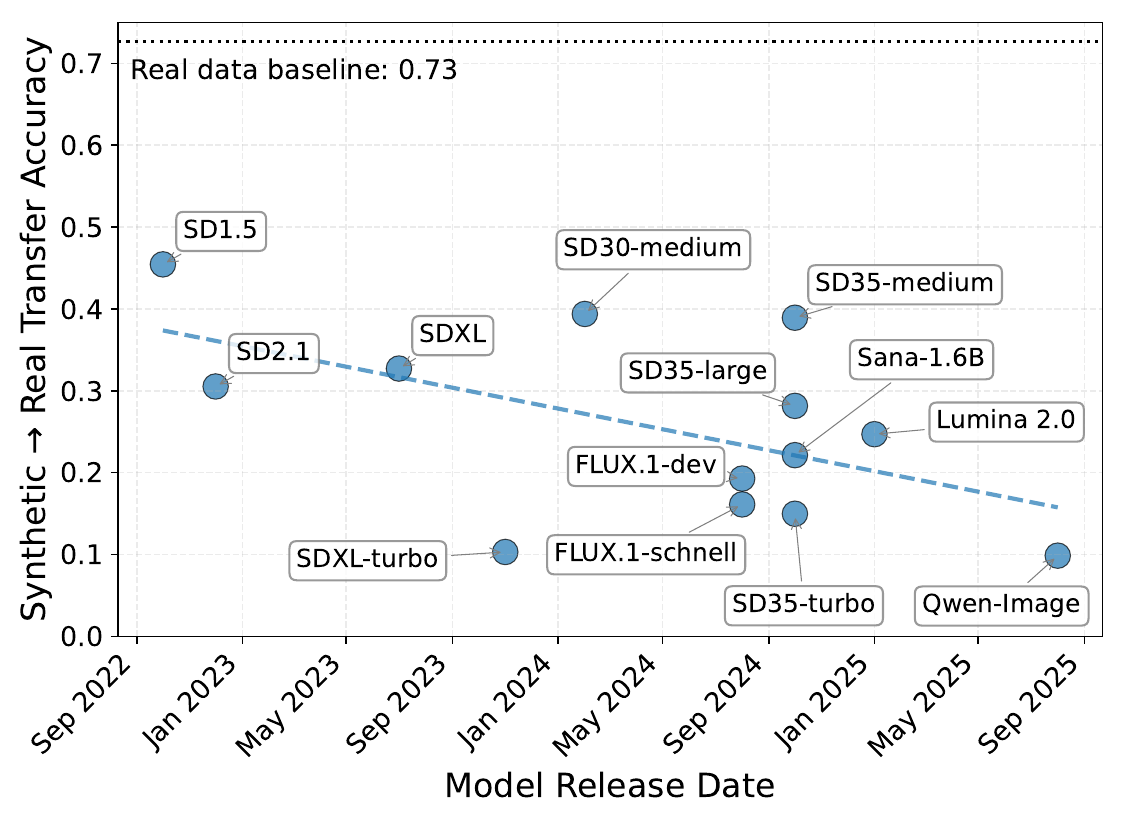}
    \caption{We train ResNet-50 classifiers on images generated by various T2I models for a subset of ImageNet-1k classes and evaluate their accuracy on real test data (Synth $\rightarrow$ Real). Our results reveal a downward trend over time. Newer models get progressively worse as reliable training data generators.
    }
    \label{fig:main_performance_drop}
\end{figure}

A growing body of work explores this question across a wide range of vision tasks. 
For classification, several studies have shown that models trained entirely on synthetic data can approach or even match real-data performance~\cite{tian2024learning,singh2024synthetic,sariyildiz2023fake,zhou2023training,koetzier2024generating}. 
Similar findings extend to object detection~\cite{fang2024data,zhu2024odgen,rothmeier2024time,voetman2023big}, contrastive learning~\cite{hammoud2024synthclip,tian2023stablerep}, continual learning~\cite{seo2024just,kim2024sddgr,masip2023continual}, and human pose estimation~\cite{saleh2025david,yang2024robust,wen2025highly}. 
These works demonstrate that, under controlled conditions, synthetic data can rival real data while avoiding the costs and constraints of manual curation.

It seems intuitive that as T2I models improve in fidelity, resolution, and photorealism, the synthetic data they produce should become correspondingly more effective for training. 
We test this intuition through a large-scale benchmark of thirteen state-of-the-art T2I models released between 2022 and 2025. 
By evaluating how well classifiers trained solely on synthetic data transfer to real test data in an image recognition task, we examine whether improvements in perceptual quality lead to improved capability as reliable training data generators.
Despite rapid progress in visual quality~\cite{podell2023sdxl,ramesh2022hierarchical} and photorealism~\cite{saharia2022photorealistic}, most of these works~\cite{gonzales2023synthetic,voetman2023big,hammoud2024synthclip,tian2023stablerep,rothmeier2024time} still rely on earlier diffusion models. 
As a result, the implications of newer advances (\ie, larger networks, internet-scale datasets, higher-resolution latent spaces, and stronger prompt conditioning) remain poorly understood in the context of synthetic data quality. 
This leads to our central and deceptively simple research question:  
\emph{Does progress in T2I generation actually translate into better synthetic training data for vision models?}

Our findings surprisingly challenge this expectation of performance improvement.
In \autoref{fig:main_performance_drop}, we show that classification accuracy on real test data consistently declines with newer T2I models as training data generators despite observable advances in visual fidelity and prompts adherence.
In this work, we investigate the reasons for this decline by probing texture, structure and spectral frequency mismatches to real data.
We find that newer models collapse to narrow, aesthetic-centric distributions that undermine diversity and label alignment.
And while using detailed text prompts, such as in \cite{lei2023image,fan2024scaling} can improve the downstream performance on real data, this gain comes at a cost: the synthetic manifolds become increasingly compact (\ie, high in density but low in coverage) signaling distributional drift and reduced intra-class diversity.

Our analysis separates sample realism from distribution realism and investigates several independent aspects of synthetic images showing that modern T2I models, though producing visually impressive images that look better, actually function worse as reliable training data generators. 
In summary, we make the following key contributions:
\begin{itemize}
    \item A \textbf{large-scale, time-spanning benchmark} of T2I models as training data generators. We evaluate thirteen open-source T2I models (2022-2025) on a controlled ImageNet-1K setup and show that, despite better visual fidelity and prompt following, Synth$\rightarrow$Real accuracy decreases for newer models.
    \item Probing \textbf{texture and frequency mismatches}.
    Using depth-based (structure) classifiers, local BagNet (texture) classifiers, and low/high-pass filtered inputs, we find that global structure and low frequencies are well retained, while texture quality and high-frequency detail are systematically degraded with newer models.
    \item Measuring \textbf{drift and diversity loss} in feature space. By measuring density \& coverage as well as comparing Real$\rightarrow$Synth vs.\ Synth$\rightarrow$Real transfer, we show that newer T2I models exhibit degraded distributional realism.
\end{itemize}

\begin{figure*}[h]
    \centering
    \setlength{\tabcolsep}{5pt}
    \begin{tabular}{>{\centering\arraybackslash}m{2cm} 
                    >{\centering\arraybackslash}m{2.6cm} 
                    >{\centering\arraybackslash}m{2.6cm} 
                    >{\centering\arraybackslash}m{2.6cm} 
                    >{\centering\arraybackslash}m{2.6cm} 
                    >{\centering\arraybackslash}m{2.6cm}}
        & \textbf{RGB} & \textbf{Depth (Structure)} & \textbf{RGB Patches (Texture)} & \textbf{Low Frequencies} & \textbf{High Frequencies} \\

        \textbf{Imagenet1k Images} &
        \includegraphics[width=\linewidth]{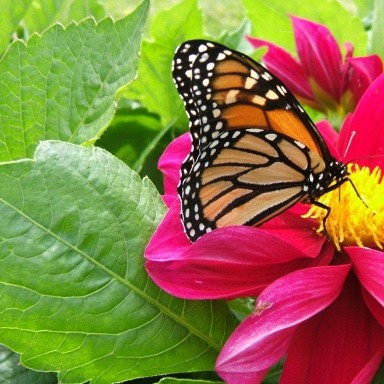} &
        \includegraphics[width=\linewidth]{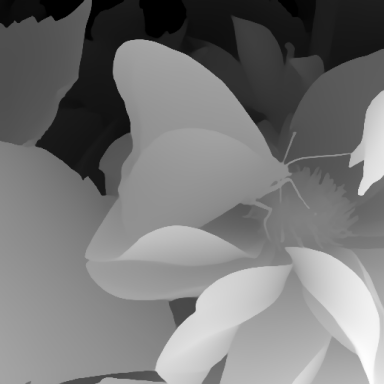} &
        \includegraphics[width=\linewidth]{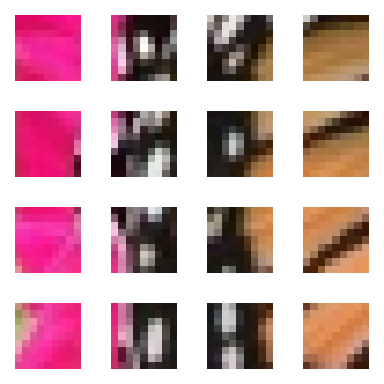} &
        \includegraphics[width=\linewidth]{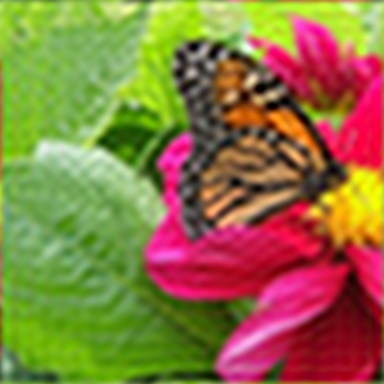} &
        \includegraphics[width=\linewidth]{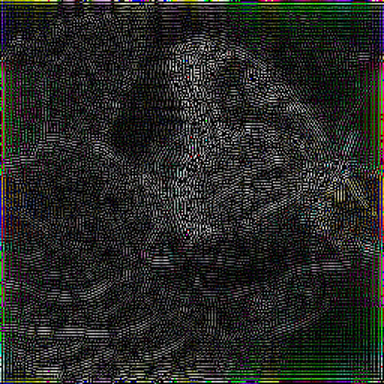} \\

        \textbf{Generated Images} &
        \includegraphics[width=\linewidth]{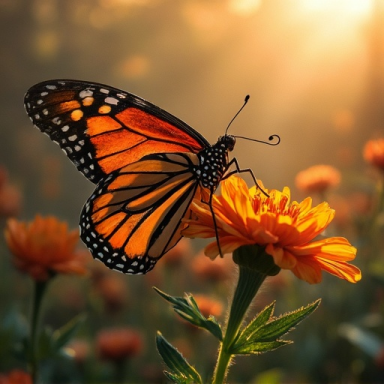} &
        \includegraphics[width=\linewidth]{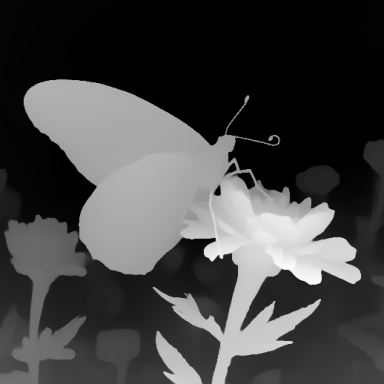} &
        \includegraphics[width=\linewidth]{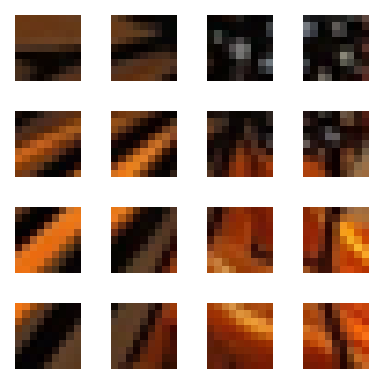} &
        \includegraphics[width=\linewidth]{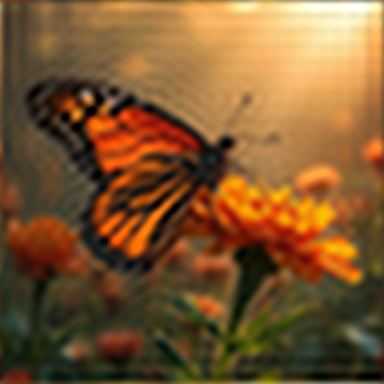} &
        \includegraphics[width=\linewidth]{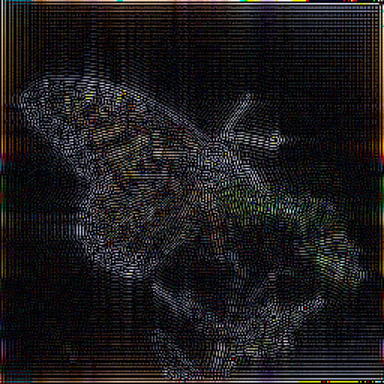} \\
    \end{tabular}
\caption{To probe which aspects of synthetic images are most affected, we transform images to suppress or amplify the effects of distortions in a given domain. To separate the effect of low and high level details, we measure the performance gap when training in depth space, which removes textures, and training a low-receptive-field (visualized in the figure) classifier which operates on $9\times9$ image patches and hence does not rely on structure. To separate the effects of high and low frequency distortions, we train on low and high-pass filtered images. Removing offending features should close the gap with relation to RGB, while removing non-offending features should widen it. }
    \label{fig:transformed_domain_figure}
\end{figure*}

\section{Related Work}

\subsection{Text-to-Image Generation}
Text-to-image (T2I) generation has advanced rapidly in recent years, largely driven by diffusion-based architectures. 
Early systems such as Stable Diffusion~\cite{rombach2022sd15} established latent diffusion as the dominant framework for controllable T2I synthesis. 
Subsequent versions like SDXL~\cite{podell2023sdxl} improved sampling efficiency through distillation and introduced higher-resolution latent spaces that enhanced visual fidelity. 
More recent models, including FLUX~\cite{blackforestlabs2024flux1dev}, Qwen-Image~\cite{wu2025qwen}, and Lumina Image 2.0~\cite{qin2025lumina}, further unify text and image tokenization and refine the diffusion process itself for sharper, more coherent generation. 
Despite these advances in fidelity and controllability, their effectiveness as generators of \emph{useful training data} remains largely untested.

\subsection{Synthetic Vision Data}
\citet{tian2023stablerep} reported that self-supervised representation learning on synthetic data can match or even surpass real data performance.
Similarly, \citet{lomurno2024stable} showed that class-conditioned and fine-tuned diffusion models can occasionally produce competitive synthetic datasets, though primarily in small-scale, low-resolution domains.
In a similar fashion, \citet{hammoud2024synthclip} successfully performs large-scale training of a CLIP model on fully synthetic vision-language data but with much lower data efficiency as compared to using real data.
On the other hand, \citet{geng2024unmet} systematically compared models trained on synthetic images from Stable Diffusion~\cite{rombach2022sd15} with those trained on images retrieved from Stable Diffusion's training dataset.
They found that synthetic datasets consistently underperform real ones when evaluated on real test data, aligning with our observations. 
Our work reveals that the performance gap identified in prior work remains inadequately addressed in the latest state-of-the-art T2I models and while they excel at visual fidelity, they increasingly diverge from the data realism required for effective learning.

\subsection{Prompt Engineering for Synthetic Training}
A complementary line of work focuses on improving the input \emph{text prompts}. 
Caption-in-Prompt (CiP)~\cite{lei2023image,fan2024scaling} extends class labels with short descriptive captions that add semantic context and enhance object-background separation. 
Without modifying the generator, this approach consistently improves classifier performance on synthetic images produced by Stable Diffusion models. 
These studies suggest that richer prompts can partially recover the diversity lost when relying solely on class names. 
Building on this idea, we also analyze how prompt detail interacts with successive generations of T2I models. 
Our results show that newer models need very detailed prompts to produce useful training data.
Yet even under detailed captioning, texture quality remains limited, and high-frequency detail continues to misalign.

\subsection{Dataset Distillation with Generative Priors}
Another direction seeks to \emph{learn} synthetic data distributions that are directly optimized for downstream performance. 
Instead of relying solely on pretrained generators, these methods train compact synthetic datasets under a generative prior to retain the training signal of a large real dataset. 
Recent approaches such as GLaD~\cite{cazenavette2023generalizing}, LD3M~\cite{moser2024latent}, and D$^4$M~\cite{su2024d} extend dataset distillation into the generative domain by coupling diffusion or adversarial priors with gradient- or feature-level matching. 
These methods explicitly optimize for \emph{data utility}, often achieving high performance with only a fraction of the original data, albeit at high computational cost. 
In contrast, our study evaluates \emph{off-the-shelf} T2I models that are optimized for visual realism rather than training utility, showing that generative progress does not automatically translate into useful data for learning. 
Given their shared reliance on generative priors, we expect similar limitations to emerge in this line of work as well, underscoring the need for learnability-centered evaluation and design.

\section{Methodology}

Building on prior work~\cite{geng2024unmet,corvi2023intriguing,astolfi2024consistency}, we identify three main factors likely responsible for the observed performance gap:  
\textbf{(i)} texture and structure distortion,  
\textbf{(ii)} high-frequency distortion, and  
\textbf{(iii)} distributional drift and diversity collapse.

To isolate their effects, we conduct controlled transformations of the data (\autoref{fig:transformed_domain_figure}).
For \textbf{(i)}, we modify images to either suppress or emphasize texture and structure, respectively.
For \textbf{(i)}, we modify images by high- and low-pass filtering.
For \textbf{(iii)}, we analyze the structural alignment of the synthetic and real data distributions using coverage and density metrics that separately quantify fidelity and diversity.

\subsection{Low-Level Texture \& High-Level Structure}

Previous work by \citet{geng2024unmet} shows that even relatively unnoticeable generator artifacts are strong enough to damage class-relevant details in fine-grained classification tasks.
Moreover, convolutional neural networks are known to exhibit a \emph{texture bias}~\cite{geirhos2018imagenet}, which potentially makes them particularly sensitive to distortions and artifacts at the texture level.

To quantify the contribution of such low-level artifacts, we compare the performance gap of models trained on transformed synthetic and real data in two settings: %
A model trained to classify solely based on \textbf{Structure} and another solely based on \textbf{Texture}. The structure classifier is a ResNet-50 operating on depth maps obtained by monocular estimation using the Depth Anything V2~\cite{yang2024depth} model (see \autoref{fig:transformed_domain_figure}: Depth).
The texture classifier is a BagNet~\cite{brendel2019approximating} classifier which operates on $9\times9$ image patches (see \autoref{fig:transformed_domain_figure}: RGB Patches).
Depth space removes all color and textural information, leaving only the overall structures and shapes intact,  while the low receptive field of BagNet makes it incapable of seeing beyond textures.
Structure models are trained \emph{and} tested in depth space to account for drop due to information removal.
If texture artifacts are a key contributor to the observed performance gap, we expect the gap to decrease in the structure-based setting.
In the structure-based setting, problematic textures are removed, in contrast to the texture-based setting, where they dominate the signal.

\subsection{High-Frequency Distortion}

Natural images follow an approximate power-law amplitude spectrum, with energy decaying as a function of spatial frequency \( f \) according to \( S(f) \propto f^{-\alpha} \)~\cite{baradad2021learning,fort2025direct}. 
However, \citet{corvi2023intriguing} observed that diffusion-based image generators deviate from this natural distribution, often exhibiting a shift toward high-frequency components. 
At the same time, CNNs are known to be biased toward high-frequency information~\cite{gavrikov2024can}, making them particularly sensitive to high-frequency spectral distortions.

To quantify the impact of such distortions, we train and evaluate ResNet-50 on high- and low-pass-filtered versions of real and synthetic images (see \autoref{fig:transformed_domain_figure}: Low \& High Frequencies).
The high-pass condition quantifies the performance gap on images that retain frequencies ${f \leq 0.2 \times f_N}$, while in the low-pass setting images retain ${f \geq 0.8 \times f_N}$ with $f_N$ referring to the Nyquist frequency. %

If the realism of high-frequency components is degraded, we expect the relative performance gap to shrink under low-pass filtering (which removes the mismatched parts of the spectrum) compared to high-pass filtering (which leaves the mismatched spectrum intact).

\subsection{Distributional Drift and Diversity Collapse}

Prior work has shown that generative models often trade diversity for fidelity, concentrating samples within a limited region of the real data manifold ~\cite{astolfi2024consistency,naeem2020reliable,borji2022pros,saharia2022photorealistic}.
We hypothesize that this fidelity-diversity trade-off (\ie, greater realism of single samples at the cost of reduced realism of the sample distribution) contributes to the performance gap observed in synthetic training data.

To assess this effect, we use coverage and density metrics introduced by \citet{naeem2020reliable} for synthetic data relative to real data.
In short, density counts how many generated samples fall inside local neighborhoods around real images, while coverage measures the fraction of real images that have least one generated sample in their local neighborhood.
These metrics provide complementary insights into how well the synthetic data distribution aligns with the real one. 
High density combined with low coverage indicates over-concentration (\ie, a collapsed or partially shifted synthetic manifold), while jointly low density and coverage imply a substantial domain shift.
We calculate density and coverage metrics with respect to the real training dataset using CLIP-ViT-L's vision head as a feature extractor.
To obtain an accurate estimate of these metrics, we use 100\,000 random samples.

We further probe the relationship between the data distribution and the Synth~→~Real transfer performance (\ie, training a classifier on synthetic data and evaluating it on real data) by comparing it with the performance obtained by testing a model fitted to real data on synthetic test data denotes as Real~→~Synth.
If there is a significant asymmetry towards the latter, it suggests that the synthetic dataset forms overly separable clusters that are easy for real-trained models to classify, yet fail to capture the complex decision boundaries present in real data.

\begin{figure*}[t]
    \centering
    \includegraphics[width=\linewidth]{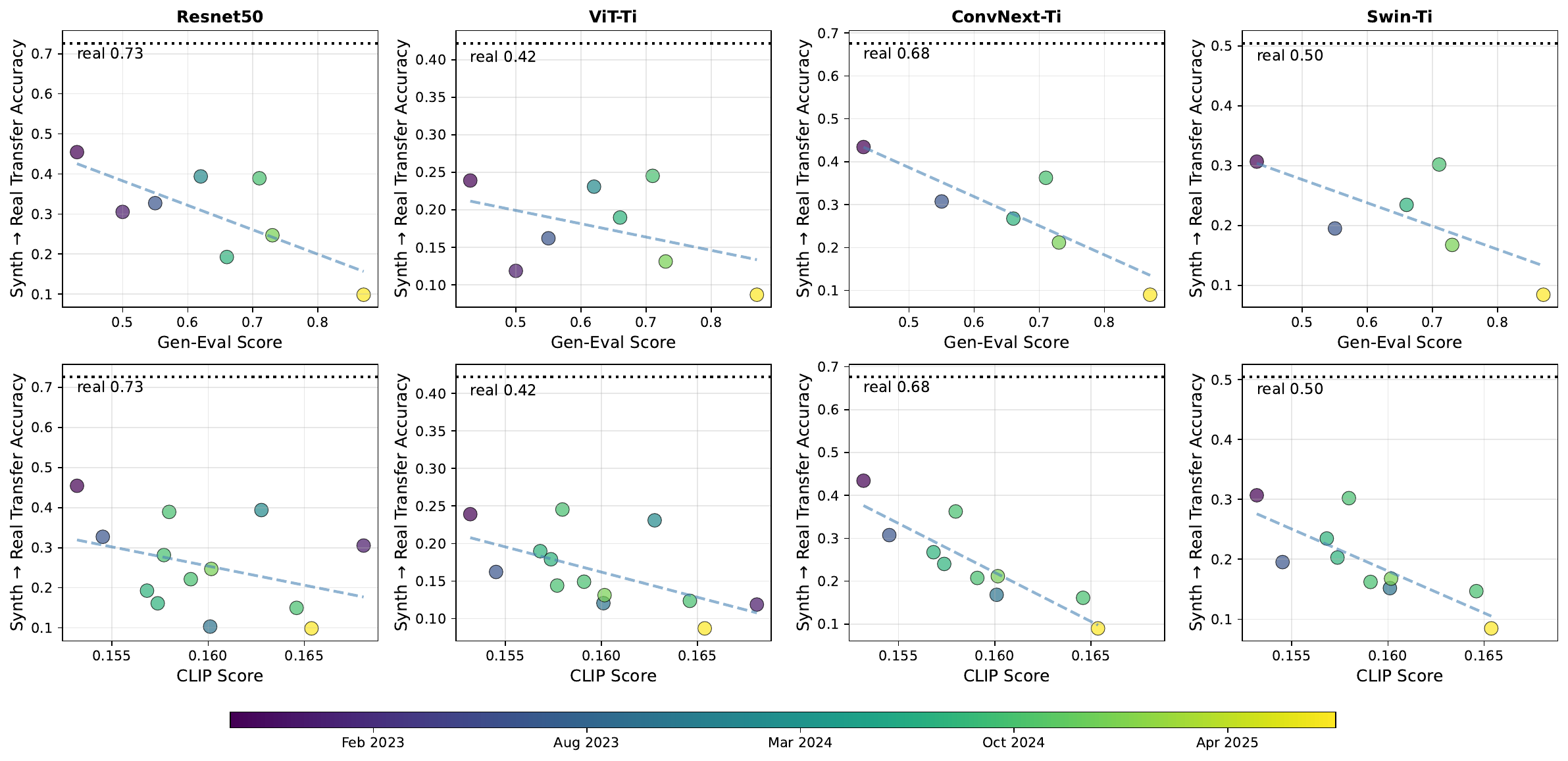}
    \caption{Accuracy on the real ImageNet-1k test set versus GenEval score \textbf{(top)} and  CLIPScore \textbf{(bottom)}. 
Each point represents the performance of a classifier trained on data synthesized by a specific T2I model; the horizontal line indicates the baseline trained on real data. 
Across architectures, we observe a downward trend; higher benchmark scores correspond to lower transfer performance for class label prompts.}
    \label{fig:benchmark_scores}
\end{figure*}

\section{Experiments}
We evaluate thirteen open-source T2I models released between 2022 and 2025 to test whether generative progress translates into better synthetic training data. 
For every model, we measure Synth~→~Real transfer performance and isolate the effects of prompt detail, texture fidelity, spectral \mbox{(im-)balance}, and distributional diversity.

\subsection{Setup}
The following paragraphs describe the data generation process, prompting schemes, and training configurations used throughout this paper.

\paragraph{Dataset Generators.} 
We generate synthetic training datasets using a representative set of open-source T2I diffusion models released between 2022 and 2025. 
Our selection spans multiple generations of model design and scale, allowing us to trace how architectural and training advances affect downstream data utility. 
Specifically, we include Stable Diffusion~V1.5 (2022)~\cite{rombach2022sd15}, Stable Diffusion 2.1 (2023)~\cite{rombach2022sd15}, Stable Diffusion~XL (2023)~\cite{podell2023sdxl}, Stable Diffusion~3.5 (2024)~\cite{stabilityai2024sd35} (large and medium), Sana (2024)~\cite{esser2024scaling}, Flux-Dev (2024)~\cite{blackforestlabs2024flux1dev}, Qwen-Image (2025)~\cite{wu2025qwen}, and Lumina 2.0 (2025)~\cite{qin2025lumina}. 
To evaluate efficiency-oriented variants, we also include distilled counterparts, \ie, SDXL-Turbo (2023)~\cite{podell2023sdxl}, Stable Diffusion~3.5-Large-Turbo (2024)~\cite{esser2024scaling}, and Flux-Schnell (2024)~\cite{blackforestlabs2024flux1dev}.

\paragraph{Datasets.} 
Following prior work on synthetic image evaluation~\cite{sariyildiz2023fake,fan2024scaling,singh2024synthetic}, we use image classification on ImageNet-1k as our benchmark task. 
To reduce computational cost associated with data generation, we create a reduced training subset of ImageNet-1k by randomly sampling 200 classes and 500 images per class (100\,000 images in total). For testing, we use images from the ImageNet-1k validation set corresponding to the sampled classes without any additional sampling (IPC=50).

\paragraph{T2I Generation.} 
We use a low classifier-free guidance scale of 2.0 and 50 denoising steps to balance fidelity and diversity~\cite{sariyildiz2023fake,hammoud2024synthclip}.
For distilled ``turbo'' models, we use four denoising steps.
All models except SDXL, which requires generation at $1024 \times 1024$ pixels, produce $512 \times 512$ pixel images.
All images are then down-sampled to $256 \times 256$.

\paragraph{Prompts.}
We use two types of prompts for T2I generation to determine the impact of text inputs: (\emph{i}) class names and (\emph{ii}) detailed captions.
Detailed captions are produced by recaptioning real images from our training set using GPT-4.1-nano.
We prompt the LLM to produce a two to three-sentence description for each image that captures the background, foreground, and describes all visible objects.
Prompts from detailed captions follow the template by \citet{singh2024synthetic} (\emph{``[class name], [caption]''}) to ensure alignment with the label, even if the prompt fails to mention the class name.

Detailed captions obtained by recaptioning real data serve as a best-case scenario for T2I generation.
However, it is worth noting that this scenario is impractical, as original images are usually \emph{not} available in the fully synthetic T2I dataset generation case.
Additionally, recaptioning each image requires substantial computational cost for large datasets.

\paragraph{Classification Models.}
Our primary architecture is ResNet-50~\cite{he2016deep}. 
To test the generality of observed trends, we additionally evaluate ViT-Ti~\cite{dosovitskiy2020image}, ConvNeXt-Ti~\cite{liu2022convnet}, and Swin-Ti~\cite{liu2021swin} on RGB data. 
All models are trained for 80 epochs with a batch size of 1024 using the Adam optimizer with a learning rate of $10^{-3}$, five epochs of linear warm-up, and a cosine-annealing learning rate schedule. 
Models are validated each epoch on validation data from the training domain (models trained on synthetic data are validated on synthetic data), and the checkpoint with the lowest validation loss is used for evaluation on real data.

\begin{figure*}
    \centering
    \includegraphics[width=\linewidth]{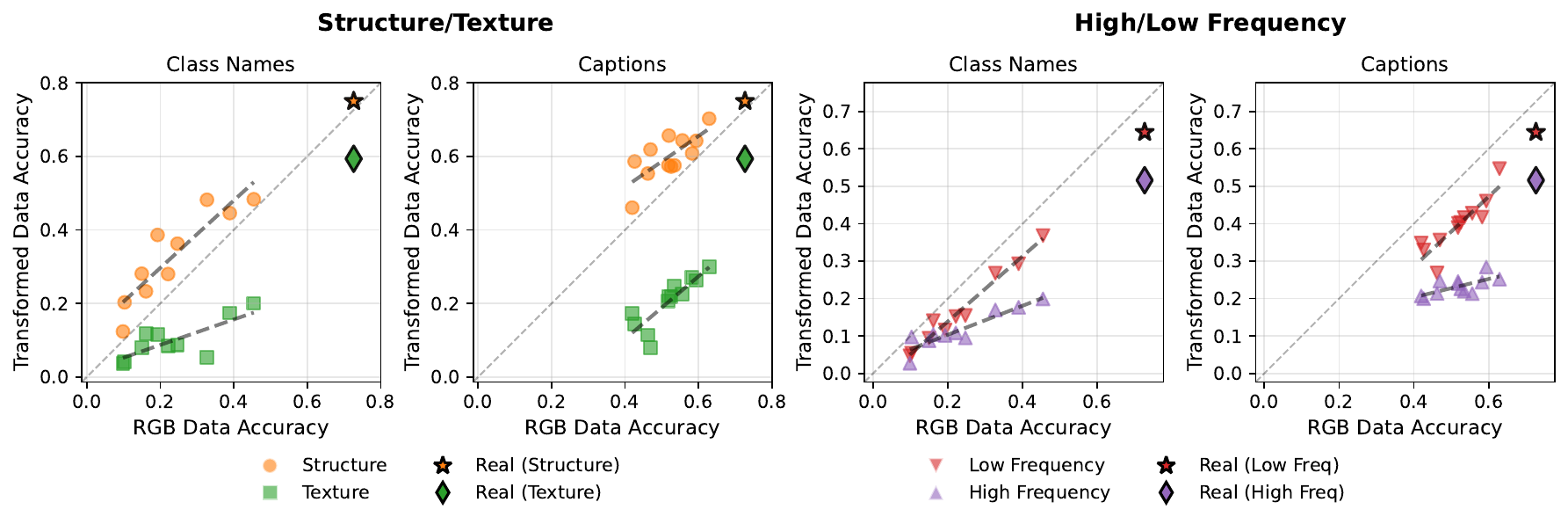}
    \caption{Performance comparison for \textbf{(left)} structure (depth-based classifier) and texture (local feature classifier), and \textbf{(right)} frequency-filtered data for class name- and caption-guided synthetic datasets. Image structure is consistently less affected than texture, while high-frequency components degrade more strongly than low frequencies (especially in better-performing models).}
    \label{fig:rgb_vs_domains}
\end{figure*}

\subsection{Performance Gap Trends}

We begin by examining whether progress in T2I model development translates to better performance when synthetic images are used as training data.

\autoref{fig:main_performance_drop} visualizes the performance of models trained class name-conditioned synthetic data on the real test set over time, while \autoref{fig:benchmark_scores} relates it to the performance on standard T2I alignment metrics.
We consider GenEval~\cite{ghosh2023geneval}, which measures how well text-specified fine-grained object properties (color, position, object count) are represented in the output image, and ClipScore~\cite{hessel2021clipscore}, which measures general alignment between prompt and images in CLIP's shared language-vision embedding space.
We report GenEval scores as computed by \citet{wu2025qwen} and calculate \mbox{CLIP-ViT-L} scores using image-text pairs from the detailed captions condition.

Benchmark results align with intuition and conclusions reported by respective models.
Newer models score higher on text-alignment metrics as the field progresses. This is not unexpected, as achieving controllable T2I synthesis is a major goal in the field of image generation.
However, we also find a surprising trend:
Text-to-image alignment is \emph{inversely} related to synthetic data performance in the class name-conditioned scenario.
These results suggest a hidden tradeoff between prompt following and the performance of a given model as a training data generator.

\subsection{Texture and Structure}

\autoref{fig:rgb_vs_domains} (left) plots ResNet-50 RGB space performance against the performance of classifiers operating in texture and structure spaces.
First, we observe that, for both class names and caption prompts and across all tested models, there is a significant difference between the performance gaps observed in texture and structure space.
Classifiers operating in the structure space consistently demonstrate a much narrower gap to real data than is the case for texture space classifiers.

Models with higher RGB accuracy also display a higher accuracy in \emph{both} texture and structure space, suggesting that \emph{both} image structure and texture affect accuracy on RGB data.
Additionally, we observe that the removal of textural information reduces the Synth.~→~Real gap, compared to the Synth.~→~Real gap in RGB space (\ie, structure space models are above the diagonal in \autoref{fig:rgb_vs_domains} (left)), suggesting that synthetic textures hinder the transferability of models from synthetic to real data.

Although using detailed captions improves performance in both RGB and structure spaces, the improvement in the texture space is only minimal.
This suggests that the presence of low-level artifacts affecting textures is largely decoupled from input captions and, unlike the image structure, cannot be significantly alleviated by better captions.

In practical terms, modern T2I models demonstrate good prompt following and produce images that can appear coherent and realistic to human observers but lack the textural diversity that neural networks rely on for generalization. Thus, improvements in global composition and realism may conceal a growing deficit in textural richness, limiting the effectiveness of synthetic data for training vision models.

\subsection{Power Spectrum}

\autoref{fig:rgb_vs_domains} (right) showcases ResNet-50 RGB space performance against the performance of classifiers operating on high and low frequency images.
The obtained results mirror those from the texture/structure experiment. Performance in the low-frequency domain closely follows the RGB domain, suggesting that smooth spatial transitions and global contrast are faithfully preserved by synthetic images.
This is in contrast to the high-frequency domain, where a more significant gap to RGB space is observed.
Additionally, as in the texture/structure domain experiment, providing detailed captions does not significantly improve high-frequency domain performance; especially compared to improvements in the low-frequency domain.

Together with the texture-structure results, these findings indicate that modern T2I models have become increasingly structure-accurate but texture-deficient. Low-frequency details are generally accurate and can be improved with better text input, while high-frequency details remain broken.

\subsection{Distribution and Diversity}

\begin{figure}
    \centering
    \includegraphics[width=0.98\linewidth]{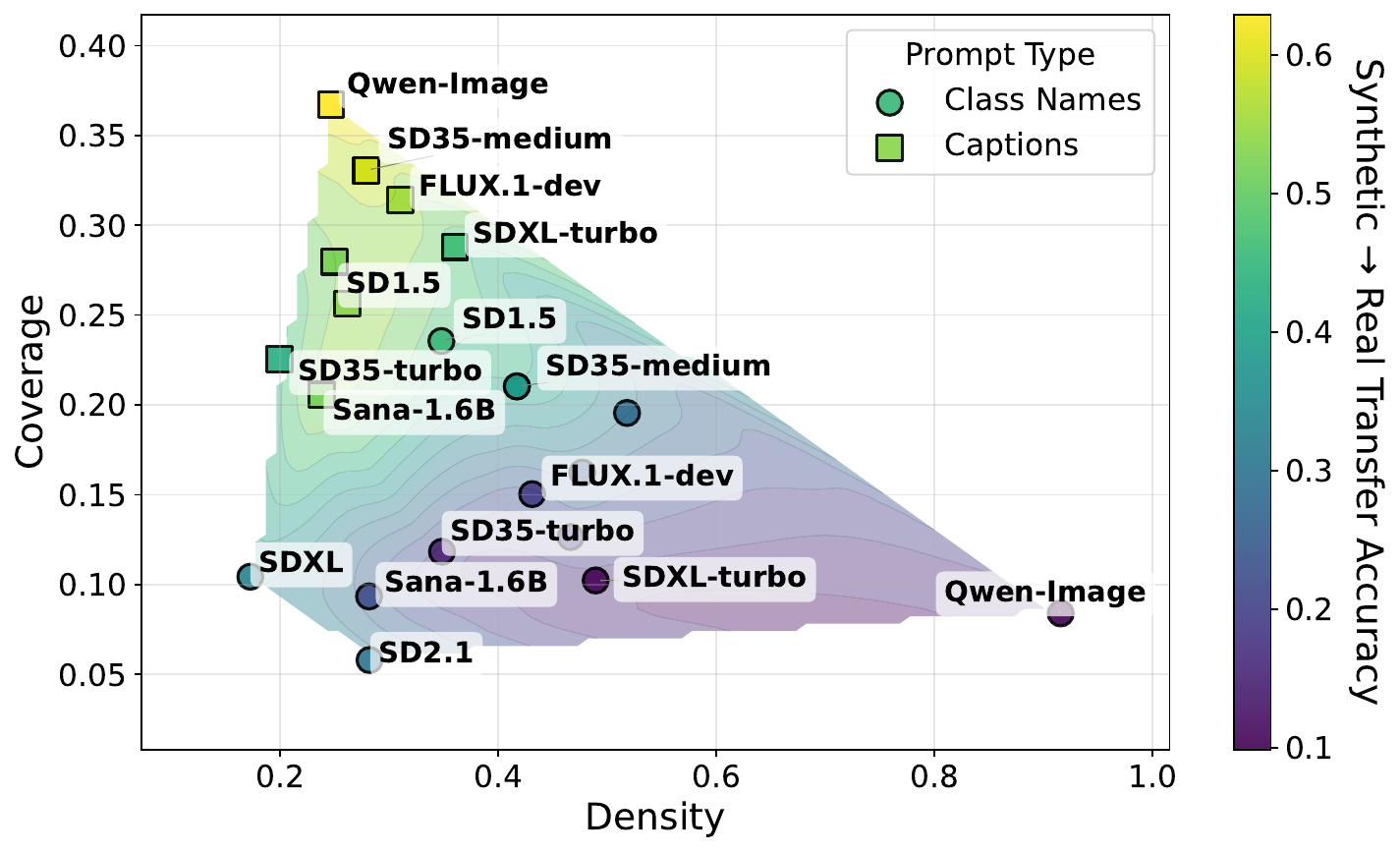}
    \caption{Dataset diversity using density and coverage metrics from \citet{naeem2020reliable}, plotted against classifier accuracy on real data (color). Models with high density but low coverage produce visually consistent yet distributionally narrow samples, while those with higher coverage span a broader portion of real data space and correlate with better generalization.
    Thus, recent T2I models achieve higher sample quality through compact, high-density clusters but sacrifice diversity essential for training quality.}
    \label{fig:density_coverage_acc}
\end{figure}

\autoref{fig:density_coverage_acc} summarizes the relationship between density, coverage, and transfer accuracy.
We find that as density decreases and coverage increases, the transfer accuracy to real data increases.
Many poorly performing models exhibit high density, suggesting their distributions are collapsed.
However, some models, such as SDXL, Sana, or SD2.1, exhibit low density with low coverage, suggesting a significant shift in image distributions.
Using detailed captions derived from real data significantly increases coverage and reduces density.
This change is especially pronounced for models that perform poorly when using class names as prompts.

\begin{figure}
    \centering
    \includegraphics[width=\linewidth]{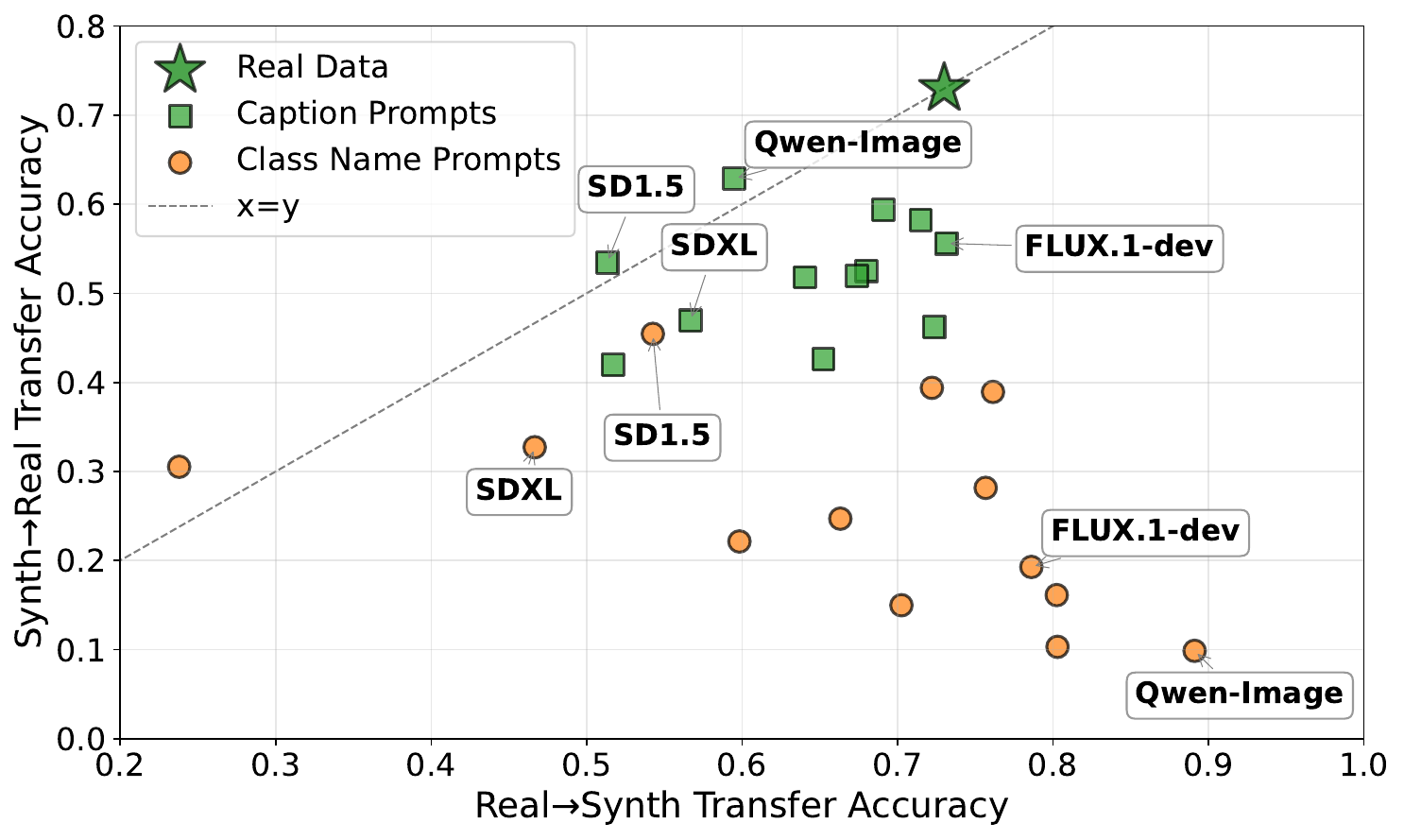}
    \caption{
Comparison of cross-domain transfer for ResNet-50: classification accuracy of models trained on real data and evaluated on synthetic images (Real $\rightarrow$ Synth) versus the reverse setting (Synth $\rightarrow$ Real). %
Synthetic data are increasingly easy for real-trained models to classify, yet models trained on synthetic data transfer progressively worse to real test images, revealing growing asymmetry between visual and learning alignment.
}
\label{fig:cross_transfer}
\end{figure}

To further probe this relationship, \autoref{fig:cross_transfer} compares cross-domain transfer:
How well does a model trained on real data classify synthetic images (Real~→~Synth.) compared to the reverse (Synth.~→~Real)?
For class name prompts, generated data results in poor Synth.~→~Real accuracy; however, Real~→~Synth transfer is comparatively high.
Using detailed captions improves Synth.~→~Real transfer and partially reduces this asymmetry.

Results from this experiment extend observations from density and coverage measurements shown in \autoref{fig:density_coverage_acc}.
Together, these two results imply that existing T2I models tend to prioritize \emph{sample realism} over \emph{distributional realism}.
Individual synthetic images tend to remain relatively easy to classify for a model trained on real data, suggesting that \emph{sample realism} remains relatively intact.
However, \emph{distributional realism} is significantly degraded as suggested by density and coverage values, leading to poor model transfer to real data.
For the class name prompts, the collapsed distribution indicated by high density values appears to be the dominant factor driving the gap, while for detailed captions, the distribution shift indicated by lower coverage appears to be the dominant factor.

The significant difference in density and coverage between the class name and detailed prompts suggests that high prompt-following comes at the cost of a distribution collapse and/or shift, making recent T2I models unsuitable for synthetic data generation unless highly detailed captions are first obtained.

\section{Discussion}

Our results identify multiple patterns that significantly affect the utility of modern T2I models as synthetic data generators.

\paragraph{Prompt Following - Performance Tradeoff.}
We find that these models exhibit a significant trade-off between prompt following and data distribution quality for simple prompts.
Newer T2I models only increase performance when the prompts describe images in detail (\autoref{fig:density_coverage_acc}), while class name prompts lead to a significant performance regression in newer models (\autoref{fig:density_coverage_acc}, \autoref{fig:main_performance_drop}). Prompt following scores increase while the classifier model transfer to real data decreases (\autoref{fig:benchmark_scores}).

\paragraph{Persistent Lack of High Frequency Realism.}
We find that texture and high-frequency details in synthetic vision data are significantly more affected than structure and low-frequency details (\autoref{fig:rgb_vs_domains}).
If textures are removed from the image (via depth estimation), the performance gap is reduced compared to the RGB space.
Respectively, the texture classifier exhibits a significantly higher gap to real data than networks operating in RGB and depth space.
Those conclusions are further supported by measurements of the performance gap on high- and low-frequency filtered images.
The gap is higher in high-frequency space than in low-frequency space.
Moreover, in the best-case scenario, using high-detail captions only significantly improves low-frequency details and the image's structure.
The performance gap in texture and high-frequency spaces remains high, suggesting that high-frequency realism is mostly decoupled from the text input.

\paragraph{Sample vs Data Realism.}
Most models show a strong mismatch between Real~→~Synth. and Synth.~→~Real transfer (\autoref{fig:cross_transfer}).
Real-trained classifiers handle synthetic images well, but synthetic-trained classifiers generalize poorly to real data.
Density-coverage analyses explain this asymmetry:
class-label prompts produce high-density, low-coverage datasets that collapse around narrow aesthetic modes.
Detailed captions improve both density and coverage, showing that low-detail prompts encourage sampling from an overly concentrated subset of the manifold.
In short, synthetic samples look realistic enough to classify, but the overall dataset does not reflect real-world variation and decision boundaries.

\section{Limitations \& Future Work}
Our experiments target image classification on a 200-class subset of ImageNet-1k, chosen to balance scale and computational feasibility.
The observed trends may differ in other domains, such as detection or segmentation, or under larger training datasets. 
We also restrict our evaluation to publicly available T2I models. 
A broader study across proprietary or closed models could refine observed trends.
However, the usefulness of proprietary models for large-scale training data synthesis is limited for a wider community due to restricted access to large-scale inference.

Future work on developing new T2I models should prioritize diversity rather than surface realism, encouraging generators to produce richer textures and more natural frequency statistics instead of optimizing primarily for aesthetic appeal.
It is equally important to evaluate usefulness directly, \ie, by incorporating learnability checks (such as density-coverage measurements), so that improvements in visual realism align more with data realism.
Finally, future work may need to couple generation with learning more, allowing lightweight classifier signals to shape the generation toward improved learnability.

\section{Conclusion}

Modern T2I models generate data that looks aesthetically better but functions worse as reliable training data. 
Across fourteen generators and multiple classifiers, we find that progress in visual fidelity and improved prompt following does \emph{not} lead to better synthetic training data. 
Our analysis points to three drivers of this trend:
First, high-frequency details are degraded, removing class-discriminative cues that image recognition models rely on. 
Second, there is a growing structure-texture imbalance: global composition and layout improve, while fine-scale variation collapses. 
Third, the data distribution narrows, leading to high density but low coverage in feature space. 
Detailed prompts can partially improve structure, but they do \emph{not} fix the core texture and high-frequency domain issues.

We thus advocate for three practical shifts:
First, train for diversity and natural image statistics, not only for photorealism. 
Second, report utility alongside perceptual quality by including density-coverage metrics, probing classification models by training on synthetic and testing on real data, and frequency-band transfer results. 
Third, design prompts and distillation pipelines that preserve intra-class variation and fine details. 
Closing this gap will require coupling generation with learning and rewarding diversity where it matters most: in the features that make models generalize.

{
    \small
    \bibliographystyle{ieeenat_fullname}
    \bibliography{main}

@String(CVPR= {IEEE Conf. Comput. Vis. Pattern Recog.})

@String(NIPS= {Adv. Neural Inform. Process. Syst.})

@String(ICLR = {Int. Conf. Learn. Represent.})

@String(CVPR  = {CVPR})

@String(NIPS  = {NeurIPS})

@String(ICLR  = {ICLR})

@article{geng2024unmet,
  title={The unmet promise of synthetic training images: Using retrieved real images performs better},
  author={Geng, Scott and Hsieh, Cheng-Yu and Ramanujan, Vivek and Wallingford, Matthew and Li, Chun-Liang and Koh, Pang Wei W and Krishna, Ranjay},
  journal=NIPS,
  year={2024}
}

@inproceedings{tian2024learning,
  title={Learning vision from models rivals learning vision from data},
  author={Tian, Yonglong and Fan, Lijie and Chen, Kaifeng and Katabi, Dina and Krishnan, Dilip and Isola, Phillip},
  booktitle=CVPR,
  year={2024}
}

@inproceedings{sariyildiz2023fake,
  title={Fake it till you make it: Learning transferable representations from synthetic imagenet clones},
  author={Sar{\i}y{\i}ld{\i}z, Mert B{\"u}lent and Alahari, Karteek and Larlus, Diane and Kalantidis, Yannis},
  booktitle=CVPR,
  year={2023}
}

@InProceedings{singh2024synthetic,
    author    = {Singh, Krishnakant and Navaratnam, Thanush and Holmer, Jannik and Schaub-Meyer, Simone and Roth, Stefan},
    title     = {Is Synthetic Data all We Need? Benchmarking the Robustness of Models Trained with Synthetic Images},
    booktitle = CVPR,
    year      = {2024}
}

@inproceedings{fang2024data,
  title={Data augmentation for object detection via controllable diffusion models},
  author={Fang, Haoyang and Han, Boran and Zhang, Shuai and Zhou, Su and Hu, Cuixiong and Ye, Wen-Ming},
  booktitle=CVPR,
  year={2024}
}

@article{zhou2023training,
  title={Training on thin air: Improve image classification with generated data},
  author={Zhou, Yongchao and Sahak, Hshmat and Ba, Jimmy},
  journal={arXiv},
  year={2023}
}

@article{hammoud2024synthclip,
  title={SynthCLIP: Are we ready for a fully synthetic CLIP training?},
  author={Hammoud, Hasan Abed Al Kader and Itani, Hani and Pizzati, Fabio and Torr, Philip and Bibi, Adel and Ghanem, Bernard},
  journal={arXiv},
  year={2024}
}

@article{tian2023stablerep,
  title={Stablerep: Synthetic images from text-to-image models make strong visual representation learners},
  author={Tian, Yonglong and Fan, Lijie and Isola, Phillip and Chang, Huiwen and Krishnan, Dilip},
  journal=NIPS,
  year={2023}
}

@inproceedings{rothmeier2024time,
  title={Time to shine: Fine-tuning object detection models with synthetic adverse weather images},
  author={Rothmeier, Thomas and Huber, Werner and Knoll, Alois C},
  booktitle=CVPR,
  year={2024}
}

@inproceedings{saleh2025david,
  title={David: Data-efficient and accurate vision models from synthetic data},
  author={Saleh, Fatemeh and Aliakbarian, Sadegh and Hewitt, Charlie and Petikam, Lohit and Xiao, Xian and Criminisi, Antonio and Cashman, Thomas J and Baltrusaitis, Tadas},
  booktitle=CVPR,
  year={2025}
}

@article{zhu2024odgen,
  title={Odgen: Domain-specific object detection data generation with diffusion models},
  author={Zhu, Jingyuan and Li, Shiyu and Liu, Yuxuan Andy and Yuan, Jian and Huang, Ping and Shan, Jiulong and Ma, Huimin},
  journal=NIPS,
  year={2024}
}

@article{baradad2021learning,
  title={Learning to see by looking at noise},
  author={Baradad Jurjo, Manel and Wulff, Jonas and Wang, Tongzhou and Isola, Phillip and Torralba, Antonio},
  journal=NIPS,
  year={2021}
}

@article{fort2025direct,
  title={Direct ascent synthesis: Revealing hidden generative capabilities in discriminative models},
  author={Fort, Stanislav and Whitaker, Jonathan},
  journal={arXiv},
  year={2025}
}

@article{qin2025lumina,
  title={Lumina-image 2.0: A unified and efficient image generative framework},
  author={Qin, Qi and Zhuo, Le and Xin, Yi and Du, Ruoyi and Li, Zhen and Fu, Bin and Lu, Yiting and Yuan, Jiakang and Li, Xinyue and Liu, Dongyang and others},
  journal={arXiv},
  year={2025}
}

@article{seo2024just,
  title={Just say the name: Online continual learning with category names only via data generation},
  author={Seo, Minhyuk and Cho, Seongwon and Lee, Minjae and Misra, Diganta and Choi, Hyeonbeom and Kim, Seon Joo and Choi, Jonghyun},
  journal={arXiv},
  year={2024}
}

@article{lomurno2024stable,
  title={Stable diffusion dataset generation for downstream classification tasks},
  author={Lomurno, Eugenio and D'Oria, Matteo and Matteucci, Matteo},
  journal={arXiv},
  year={2024}
}

@InProceedings{rombach2022sd15,
    author    = {Rombach, Robin and Blattmann, Andreas and Lorenz, Dominik and Esser, Patrick and Ommer, Bj\"orn},
    title     = {High-Resolution Image Synthesis With Latent Diffusion Models},
    booktitle = CVPR,
    year      = {2022}
}

@article{podell2023sdxl,
  title={Sdxl: Improving latent diffusion models for high-resolution image synthesis},
  author={Podell, Dustin and English, Zion and Lacey, Kyle and Blattmann, Andreas and Dockhorn, Tim and M{\"u}ller, Jonas and Penna, Joe and Rombach, Robin},
  journal={arXiv},
  year={2023}
}

@misc{blackforestlabs2024flux1dev,
  author       = {Black Forest Labs},
  title        = {FLUX.1 [dev] – 12 B-parameter Text-to-Image Model},
  year         = {2024},
  howpublished = {\url{https://huggingface.co/black-forest-labs/FLUX.1-dev}},
}

@article{wu2025qwen,
  title={Qwen-image technical report},
  author={Wu, Chenfei and Li, Jiahao and Zhou, Jingren and Lin, Junyang and Gao, Kaiyuan and Yan, Kun and Yin, Sheng-ming and Bai, Shuai and Xu, Xiao and Chen, Yilei and others},
  journal={arXiv},
  year={2025}
}

@article{kaplan2020scaling,
  title={Scaling laws for neural language models},
  author={Kaplan, Jared and McCandlish, Sam and Henighan, Tom and Brown, Tom B and Chess, Benjamin and Child, Rewon and Gray, Scott and Radford, Alec and Wu, Jeffrey and Amodei, Dario},
  journal={arXiv},
  year={2020}
}

@article{ramanujan2023connection,
  title={On the connection between pre-training data diversity and fine-tuning robustness},
  author={Ramanujan, Vivek and Nguyen, Thao and Oh, Sewoong and Farhadi, Ali and Schmidt, Ludwig},
  journal=NIPS,
  year={2023}
}

@article{brendel2019approximating,
  title={Approximating cnns with bag-of-local-features models works surprisingly well on imagenet},
  author={Brendel, Wieland and Bethge, Matthias},
  journal={arXiv},
  year={2019}
}

@article{yang2024depth,
  title={Depth anything v2},
  author={Yang, Lihe and Kang, Bingyi and Huang, Zilong and Zhao, Zhen and Xu, Xiaogang and Feng, Jiashi and Zhao, Hengshuang},
  journal=NIPS,
  year={2024}
}

@article{hestness2017deep,
  title={Deep learning scaling is predictable, empirically},
  author={Hestness, Joel and Narang, Sharan and Ardalani, Newsha and Diamos, Gregory and Jun, Heewoo and Kianinejad, Hassan and Patwary, Md Mostofa Ali and Yang, Yang and Zhou, Yanqi},
  journal={arXiv},
  year={2017}
}

@article{udandarao2024no,
  title={No" zero-shot" without exponential data: Pretraining concept frequency determines multimodal model performance},
  author={Udandarao, Vishaal and Prabhu, Ameya and Ghosh, Adhiraj and Sharma, Yash and Torr, Philip and Bibi, Adel and Albanie, Samuel and Bethge, Matthias},
  journal=NIPS,
  year={2024}
}

@article{chen2023meditron,
  title={Meditron-70b: Scaling medical pretraining for large language models},
  author={Chen, Zeming and Cano, Alejandro Hern{\'a}ndez and Romanou, Angelika and Bonnet, Antoine and Matoba, Kyle and Salvi, Francesco and Pagliardini, Matteo and Fan, Simin and K{\"o}pf, Andreas and Mohtashami, Amirkeivan and others},
  journal={arXiv},
  year={2023}
}

@misc{oquab2023dinov2,
  title={DINOv2: Learning Robust Visual Features without Supervision},
  author={Oquab, Maxime and Darcet, Timothée and Moutakanni, Theo and Vo, Huy V. and Szafraniec, Marc and Khalidov, Vasil and Fernandez, Pierre and Haziza, Daniel and Massa, Francisco and El-Nouby, Alaaeldin and Howes, Russell and Huang, Po-Yao and Xu, Hu and Sharma, Vasu and Li, Shang-Wen and Galuba, Wojciech and Rabbat, Mike and Assran, Mido and Ballas, Nicolas and Synnaeve, Gabriel and Misra, Ishan and Jegou, Herve and Mairal, Julien and Labatut, Patrick and Joulin, Armand and Bojanowski, Piotr},
  journal={arXiv},
  year={2023}
}

@inproceedings{esser2024scaling,
  title={Scaling rectified flow transformers for high-resolution image synthesis},
  author={Esser, Patrick and Kulal, Sumith and Blattmann, Andreas and Entezari, Rahim and M{\"u}ller, Jonas and Saini, Harry and Levi, Yam and Lorenz, Dominik and Sauer, Axel and Boesel, Frederic and others},
  booktitle=ICML,
  year={2024}
}

@misc{stabilityai2024sd35,
  title        = {Introducing Stable Diffusion 3.5},
  author       = {{Stability AI}},
  year         = {2024},
  month        = {October},
  howpublished = {\url{https://stability.ai/news/introducing-stable-diffusion-3-5}},
  note         = {Accessed: 2025-11-12}
}

@inproceedings{geirhos2018imagenet,
  title={ImageNet-trained CNNs are biased towards texture; increasing shape bias improves accuracy and robustness},
  author={Geirhos, Robert and Rubisch, Patricia and Michaelis, Claudio and Bethge, Matthias and Wichmann, Felix A and Brendel, Wieland},
  booktitle=ICLR,
  year={2018}
}

@article{borji2022pros,
  title={Pros and cons of GAN evaluation measures: New developments},
  author={Borji, Ali},
  journal={Computer Vision and Image Understanding},
  year={2022}
}

@article{lei2023image,
  title={Image captions are natural prompts for text-to-image models},
  author={Lei, Shiye and Chen, Hao and Zhang, Sen and Zhao, Bo and Tao, Dacheng},
  journal={arXiv},
  year={2023}
}

@inproceedings{naeem2020reliable,
  title={Reliable fidelity and diversity metrics for generative models},
  author={Naeem, Muhammad Ferjad and Oh, Seong Joon and Uh, Youngjung and Choi, Yunjey and Yoo, Jaejun},
  booktitle=ICML,
  year={2020}
}

@inproceedings{su2024d,
  title={D\^{4}M: Dataset Distillation via Disentangled Diffusion Model},
  author={Su, Duo and Hou, Junjie and Gao, Weizhi and Tian, Yingjie and Tang, Bowen},
  booktitle=CVPR,
  year={2024}
}

@inproceedings{cazenavette2023generalizing,
  title={Generalizing dataset distillation via deep generative prior},
  author={Cazenavette, George and Wang, Tongzhou and Torralba, Antonio and Efros, Alexei A and Zhu, Jun-Yan},
  booktitle=CVPR,
  year={2023}
}

@article{moser2024latent,
  title={Latent dataset distillation with diffusion models},
  author={Moser, Brian B and Raue, Federico and Palacio, Sebastian and Frolov, Stanislav and Dengel, Andreas},
  journal={arXiv},
  year={2024}
}

@inproceedings{liu2022convnet,
  title={A convnet for the 2020s},
  author={Liu, Zhuang and Mao, Hanzi and Wu, Chao-Yuan and Feichtenhofer, Christoph and Darrell, Trevor and Xie, Saining},
  booktitle=CVPR,
  year={2022}
}

@inproceedings{liu2021swin,
  title={Swin transformer: Hierarchical vision transformer using shifted windows},
  author={Liu, Ze and Lin, Yutong and Cao, Yue and Hu, Han and Wei, Yixuan and Zhang, Zheng and Lin, Stephen and Guo, Baining},
  booktitle=CVPR,
  year={2021}
}

@article{dosovitskiy2020image,
  title={An image is worth 16x16 words: Transformers for image recognition at scale},
  author={Dosovitskiy, Alexey},
  journal={arXiv},
  year={2020}
}

@article{ravi2024sam2,
  title={SAM 2: Segment Anything in Images and Videos},
  author={Ravi, Nikhila and Gabeur, Valentin and Hu, Yuan-Ting and Hu, Ronghang and Ryali, Chaitanya and Ma, Tengyu and Khedr, Haitham and R{\"a}dle, Roman and Rolland, Chloe and Gustafson, Laura and Mintun, Eric and Pan, Junting and Alwala, Kalyan Vasudev and Carion, Nicolas and Wu, Chao-Yuan and Girshick, Ross and Doll{\'a}r, Piotr and Feichtenhofer, Christoph},
  journal={arXiv},
  year={2024}
}

@inproceedings{he2016deep,
  title={Deep residual learning for image recognition},
  author={He, Kaiming and Zhang, Xiangyu and Ren, Shaoqing and Sun, Jian},
  booktitle=CVPR,
  year={2016}
}

@article{grattafiori2024llama,
  title={The llama 3 herd of models},
  author={Grattafiori, Aaron and Dubey, Abhimanyu and Jauhri, Abhinav and Pandey, Abhinav and Kadian, Abhishek and Al-Dahle, Ahmad and Letman, Aiesha and Mathur, Akhil and Schelten, Alan and Vaughan, Alex and others},
  journal={arXiv},
  year={2024}
}

@article{gonzales2023synthetic,
  title={Synthetic data in health care: A narrative review},
  author={Gonzales, Aldren and Guruswamy, Guruprabha and Smith, Scott R},
  journal={PLOS Digital Health},
  year={2023},
  publisher={Public Library of Science San Francisco, CA USA}
}

@article{barisic2022sim2air,
  title={Sim2air-synthetic aerial dataset for uav monitoring},
  author={Barisic, Antonella and Petric, Frano and Bogdan, Stjepan},
  journal={IEEE Robotics and Automation Letters},
  year={2022},
  publisher={IEEE}
}

@inproceedings{chen2023deep,
  title={Deep data augmentation for weed recognition enhancement: A diffusion probabilistic model and transfer learning based approach},
  author={Chen, Dong and Qi, Xinda and Zheng, Yu and Lu, Yuzhen and Huang, Yanbo and Li, Zhaojian},
  booktitle={2023 ASABE Annual International Meeting},
  year={2023},
  organization={American Society of Agricultural and Biological Engineers}
}

@article{koetzier2024generating,
  title={Generating synthetic data for medical imaging},
  author={Koetzier, Lennart R and Wu, Jie and Mastrodicasa, Domenico and Lutz, Aline and Chung, Matthew and Koszek, W Adam and Pratap, Jayanth and Chaudhari, Akshay S and Rajpurkar, Pranav and Lungren, Matthew P and others},
  journal={Radiology},
  year={2024},
  publisher={Radiological Society of North America}
}

@article{xu2023innovative,
  title={Innovative synthetic data augmentation for dam crack detection, segmentation, and quantification},
  author={Xu, Jia and Yuan, Cheng and Gu, Jiaxuan and Liu, Jian and An, Jiong and Kong, Qingzhao},
  journal={Structural Health Monitoring},
  year={2023},
  publisher={SAGE Publications Sage UK: London, England}
}

@article{voetman2023big,
  title={The big data myth: Using diffusion models for dataset generation to train deep detection models},
  author={Voetman, Roy and Aghaei, Maya and Dijkstra, Klaas},
  journal={arXiv},
  year={2023}
}

@inproceedings{kim2024sddgr,
  title={Sddgr: Stable diffusion-based deep generative replay for class incremental object detection},
  author={Kim, Junsu and Cho, Hoseong and Kim, Jihyeon and Tiruneh, Yihalem Yimolal and Baek, Seungryul},
  booktitle=CVPR,
  year={2024}
}

@article{masip2023continual,
  title={Continual learning of diffusion models with generative distillation},
  author={Masip, Sergi and Rodriguez, Pau and Tuytelaars, Tinne and van de Ven, Gido M},
  journal={arXiv},
  year={2023}
}

@article{wen2025highly,
  title={Highly realistic synthetic dataset for pixel-level DensePose estimation via diffusion model},
  author={Wen, Jiaxiao and Chu, Tao and Liu, Qiong},
  journal={Pattern Recognition},
  year={2025}
}

@inproceedings{yang2024robust,
  title={Robust category-level 3d pose estimation from diffusion-enhanced synthetic data},
  author={Yang, Jiahao and Ma, Wufei and Wang, Angtian and Yuan, Xiaoding and Yuille, Alan and Kortylewski, Adam},
  booktitle=CVPR,
  year={2024}
}

@article{saharia2022photorealistic,
  title={Photorealistic text-to-image diffusion models with deep language understanding},
  author={Saharia, Chitwan and Chan, William and Saxena, Saurabh and Li, Lala and Whang, Jay and Denton, Emily L and Ghasemipour, Kamyar and Gontijo Lopes, Raphael and Karagol Ayan, Burcu and Salimans, Tim and others},
  journal=NIPS,
  year={2022}
}

@article{ramesh2022hierarchical,
  title={Hierarchical text-conditional image generation with clip latents},
  author={Ramesh, Aditya and Dhariwal, Prafulla and Nichol, Alex and Chu, Casey and Chen, Mark},
  journal={arXiv},
  year={2022}
}

@inproceedings{fan2024scaling,
  title={Scaling laws of synthetic images for model training... for now},
  author={Fan, Lijie and Chen, Kaifeng and Krishnan, Dilip and Katabi, Dina and Isola, Phillip and Tian, Yonglong},
  booktitle=CVPR,
  year={2024}
}

@inproceedings{corvi2023intriguing,
  title={Intriguing properties of synthetic images: from generative adversarial networks to diffusion models},
  author={Corvi, Riccardo and Cozzolino, Davide and Poggi, Giovanni and Nagano, Koki and Verdoliva, Luisa},
  booktitle=CVPR,
  year={2023}
}

@article{ghosh2023geneval,
  title={Geneval: An object-focused framework for evaluating text-to-image alignment},
  author={Ghosh, Dhruba and Hajishirzi, Hannaneh and Schmidt, Ludwig},
  journal=NIPS,
  year={2023}
}

@inproceedings{gavrikov2024can,
  title={Can biases in ImageNet models explain generalization?},
  author={Gavrikov, Paul and Keuper, Janis},
  booktitle=CVPR,
  year={2024}
}

@article{astolfi2024consistency,
  title={Consistency-diversity-realism Pareto fronts of conditional image generative models},
  author={Astolfi, Pietro and Careil, Marlene and Hall, Melissa and Ma{\~n}as, Oscar and Muckley, Matthew and Verbeek, Jakob and Soriano, Adriana Romero and Drozdzal, Michal},
  journal={arXiv},
  year={2024}
}

@inproceedings{hessel2021clipscore,
  title={Clipscore: A reference-free evaluation metric for image captioning},
  author={Hessel, Jack and Holtzman, Ari and Forbes, Maxwell and Le Bras, Ronan and Choi, Yejin},
  booktitle={Proceedings of the 2021 conference on empirical methods in natural language processing},
  year={2021}
}
}

\newpage

\appendix

\section*{Appendix}

\section{Impact of the VAE}

\begin{table}[h]
\centering
\begin{tabular}{lccc}
\hline
VAE & $Acc_{pixel}$ & $Acc_{highpass}$ & $Acc_{lowpass}$ \\
\hline
- & 73.0 & 51.6 & 64.5 \\
SD1.5 & 70.5 & 37.6 & 63.1 \\
SDXL & 69.7 & 37.1 & 62.8 \\
Flux & 70.6 & 37.7 & 63.4 \\
\hline
\end{tabular}
\caption{We quantify the impact of the VAE by performing the reconstruction of our ImageNet subset and then training on it. An evaluation on unaltered real data reveals that the VAE accounts for a significant drop in accuracy in the high-frequency regime.}
\label{tab:model_accuracy}
\end{table}

\section{Additional Vision Tasks}
In addition to classification, we test object detection and segmentation to determine whether the observed trend holds across other vision tasks.
Tested models do not all support additional spatial ques, so we label generated images using the SAM3 model to obtain bounding boxes and segmentation masks. Due to computational constraints, we reuse the pixel-space images from the class-name case of the main experiment. To provide a fair comparison with real data, the same labelling method is applied to the ImageNet images.

For object detection, we follow torchvision~\footnote{https://github.com/pytorch/vision/tree/6f131f1f56f1b78c6301eb4} recipe and train object detection Faster-RCNN with Imagenet1k Resnet-50 backbone for 26 epochs with Adam optimizer and batch size of 16. For segmentation, we also follow the torchvision recipe for the DeepLabv3 model with Imagenet1k Resnet-50 backbone. We train the model using the Adam optimizer for 30 epochs with a batch size of 20.

\subsection{Object Detection}
\begin{table}[ht]
\centering
\begin{tabular}{l ccc}
\toprule

& \multicolumn{3}{c}{\textbf{Test Scores}} \\
\cmidrule(lr){2-4}

\textbf{Data Source} & AP & AP$_{50}$ & AP$_{75}$ \\

\midrule

ImageNet & 20.2 & 47.6 & 14.1 \\ 
SD V1.5 (Oct 2022) & 11.9 & 32.1 & 6.2 \\
SD V3.0 (Feb 2024) & 11.0 & 28.7 & 6.4 \\
Flux-dev (Aug 2024) & 6.5 & 17.2 & 3.2 \\
Lumina2 (Jan 2025) & 6.1 & 17.3 & 3.2 \\
Qwen-Image (Aug 2025) & 3.1 & 8.2 & 1.7 \\

\bottomrule
\end{tabular}
\caption{Performance of the Faster-RCNN model evaluated on the ImageNet validation subset labeled with SAM3 using the COCO evaluation protocol. We observe that the downward trend also applies to the object detection task, with the SD15 model outperforming recent state-of-the-art models.}
\label{tab:object_detection_results}
\end{table}

\subsection{Semantic Segmentation}

\begin{table}[ht]
\centering
\begin{tabular}{lcccc}
\toprule
 
& \multicolumn{4}{c}{\textbf{Test scores}} \\

\cmidrule(lr){2-5}

\textbf{Data Source}
& \rotatebox{90}{Pixel Acc}
& \rotatebox{90}{mIoU}
& \rotatebox{90}{FWIoU}
& \rotatebox{90}{Dice Score} \\

\midrule

ImageNet & 94.3 & 75.2 & 89.6 & 84.6 \\
SD V1.5 (Oct 2022) & 87.9 & 49.3 & 79.1 & 62.6 \\
SD V3.0 (Feb 2024) & 85.2 & 40.2 & 75.0 & 54.1 \\
Flux-dev (Aug 2024) & 78.2 & 27.9 & 66.5 & 40.6 \\
Lumina2 (Jan 2025) & 80.9 & 26.0 & 67.6 & 38.2 \\
Qwen-Image (Aug 2025) & 75.7 & 12.6 & 59.1 & 19.8 \\

\bottomrule
\end{tabular}
\caption{Performance of the DeepLabv3 model evaluated on the ImageNet validation subset labeled with SAM3. We observe that the downward trend also applies to the semantic segmentation task, with the SD15 model outperforming recent state-of-the-art models. }
\label{tab:segmentation_results}
\end{table}

\section{Performance on Synthetic Test Set}

\begin{table}[ht]
\centering
\begin{tabular}{lccc|ccc}
\toprule
& \multicolumn{3}{c}{\textbf{Real Test Set}}
& \multicolumn{3}{c}{\textbf{Synth Test Set}} \\

\cmidrule(lr){2-4} \cmidrule(lr){5-7}

\textbf{Data Source}
& \rotatebox{90}{Acc}
& \rotatebox{90}{AP}
& \rotatebox{90}{mIoU}
& \rotatebox{90}{Acc}
& \rotatebox{90}{AP}
& \rotatebox{90}{mIoU} \\

\midrule

Imagenet & 73.0 & 20.2 & 47.6 & - & - & - \\
SD1.5 & 45.5 & 11.9 & 49.3 & 79.6 & 25.5 & 71.2 \\
SD2.1 & 30.5 & - & - & 51.6 & - & - \\
SDXL & 30.5 & - & - & 83.2 & - & - \\
SDXL turbo & 10.3 & - & - & 99.2 & - & - \\
SD3.0 & 39.4 & 11.0 & 40.2 & 92.6 & 47.7 & 81.9 \\
SD3.5 medium & 38.9 & - & - & 94.6 & - & - \\
SD3.5 large & 28.2 & - & - & 77.6 & - & - \\
SD3.5 turbo & 15.0 & - & - & 97.6 & - & - \\
Sana & 22.1 & - & - & 97.3 & - & - \\
Flux-dev  & 19.3 & 6.5 & 27.9 & 95.5 & 50.1 & 88.1 \\
Flux-schnell & 16.1 & - & - & 94.4 & - & - \\
Lumina2 & 24.7 & 6.1 & 26.0 & 93.3 & 41.9 & 81.6 \\
Qwen & 9.9 & 3.1 & 12.6 & 97.9 & 69.5 & 93.7 \\

\bottomrule
\end{tabular}
\caption{We evaluate the trained model on the same-sized synthetic validation set obtained from the same data source. We find that synthetic models generally achieve a much higher performance than on the real test set. This excludes bad fit as the source of the observed trend and instead strongly hints at a more easily class-separable data manifold. Accuracy (Acc) reported for the Resnet-50 network.}
\label{tab:real_vs_synth}
\end{table}

\newpage
\onecolumn
\section{Scaling Behavior of Synthetic Data}

\begin{figure*}[!h]
    \centering
    \includegraphics[width=\linewidth]{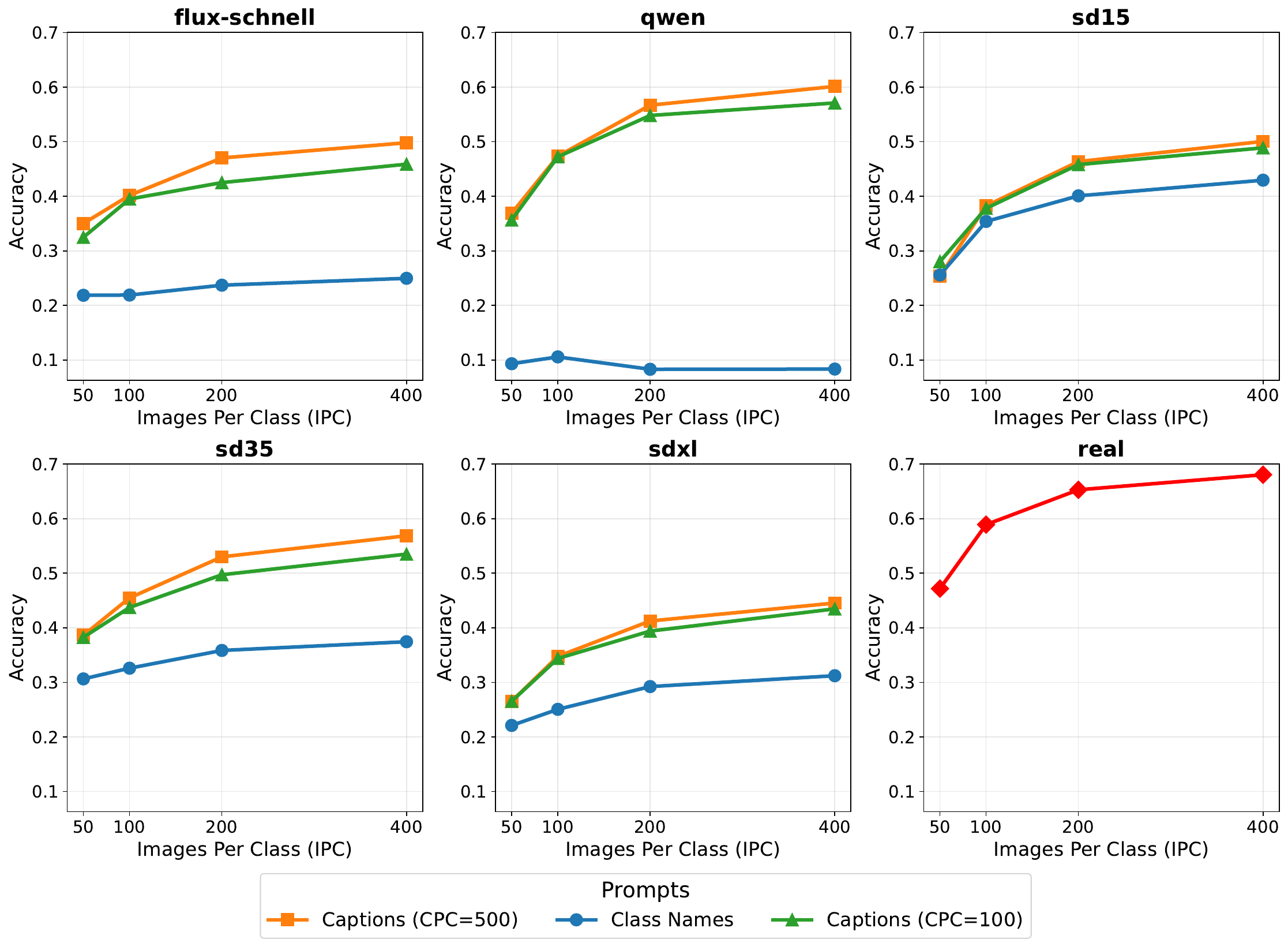}
    \caption{We investigate the scaling of the performance as a function of the number of Images per Class (IPC). We compare class-name prompts to image captions. We additionally test whether generating multiple images from the same caption impacts scaling. To test this, we also add a case in which the number of captions per class (CPC) is reduced to 100 (we generate 5 images per prompt). We observe that models which perform well with captions (qwen, flux-schnell, sd35). Exhibit a significant disparity between the scaling law of class name prompts and caption prompts. Additionally, we observe that when the number of unique prompts is reduced, those models also exhibit a bigger gap in scaling than models performing well with class names. This highlights the importance of using a diverse set of detail prompts for generation with recent models. }
    \label{fig:scaling_figure}
\end{figure*}

\newpage
\section{Synthetic Image Samples}

\begin{longtable}{|l|c|}

\caption{
We visualize random samples from selected classes in the synthetic training dataset. 
Models are sorted by release date. Many of the tested models exhibit a distinct 
"visual style" that differs from real data. For example, select models (SDXL and 
pixelart) are biased towards stylistic, art-like images by default, rather than 
photorealistic images when the prompt is underspecified. Newer models tend to produce 
high-fidelity images; however, backgrounds often become blurry or plain, with the main 
object clearly visible and centered in the frame.
} \\
\hline
\textbf{Data Source} & \textbf{Sample Images (Class Name Prompts)} \\
\hline
\endfirsthead

\multicolumn{2}{c}%
{\tablename\ \thetable\ -- \textit{Continued from previous page}} \\
\hline
\textbf{Data Source} & \textbf{Sample Images (Class Name Prompts)} \\
\hline
\endhead

\hline
\multicolumn{2}{r}{\textit{Continued on next page}} \\
\endfoot

\hline
\endlastfoot

\hline
\multicolumn{2}{|c|}{\textbf{Goldfish}} \\
\hline
ImageNet & \includegraphics[height=3.0cm]{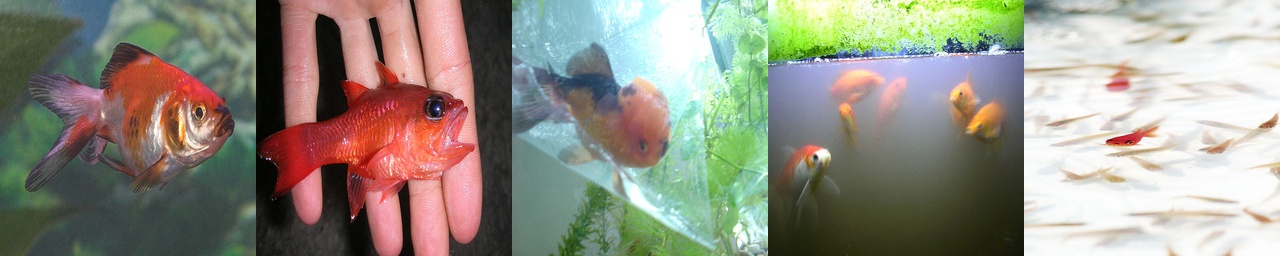} \\
\hline
sd15 & \includegraphics[height=3.0cm]{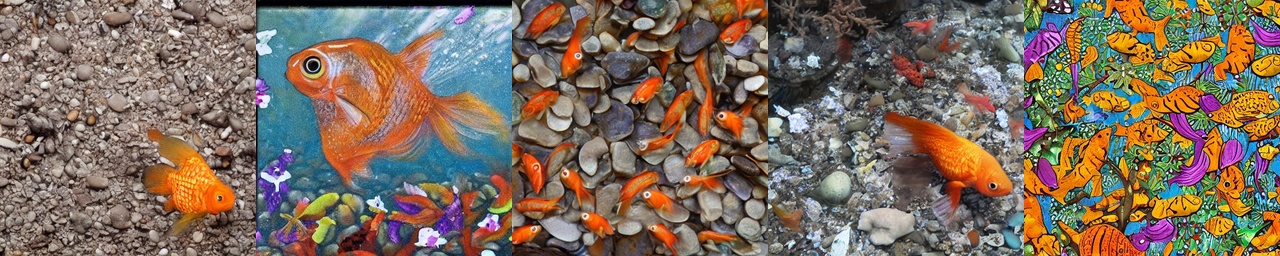} \\
\hline
sd21 & \includegraphics[height=3.0cm]{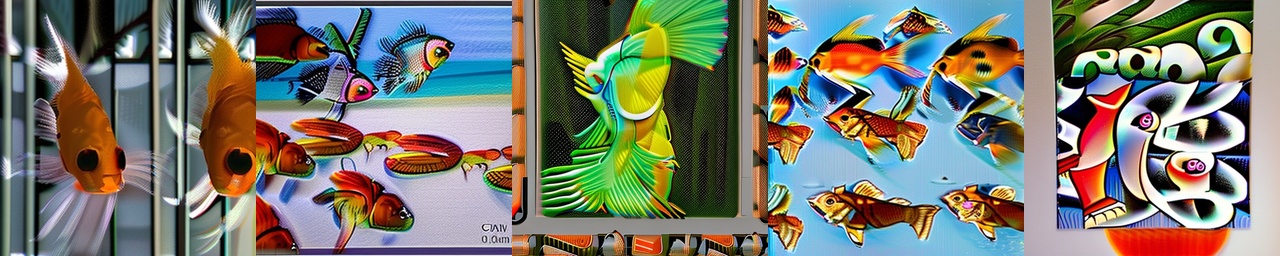} \\
\hline
sdxl & \includegraphics[height=3.0cm]{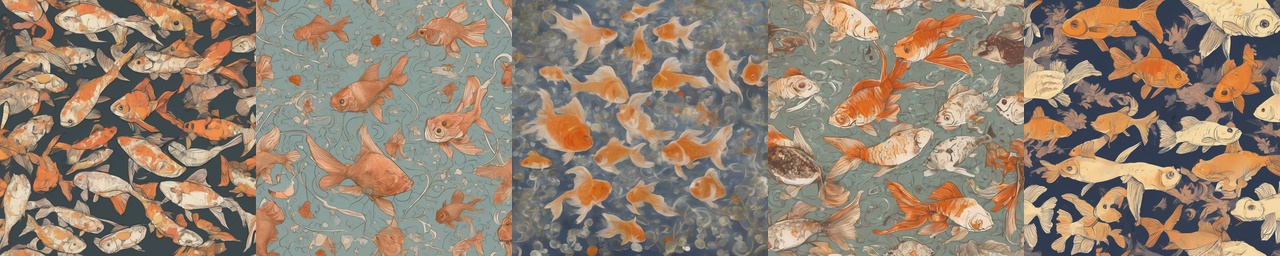} \\
\hline
pixelart & \includegraphics[height=3.0cm]{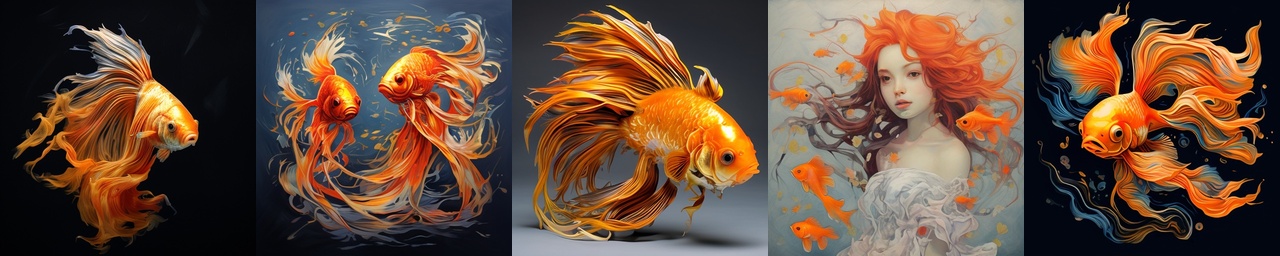} \\
\hline
sdxl-turbo & \includegraphics[height=3.0cm]{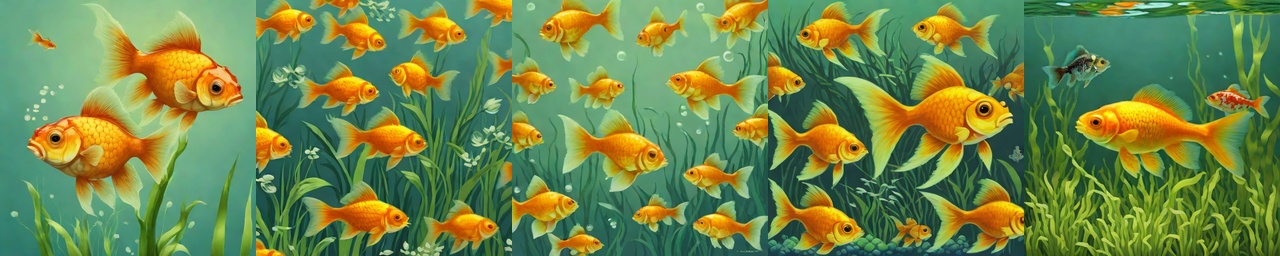} \\
\hline
sd30 & \includegraphics[height=3.0cm]{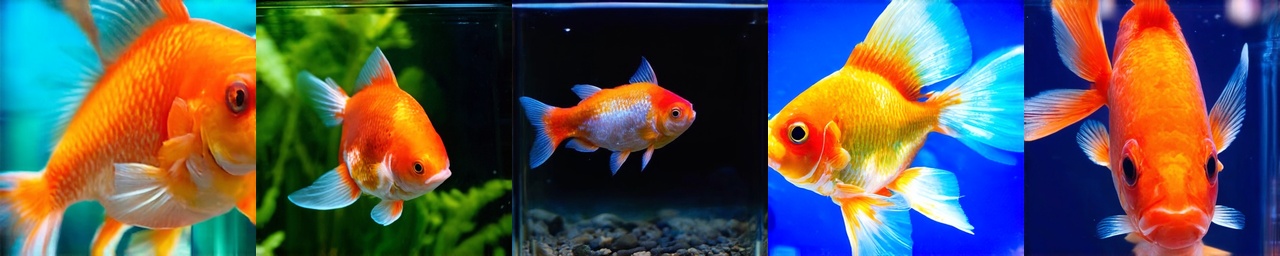} \\
\hline
flux-dev & \includegraphics[height=3.0cm]{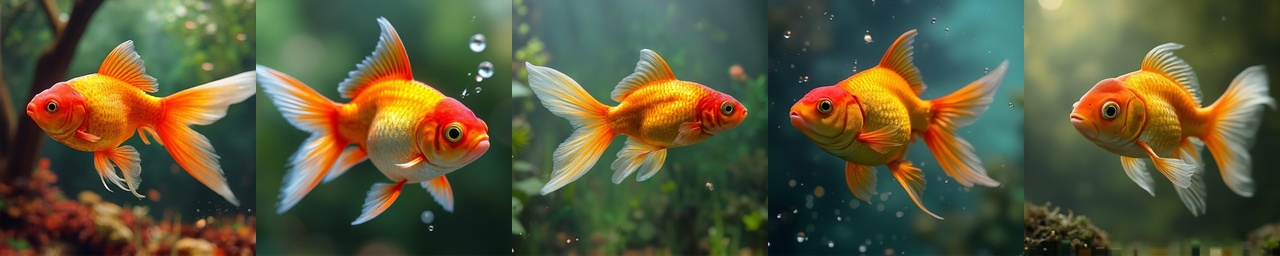} \\
\hline
flux-schnell & \includegraphics[height=3.0cm]{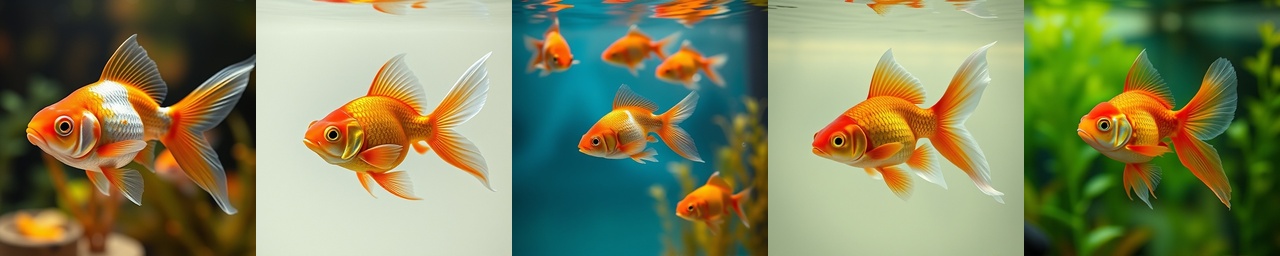} \\
\hline
sd35 & \includegraphics[height=3.0cm]{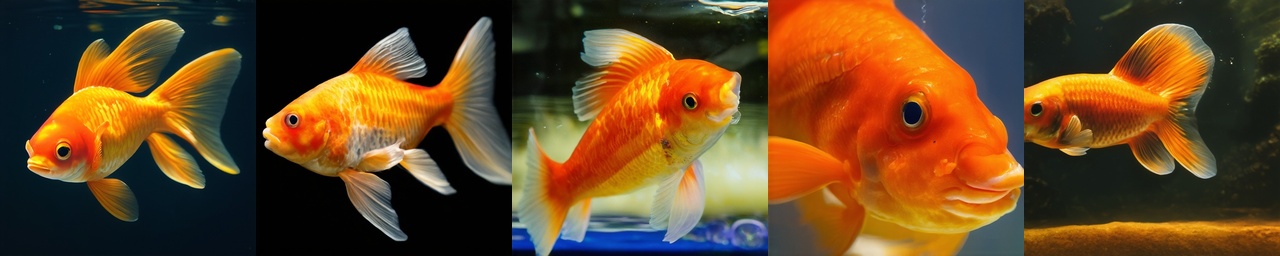} \\
\hline
sd35-large & \includegraphics[height=3.0cm]{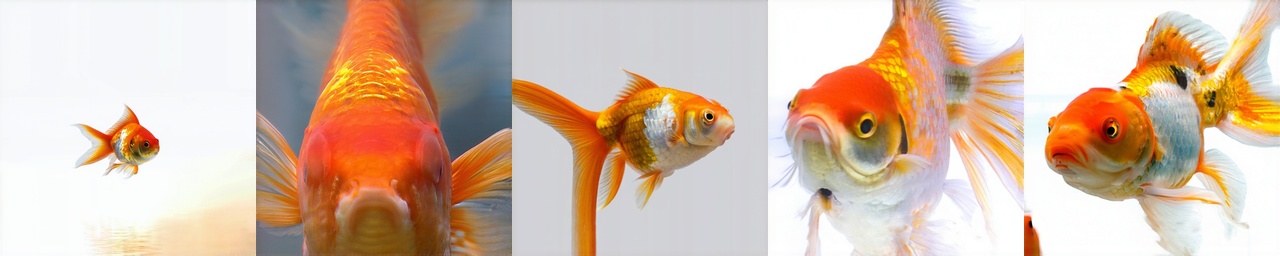} \\
\hline
sd35-turbo & \includegraphics[height=3.0cm]{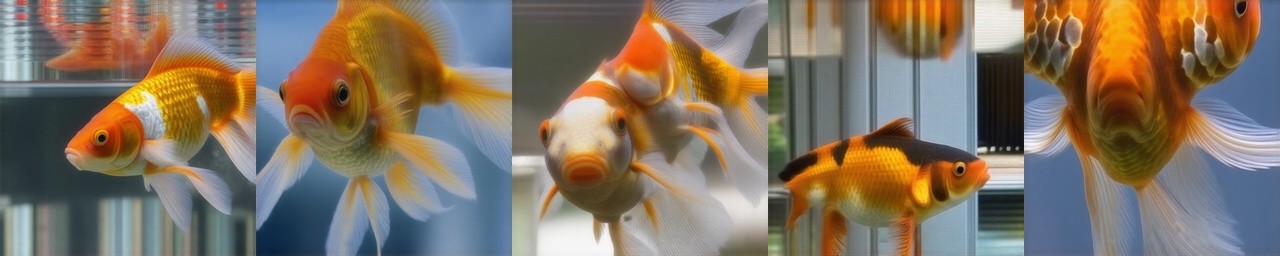} \\
\hline
sana & \includegraphics[height=3.0cm]{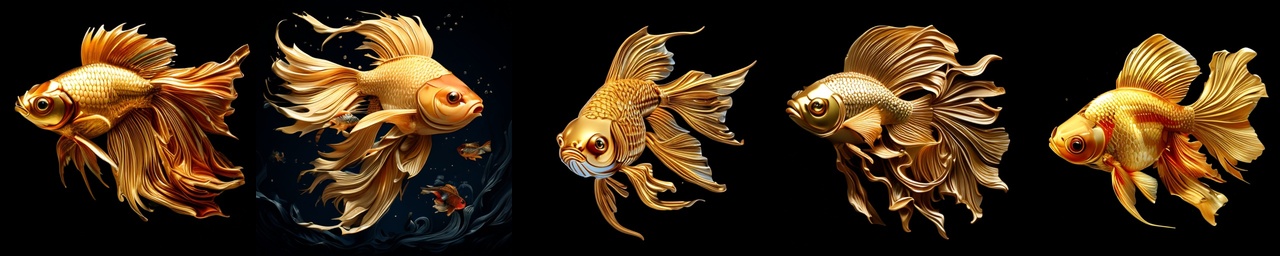} \\
\hline
lumina2 & \includegraphics[height=3.0cm]{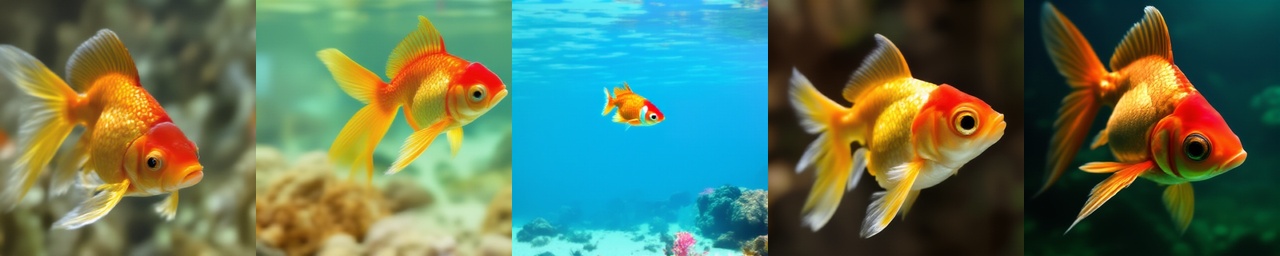} \\
\hline
qwen & \includegraphics[height=3.0cm]{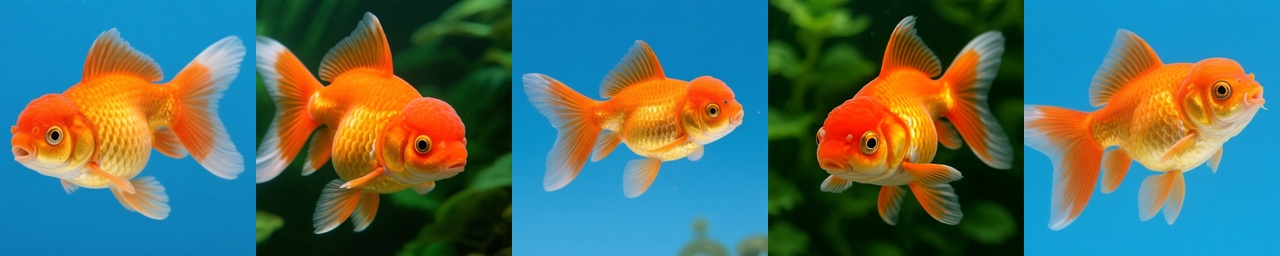} \\
\hline

\hline
\multicolumn{2}{|c|}{\textbf{Monarch}} \\
\hline
ImageNet & \includegraphics[height=3.0cm]{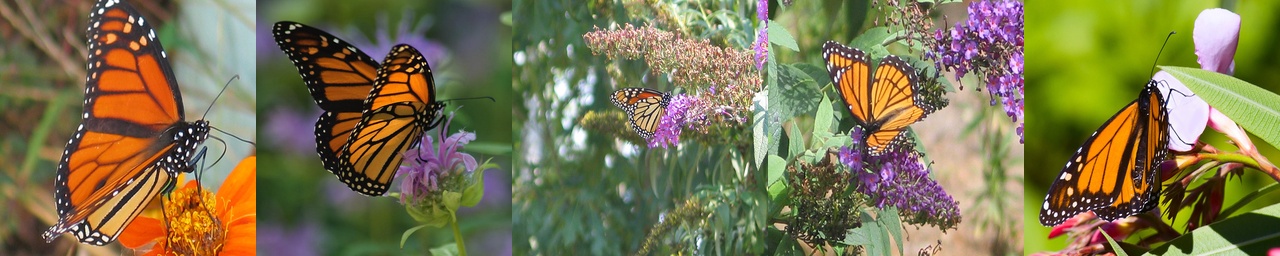} \\
\hline
sd15 & \includegraphics[height=3.0cm]{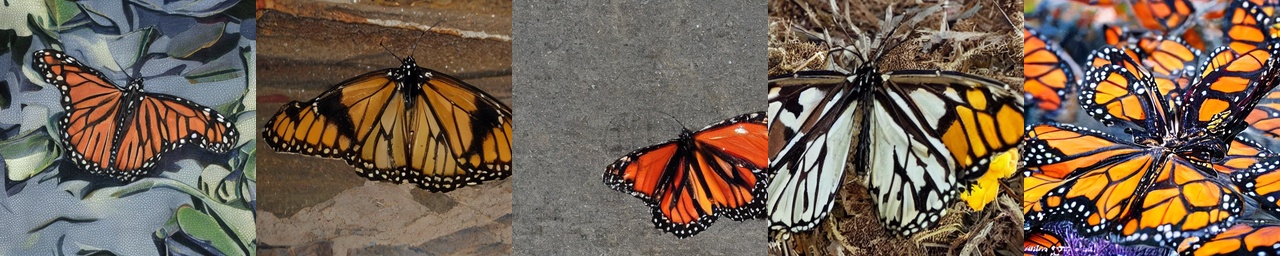} \\
\hline
sd21 & \includegraphics[height=3.0cm]{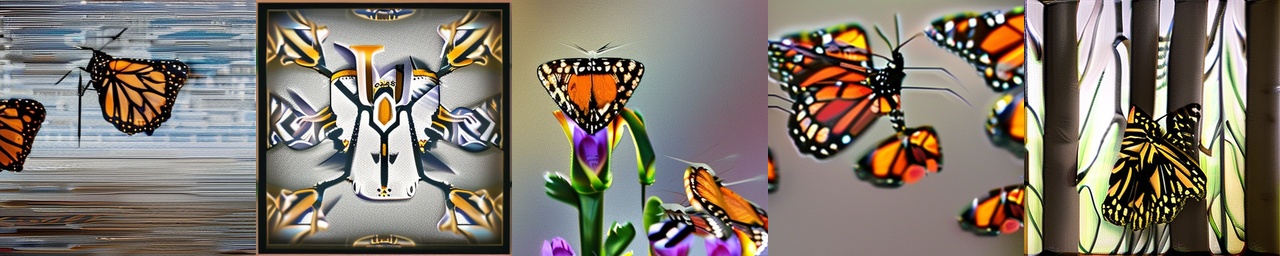} \\
\hline
sdxl & \includegraphics[height=3.0cm]{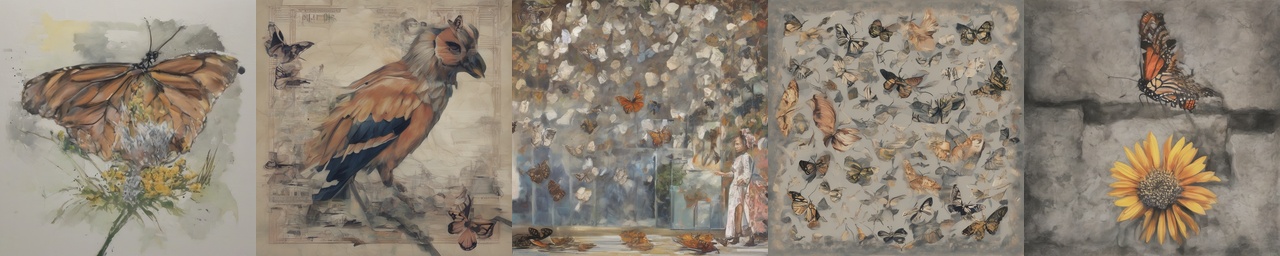} \\
\hline
pixelart & \includegraphics[height=3.0cm]{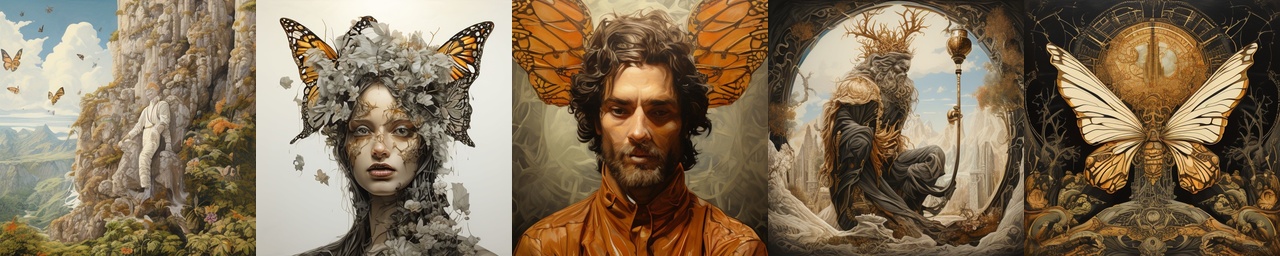} \\
\hline
sdxl-turbo & \includegraphics[height=3.0cm]{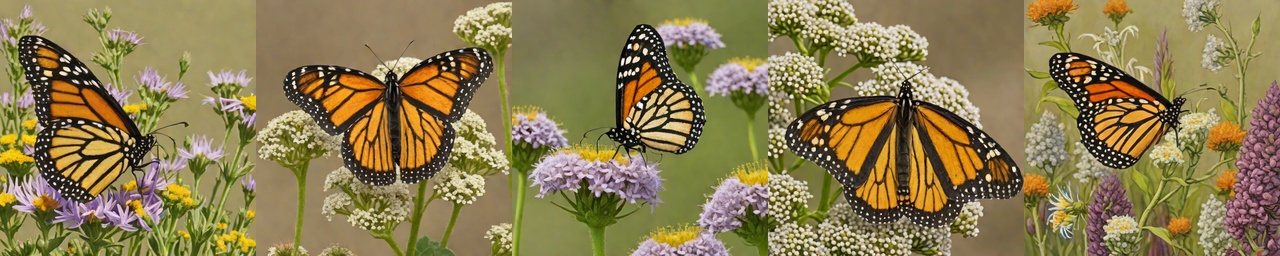} \\
\hline
sd30 & \includegraphics[height=3.0cm]{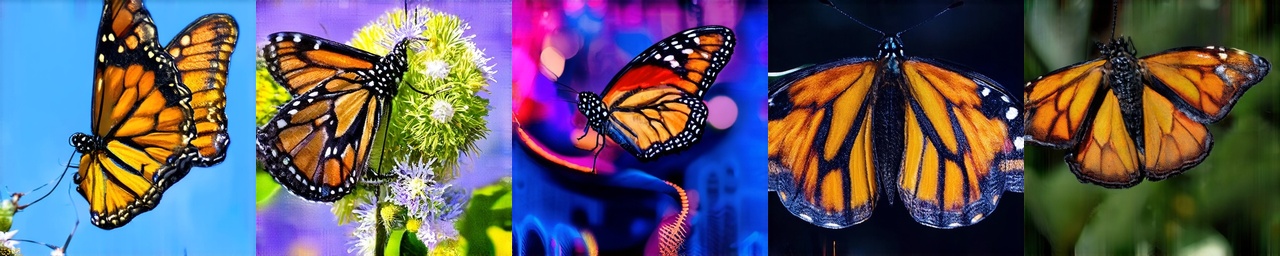} \\
\hline
flux-dev & \includegraphics[height=3.0cm]{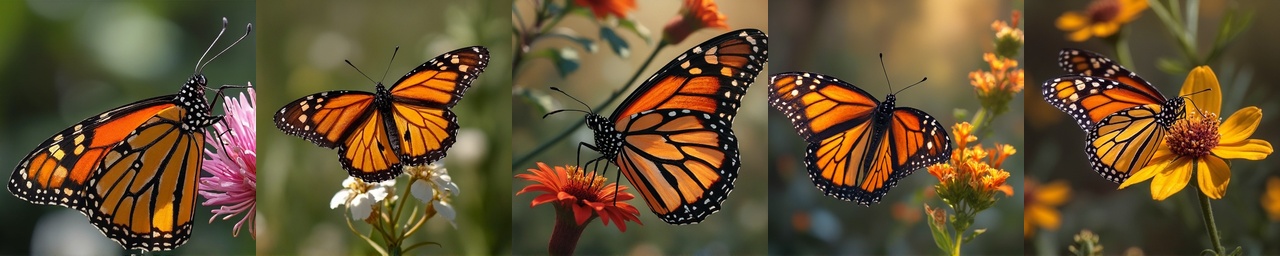} \\
\hline
flux-schnell & \includegraphics[height=3.0cm]{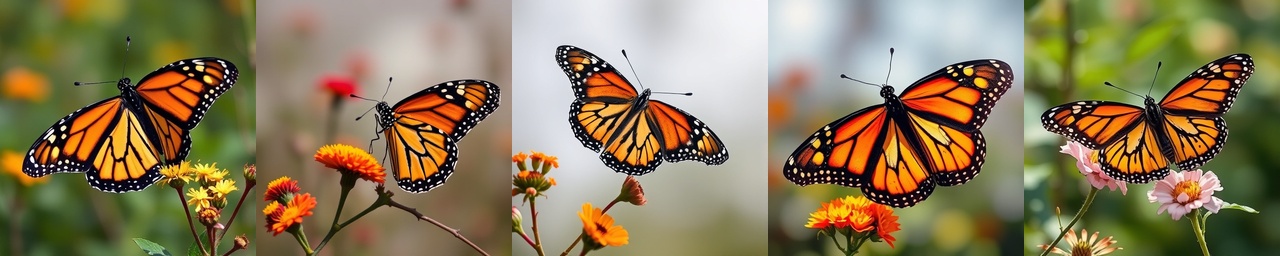} \\
\hline
sd35 & \includegraphics[height=3.0cm]{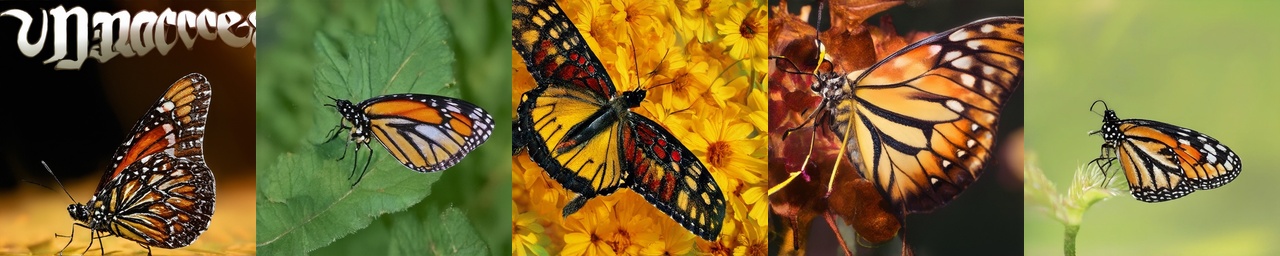} \\
\hline
sd35-large & \includegraphics[height=3.0cm]{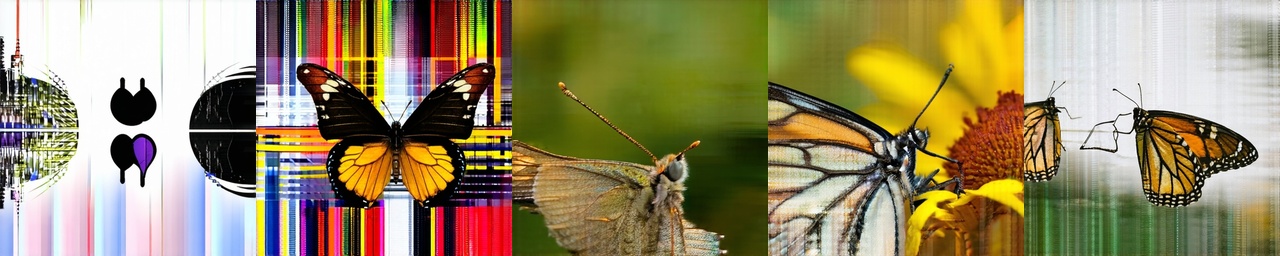} \\
\hline
sd35-turbo & \includegraphics[height=3.0cm]{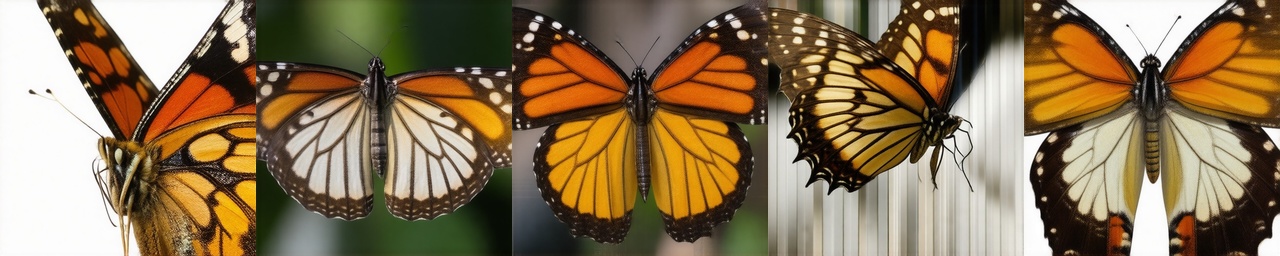} \\
\hline
sana & \includegraphics[height=3.0cm]{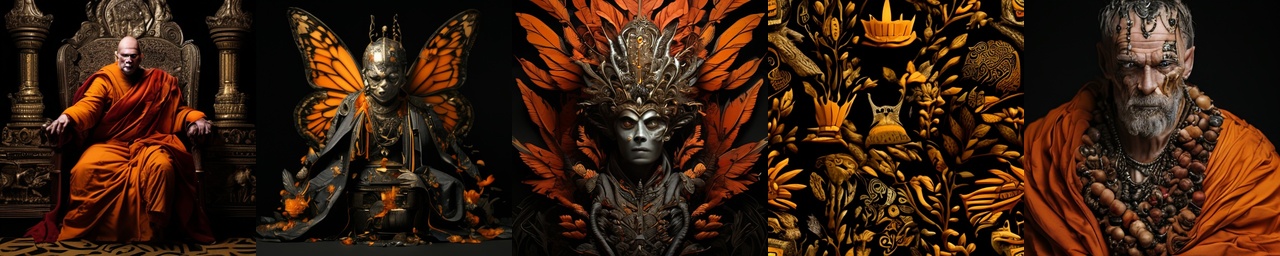} \\
\hline
lumina2 & \includegraphics[height=3.0cm]{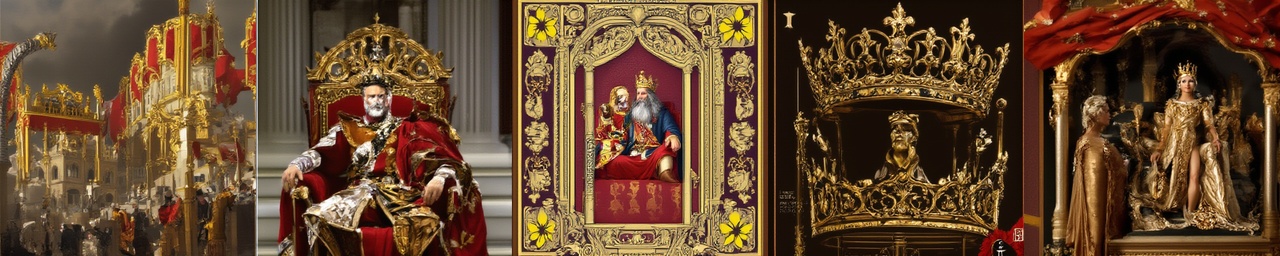} \\
\hline
qwen & \includegraphics[height=3.0cm]{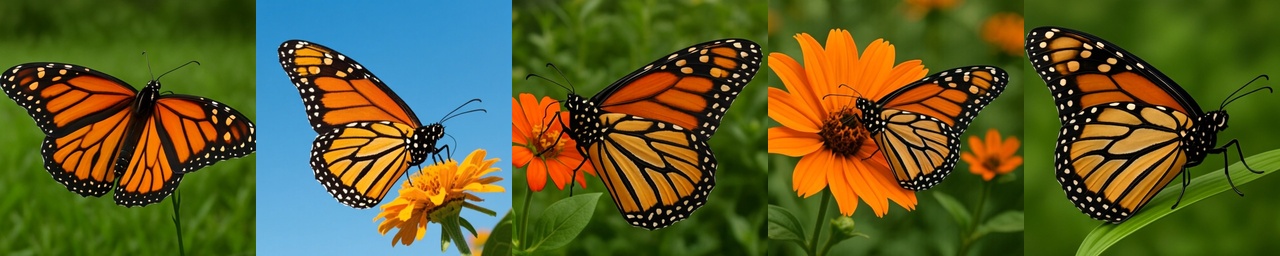} \\
\hline

\hline
\multicolumn{2}{|c|}{\textbf{Koala}} \\
\hline
ImageNet & \includegraphics[height=3.0cm]{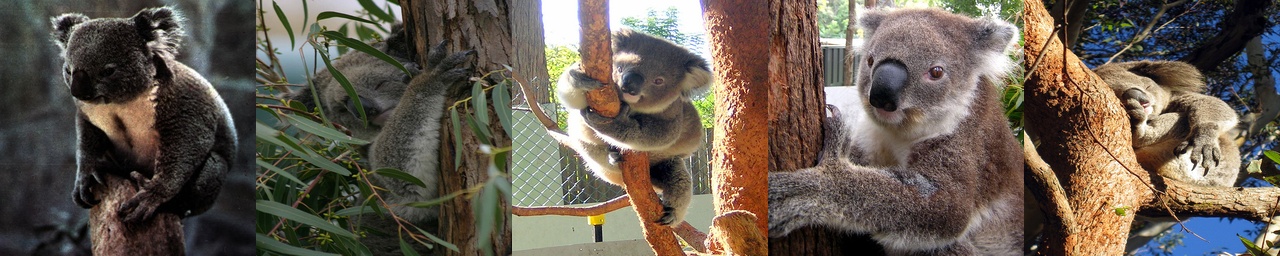} \\
\hline
sd15 & \includegraphics[height=3.0cm]{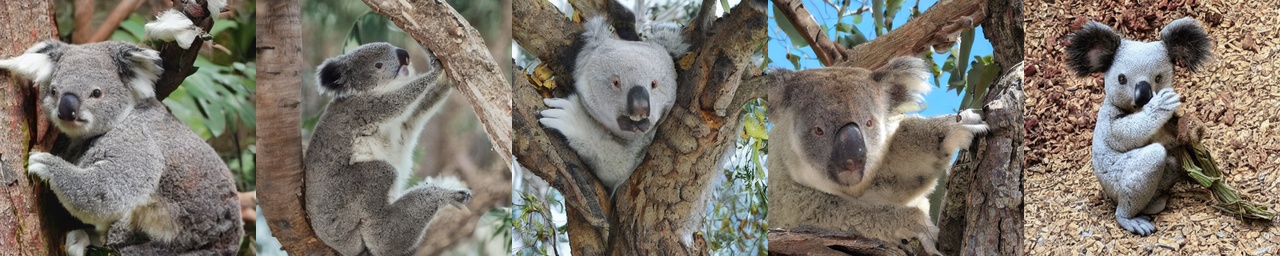} \\
\hline
sd21 & \includegraphics[height=3.0cm]{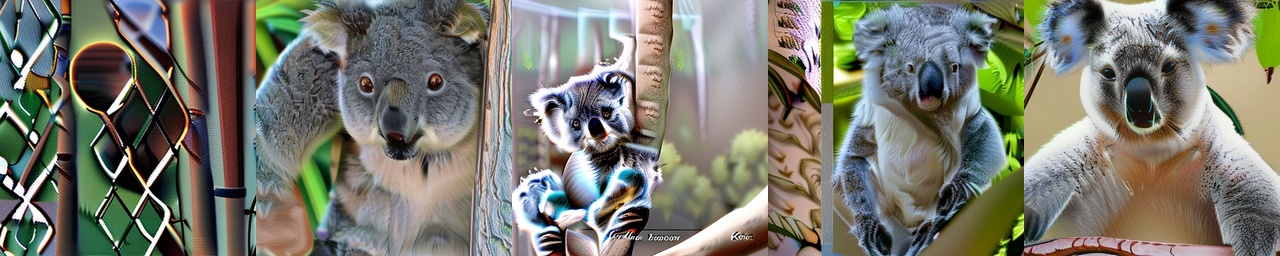} \\
\hline
sdxl & \includegraphics[height=3.0cm]{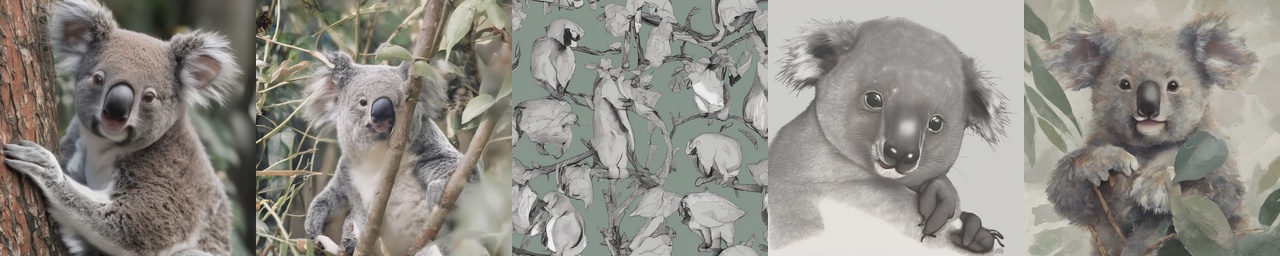} \\
\hline
pixelart & \includegraphics[height=3.0cm]{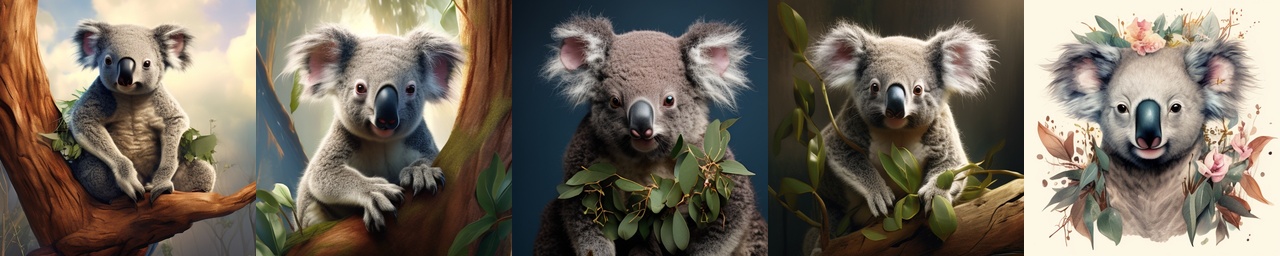} \\
\hline
sdxl-turbo & \includegraphics[height=3.0cm]{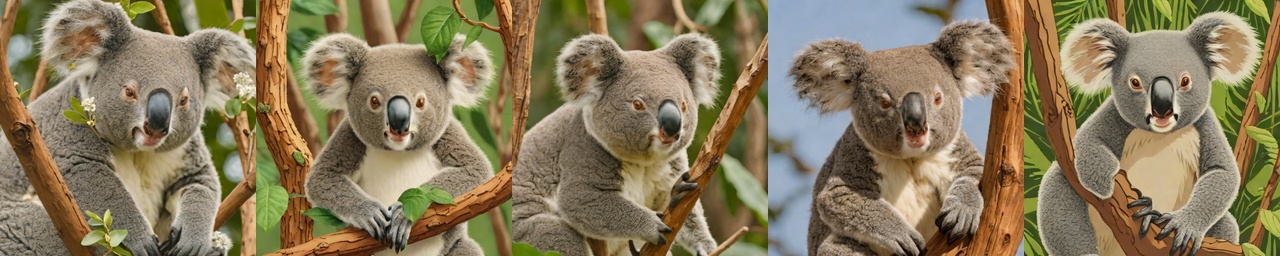} \\
\hline
sd30 & \includegraphics[height=3.0cm]{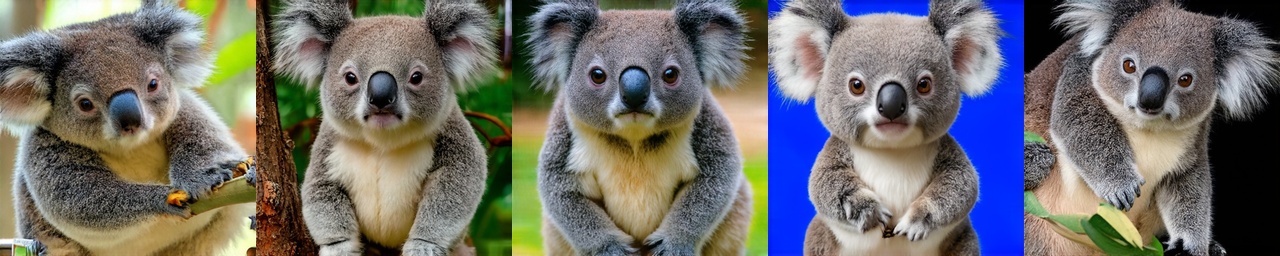} \\
\hline
flux-dev & \includegraphics[height=3.0cm]{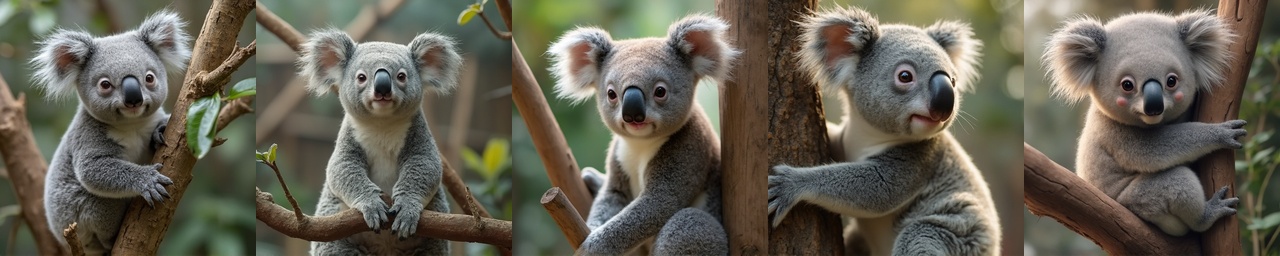} \\
\hline
flux-schnell & \includegraphics[height=3.0cm]{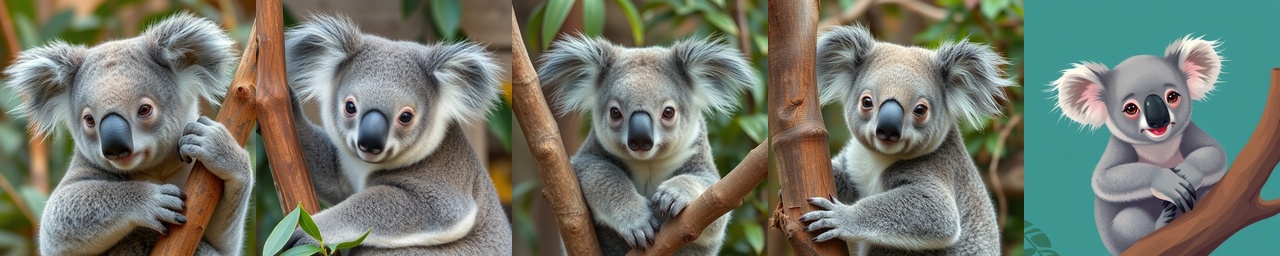} \\
\hline
sd35 & \includegraphics[height=3.0cm]{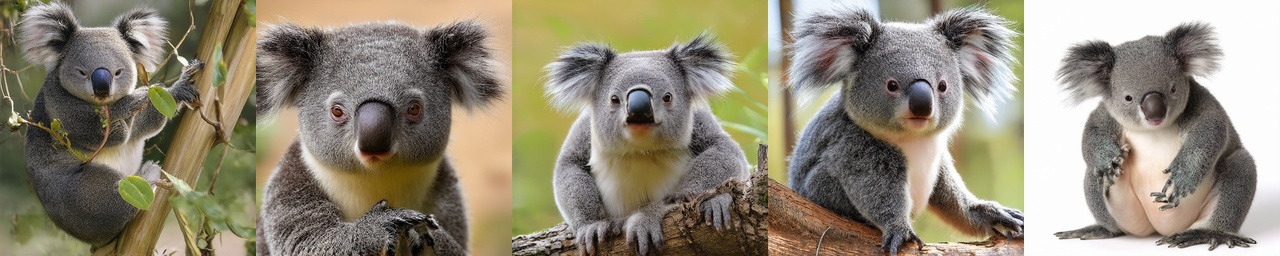} \\
\hline
sd35-large & \includegraphics[height=3.0cm]{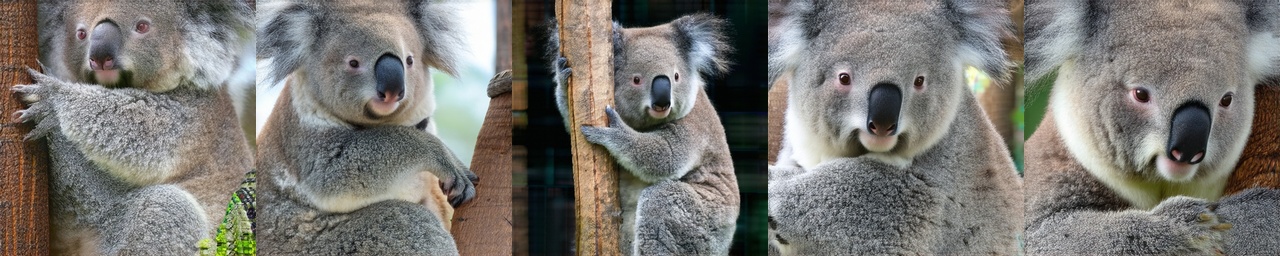} \\
\hline
sd35-turbo & \includegraphics[height=3.0cm]{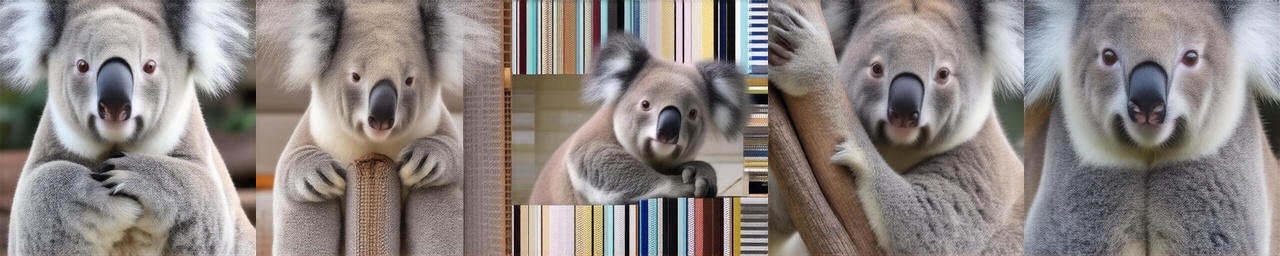} \\
\hline
sana & \includegraphics[height=3.0cm]{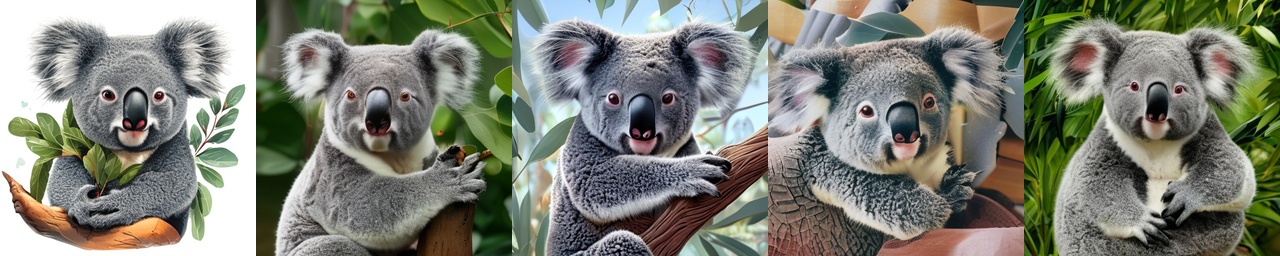} \\
\hline
lumina2 & \includegraphics[height=3.0cm]{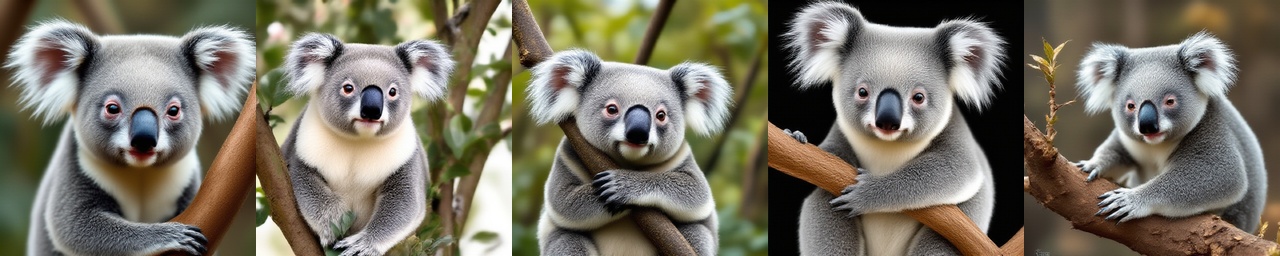} \\
\hline
qwen & \includegraphics[height=3.0cm]{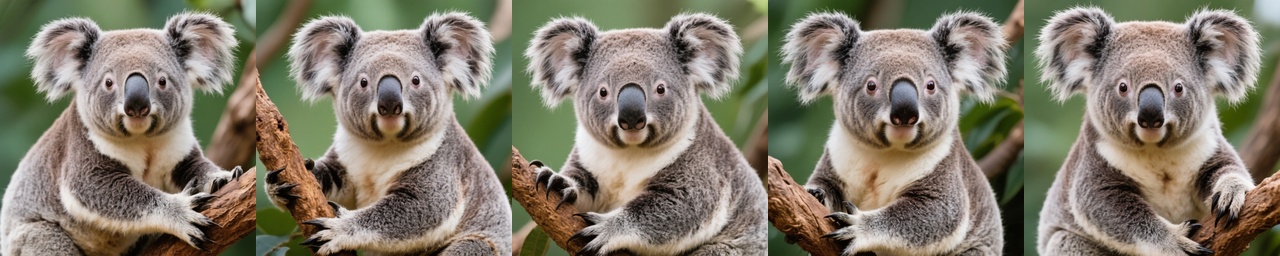} \\
\hline

\hline
\multicolumn{2}{|c|}{\textbf{Broom}} \\
\hline
ImageNet & \includegraphics[height=3.0cm]{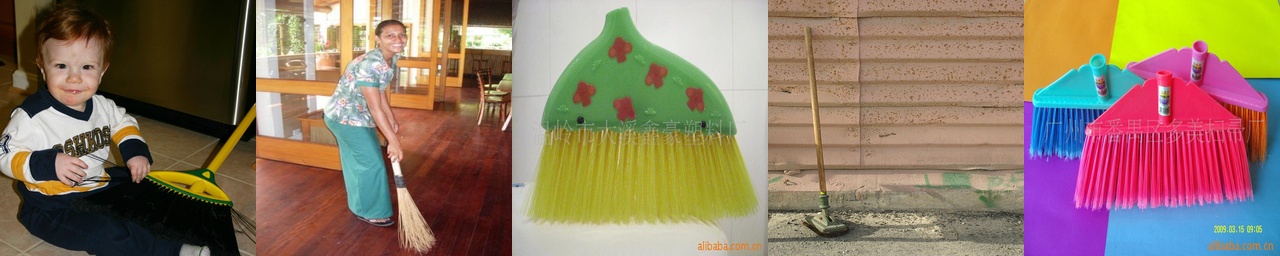} \\
\hline
sd15 & \includegraphics[height=3.0cm]{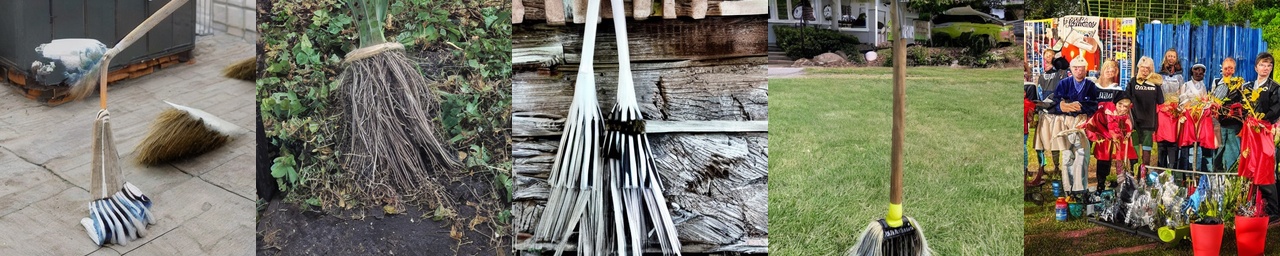} \\
\hline
sd21 & \includegraphics[height=3.0cm]{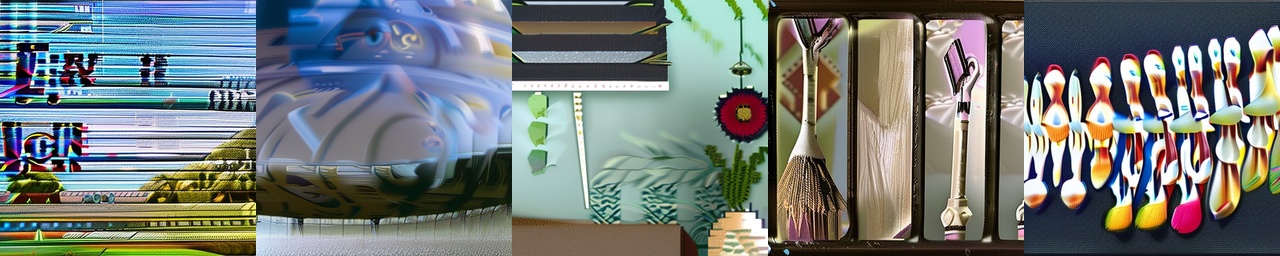} \\
\hline
sdxl & \includegraphics[height=3.0cm]{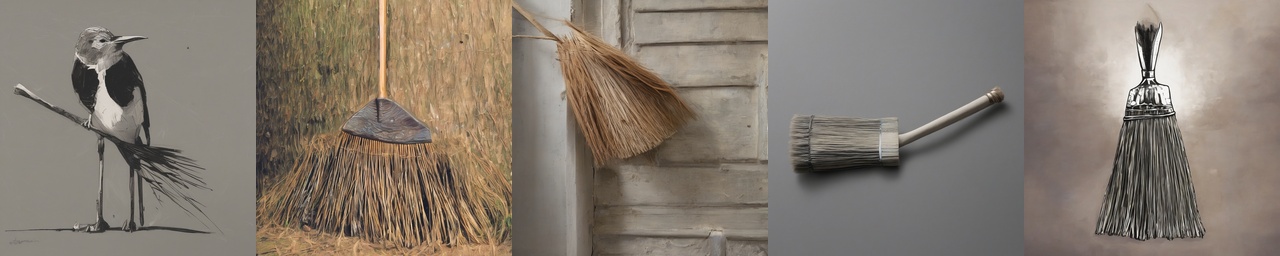} \\
\hline
pixelart & \includegraphics[height=3.0cm]{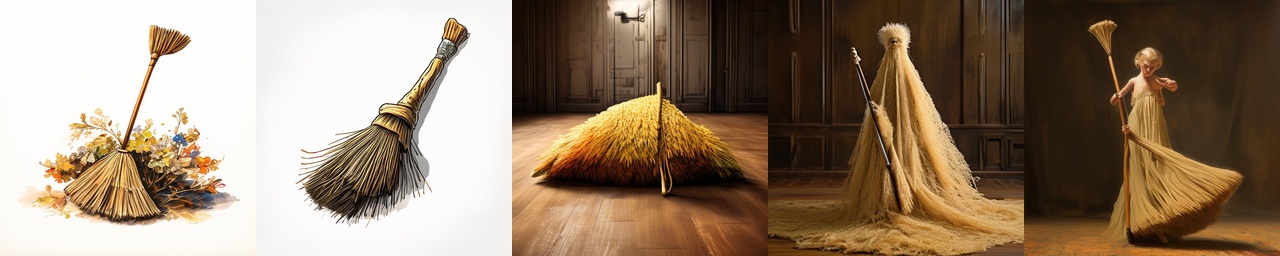} \\
\hline
sdxl-turbo & \includegraphics[height=3.0cm]{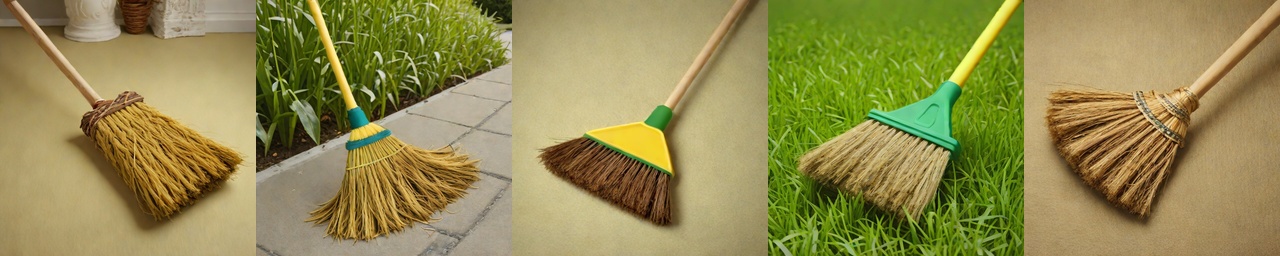} \\
\hline
sd30 & \includegraphics[height=3.0cm]{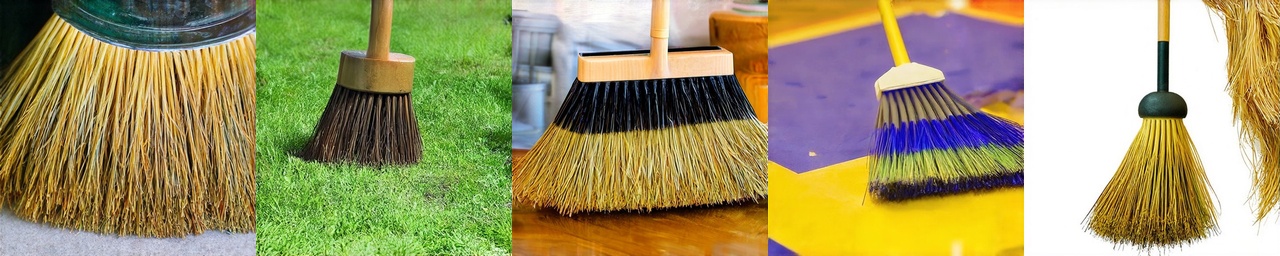} \\
\hline
flux-dev & \includegraphics[height=3.0cm]{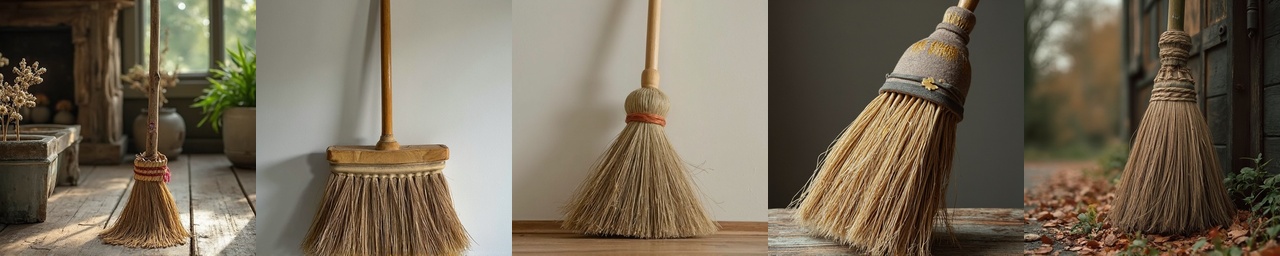} \\
\hline
flux-schnell & \includegraphics[height=3.0cm]{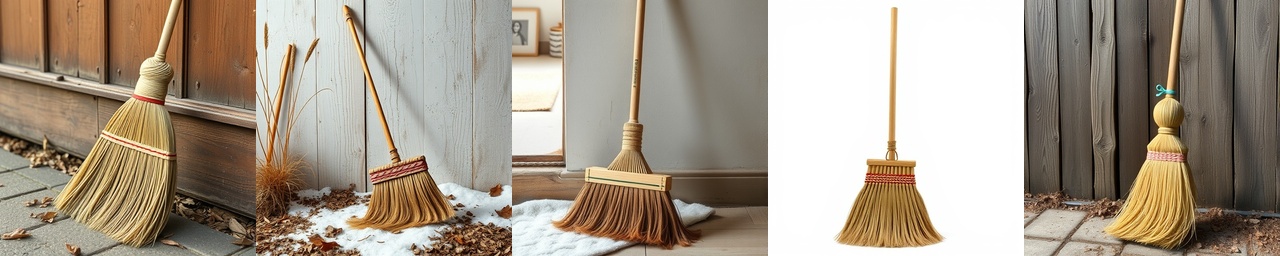} \\
\hline
sd35 & \includegraphics[height=3.0cm]{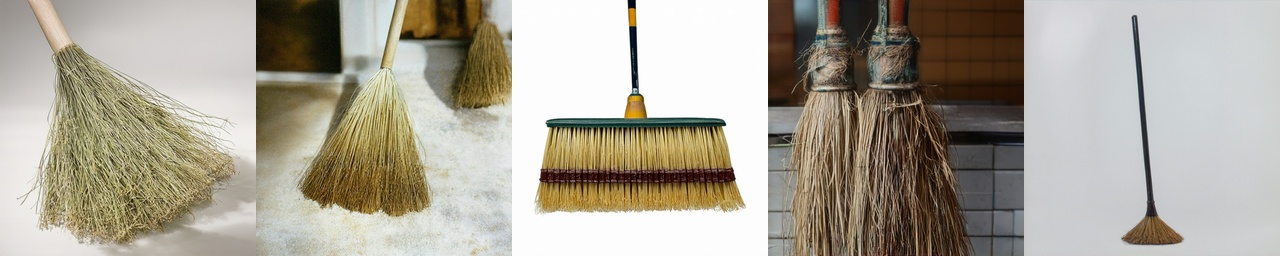} \\
\hline
sd35-large & \includegraphics[height=3.0cm]{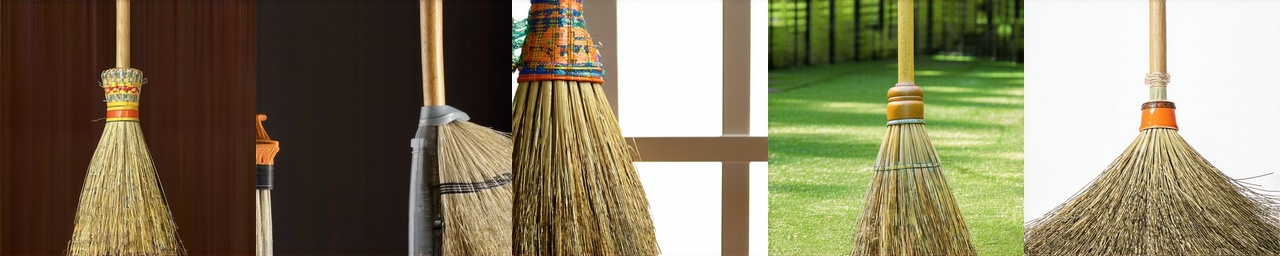} \\
\hline
sd35-turbo & \includegraphics[height=3.0cm]{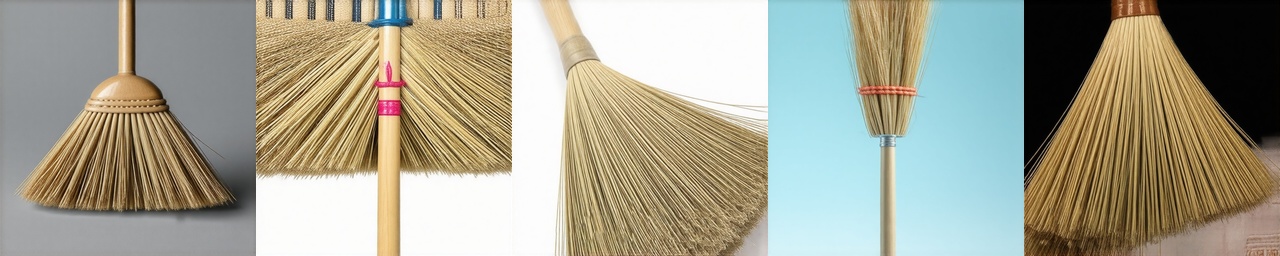} \\
\hline
sana & \includegraphics[height=3.0cm]{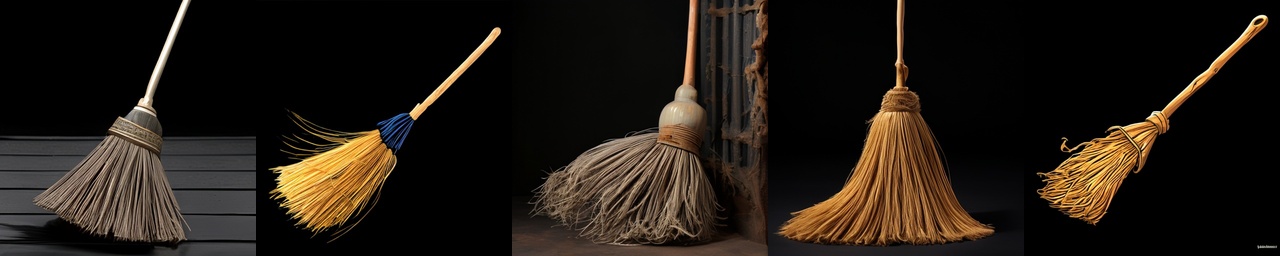} \\
\hline
lumina2 & \includegraphics[height=3.0cm]{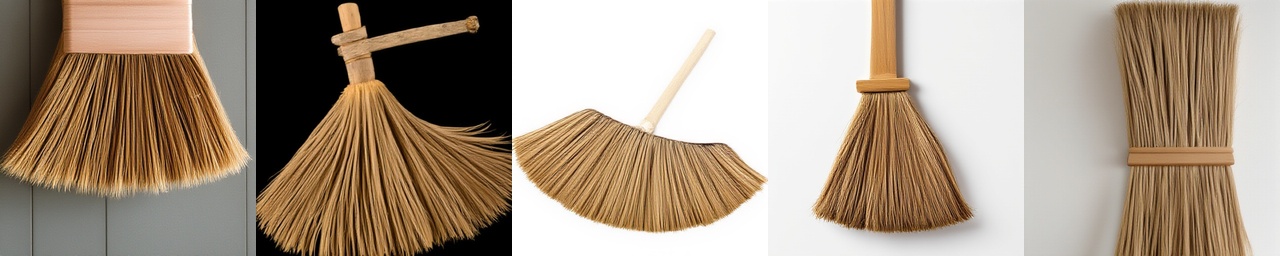} \\
\hline
qwen & \includegraphics[height=3.0cm]{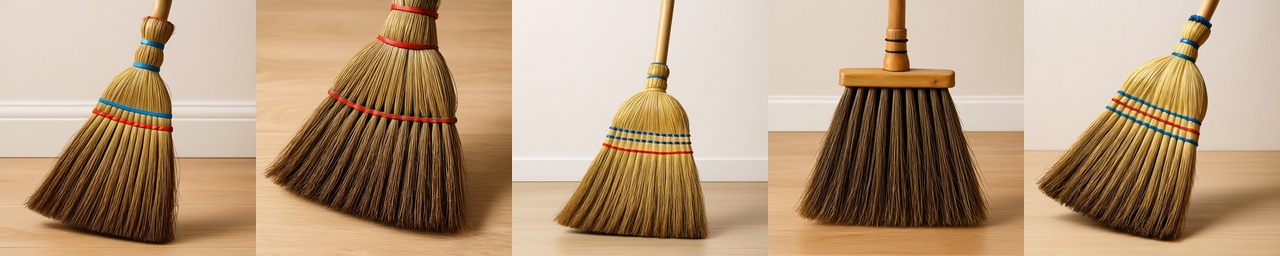} \\
\hline

\end{longtable}

\begin{longtable}{|l|c|c|}

\caption{We visualize samples generated from images with class name and caption prompts. We see that using detailed captions significantly improves the diversity and realism of the synthetic data, matching the observed boost in model performance.}\\
\hline
\textbf{Model} & \textbf{Class Name Prompt} & \textbf{Caption Prompt} \\
\hline
\endfirsthead

\hline
\multicolumn{3}{|c|}%
{\tablename\ \thetable\ -- \textit{Continued from previous page}} \\
\hline
\textbf{Model} & \textbf{Class Name Prompt} & \textbf{Caption Prompt} \\
\hline
\endhead

\hline
\multicolumn{3}{|r|}{\textit{Continued on next page}} \\
\hline
\endfoot

\hline
\endlastfoot

\hline
\multicolumn{3}{|c|}{\textbf{goldfish}} \\
\hline
sd15 & \includegraphics[height=1.87cm]{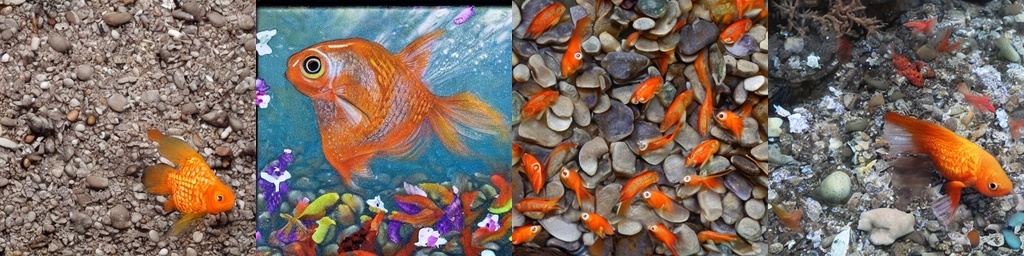} & \includegraphics[height=1.87cm]{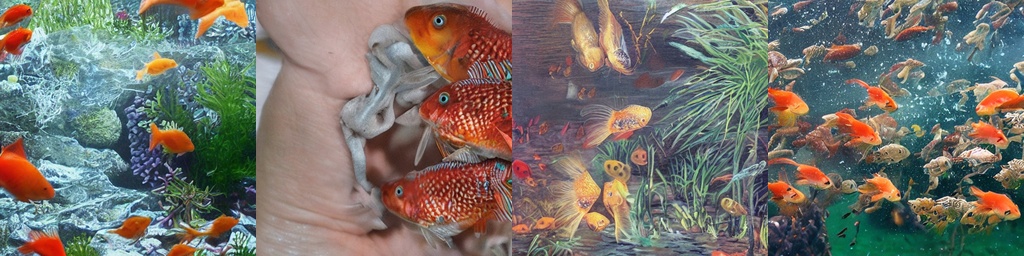} \\
\hline
sd21 & \includegraphics[height=1.87cm]{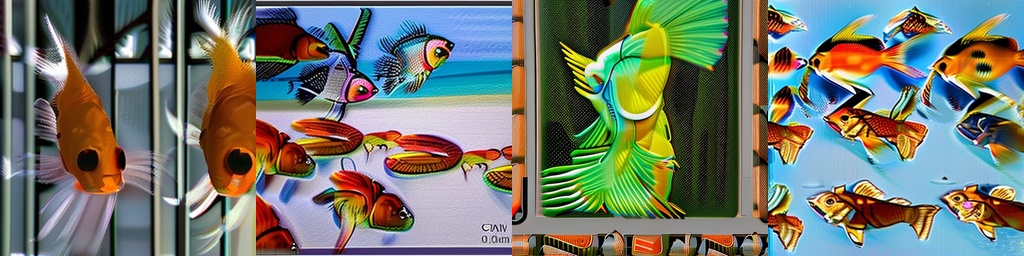} & \includegraphics[height=1.87cm]{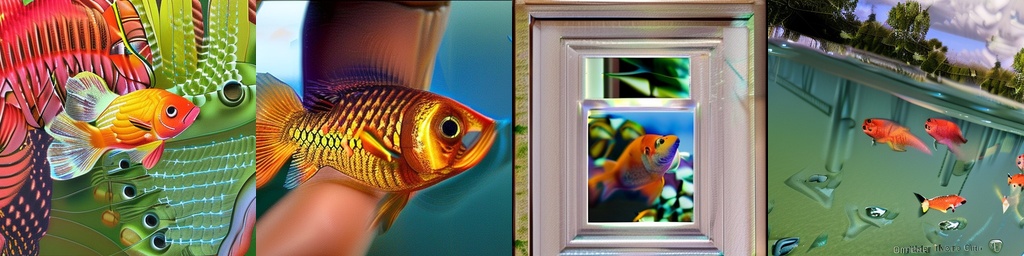} \\
\hline
sdxl & \includegraphics[height=1.87cm]{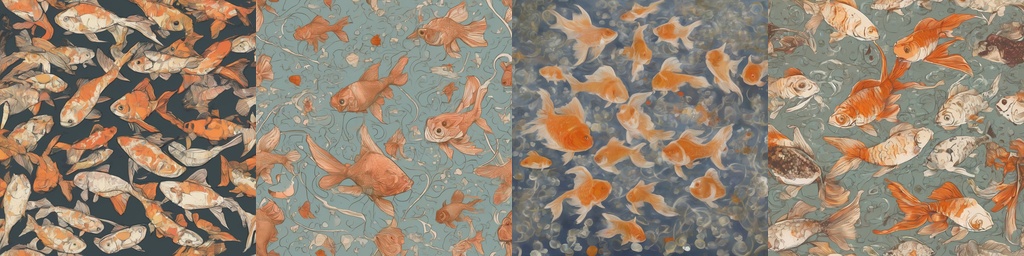} & \includegraphics[height=1.87cm]{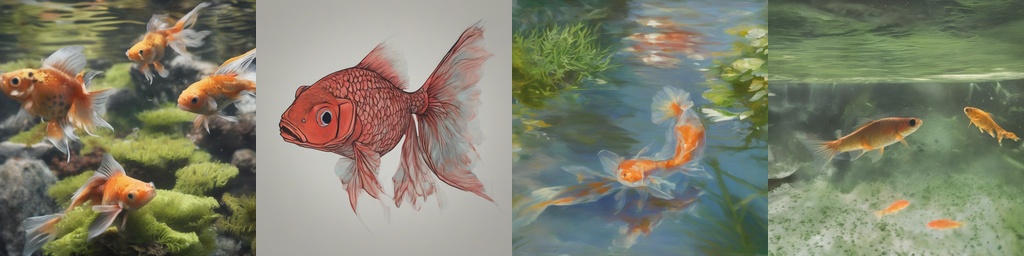} \\
\hline
pixelart & \includegraphics[height=1.87cm]{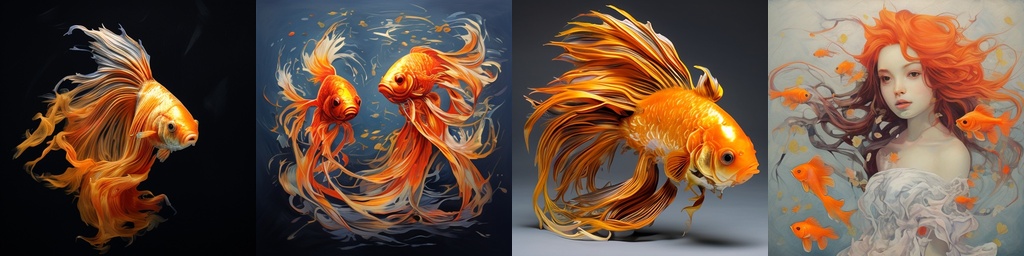} & \includegraphics[height=1.87cm]{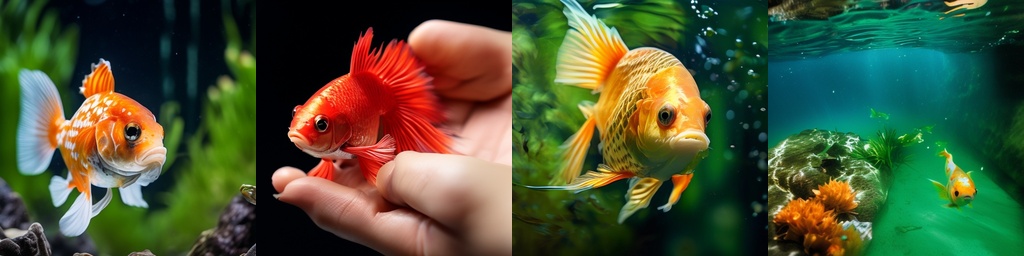} \\
\hline
sdxl-turbo & \includegraphics[height=1.87cm]{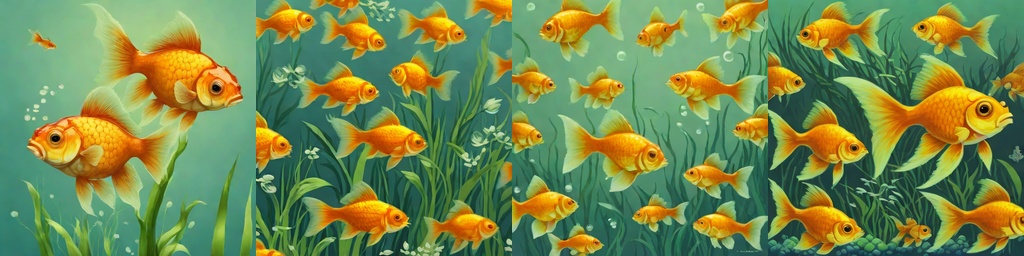} & \includegraphics[height=1.87cm]{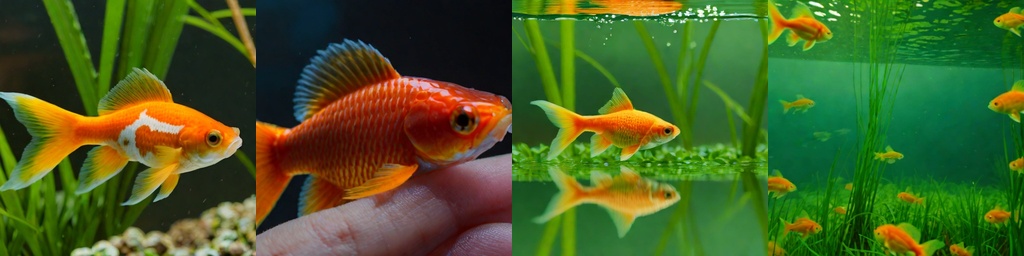} \\
\hline
sd30 & \includegraphics[height=1.87cm]{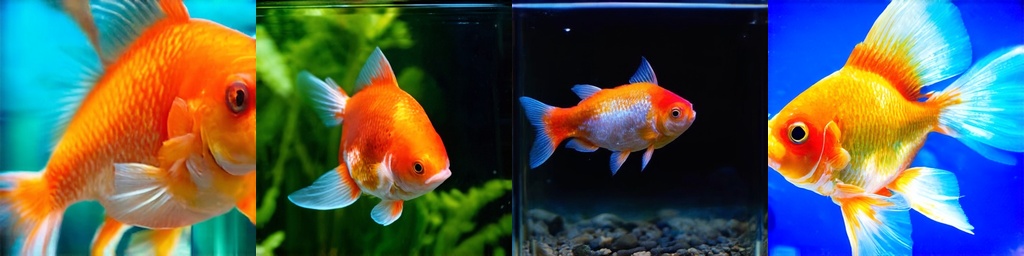} & \includegraphics[height=1.87cm]{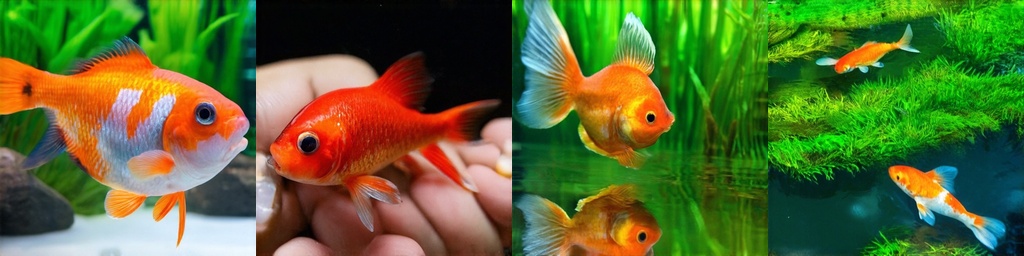} \\
\hline
flux dev & \includegraphics[height=1.87cm]{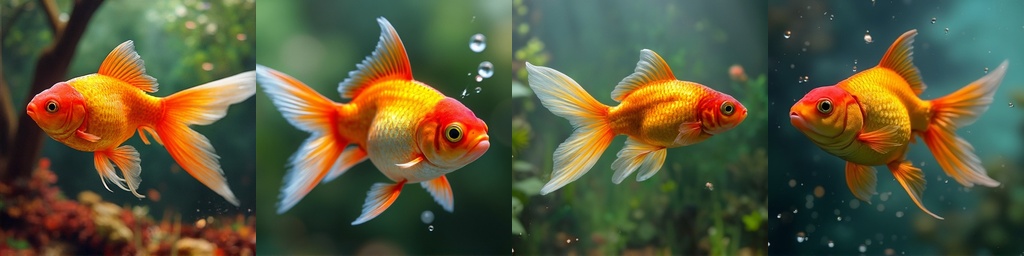} & \includegraphics[height=1.87cm]{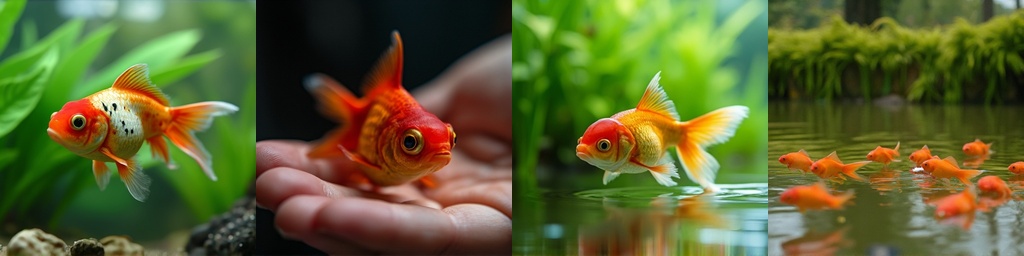} \\
\hline
flux-schnell & \includegraphics[height=1.87cm]{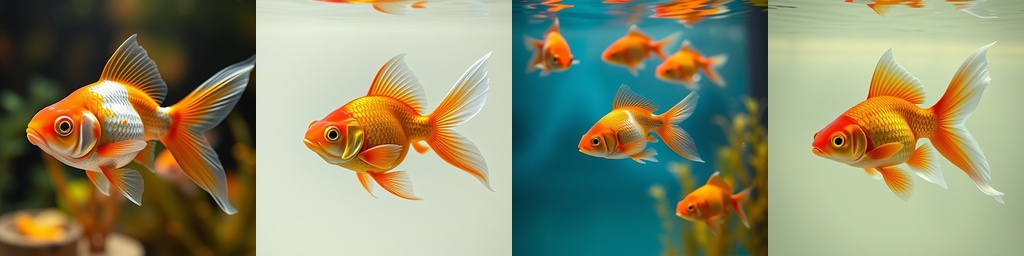} & \includegraphics[height=1.87cm]{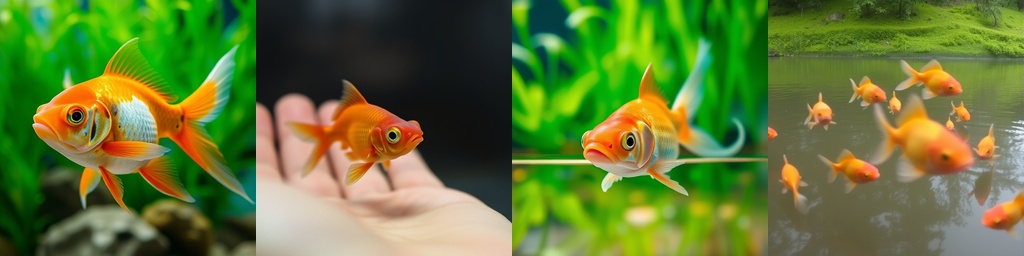} \\
\hline
sd35 & \includegraphics[height=1.87cm]{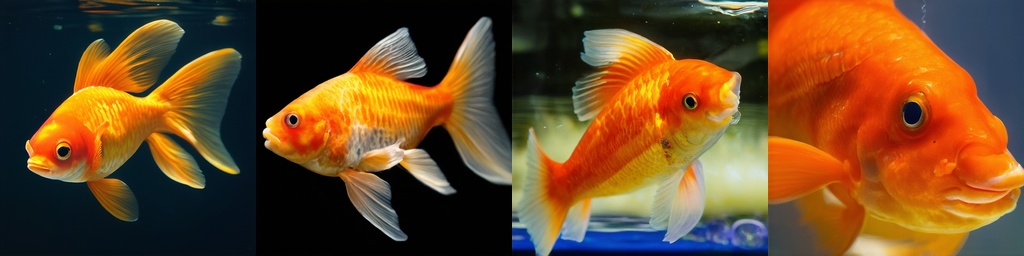} & \includegraphics[height=1.87cm]{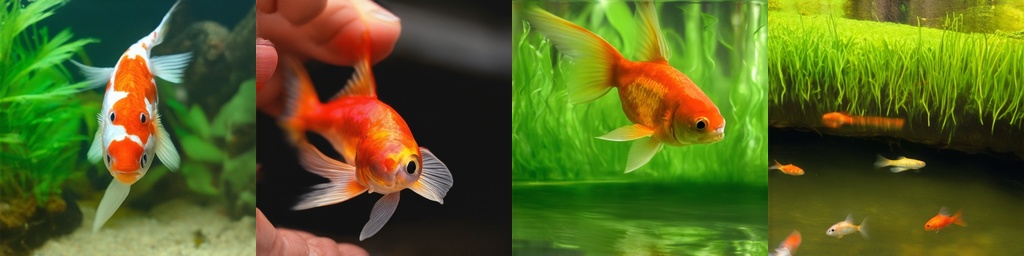} \\
\hline
sd35-large & \includegraphics[height=1.87cm]{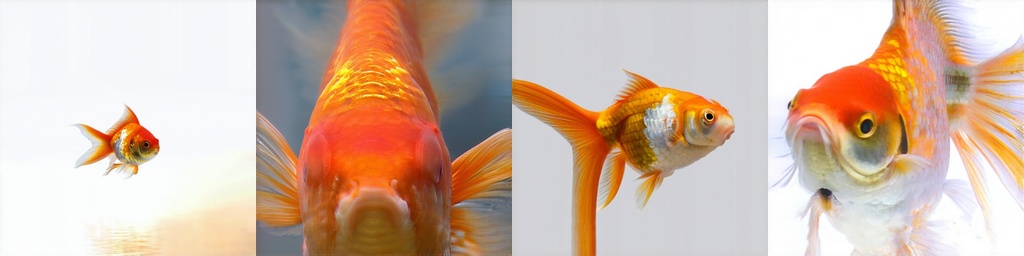} & \includegraphics[height=1.87cm]{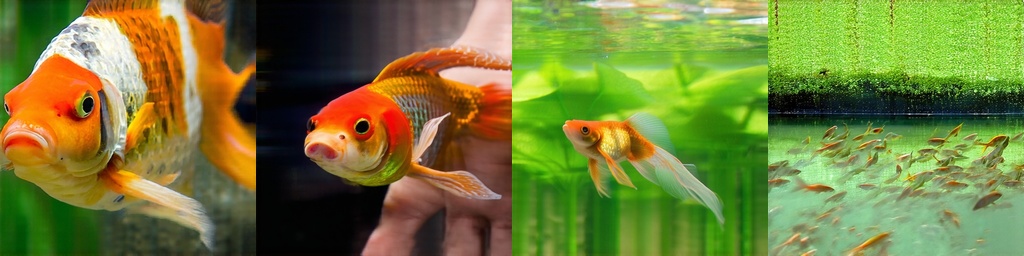} \\
\hline
sd35-turbo & \includegraphics[height=1.87cm]{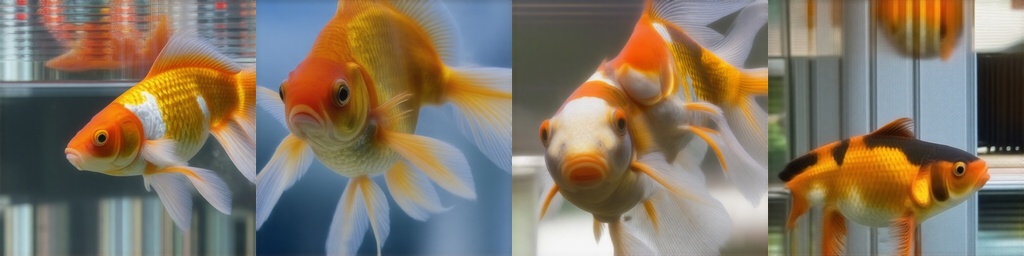} & \includegraphics[height=1.87cm]{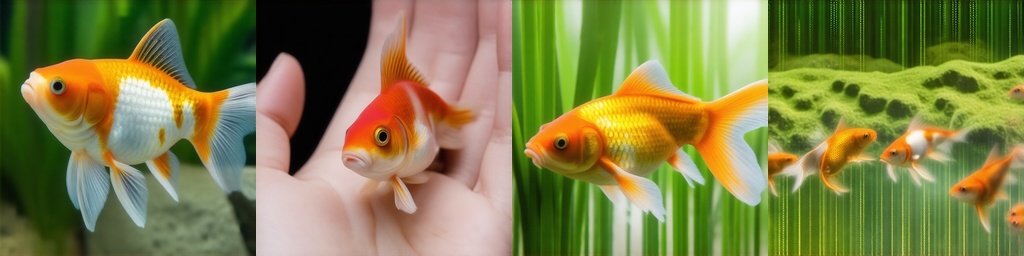} \\
\hline
sana & \includegraphics[height=1.87cm]{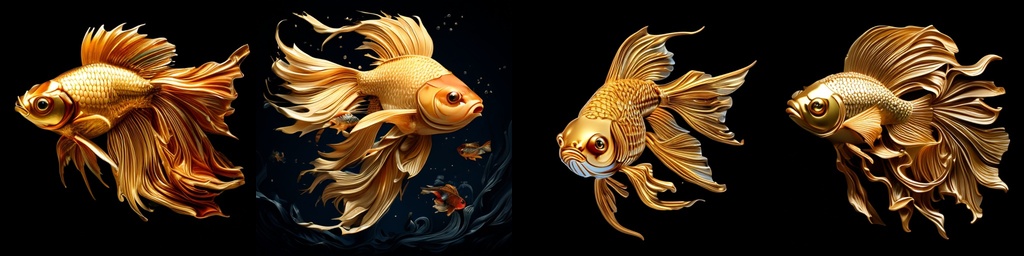} & \includegraphics[height=1.87cm]{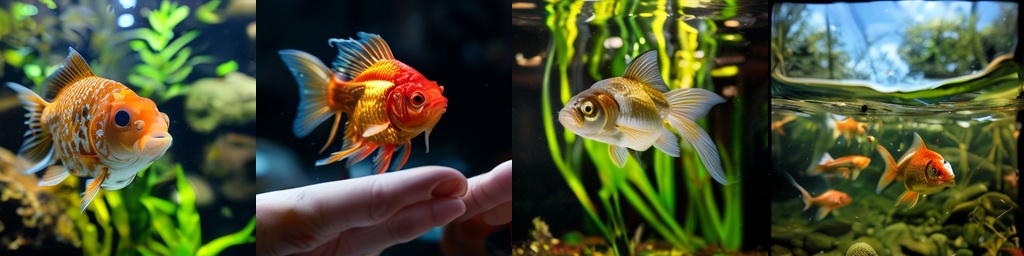} \\
\hline
qwen & \includegraphics[height=1.87cm]{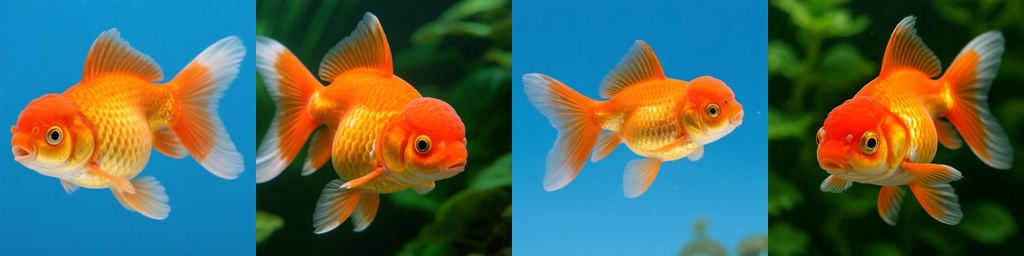} & \includegraphics[height=1.87cm]{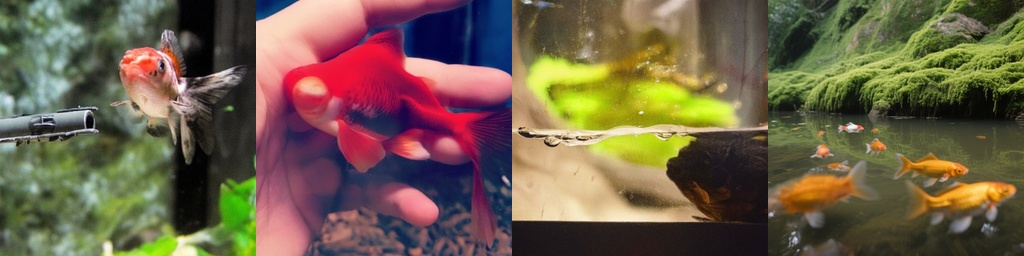} \\
\hline

\hline
\multicolumn{3}{|c|}{\textbf{monarch}} \\
\hline
sd15 & \includegraphics[height=1.87cm]{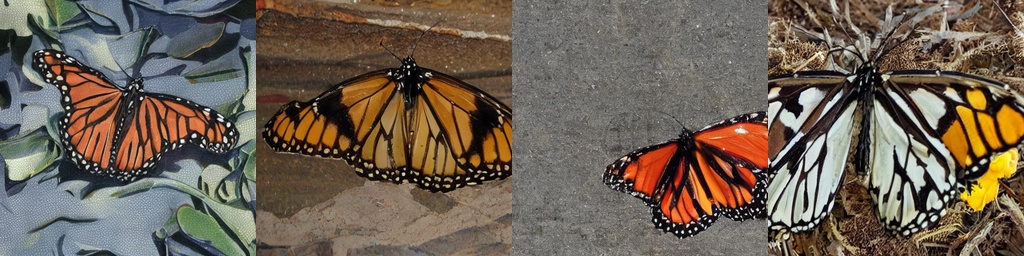} & \includegraphics[height=1.87cm]{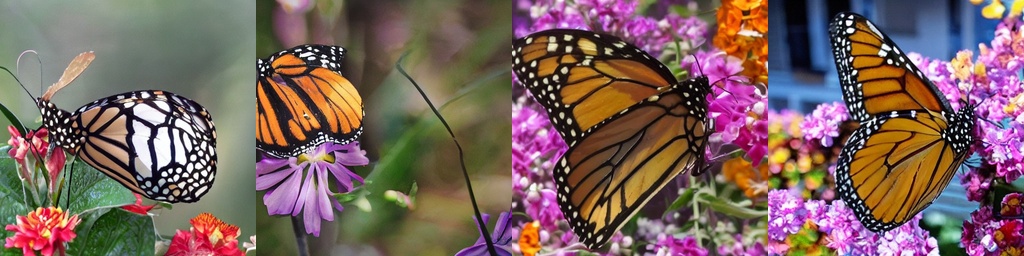} \\
\hline
sd21 & \includegraphics[height=1.87cm]{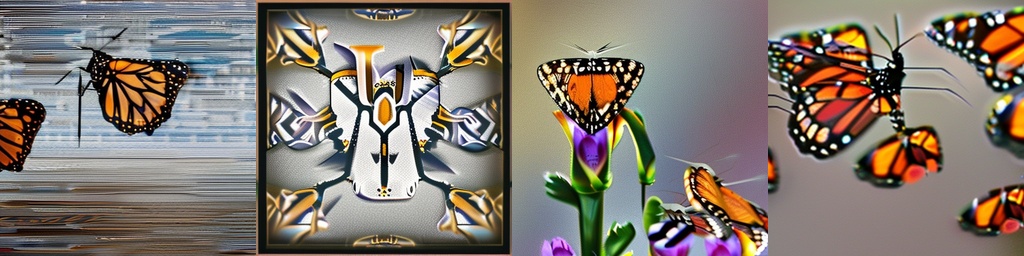} & \includegraphics[height=1.87cm]{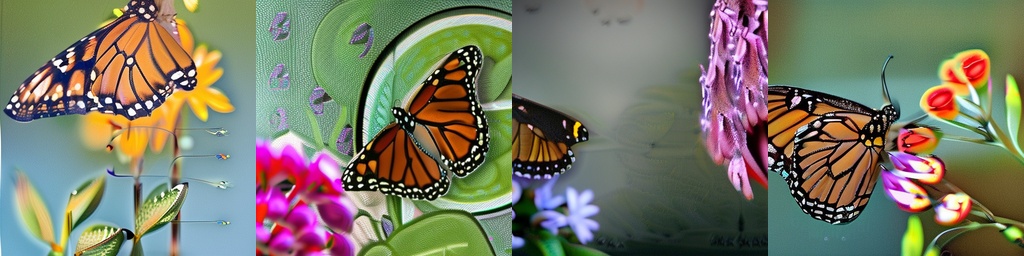} \\
\hline
sdxl & \includegraphics[height=1.87cm]{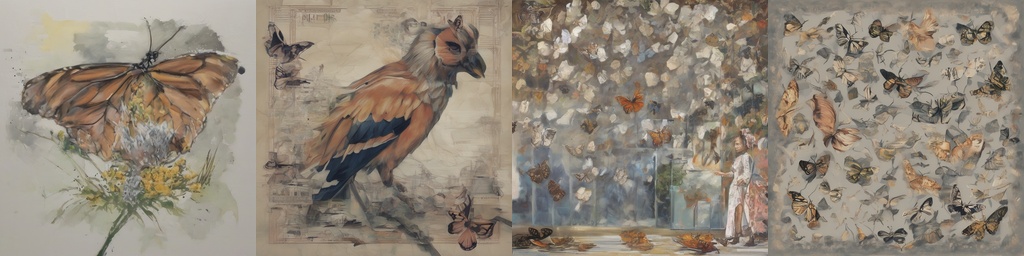} & \includegraphics[height=1.87cm]{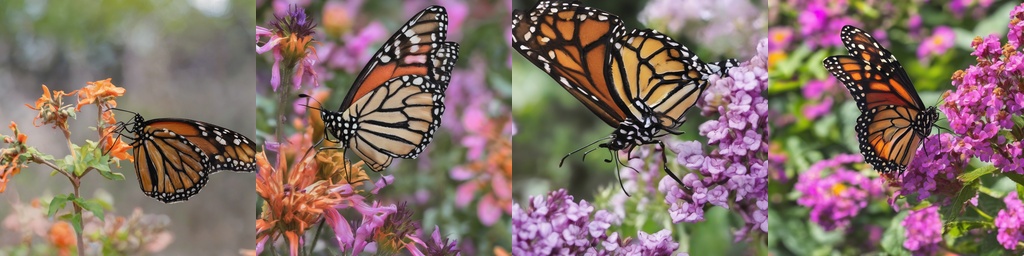} \\
\hline
pixelart & \includegraphics[height=1.87cm]{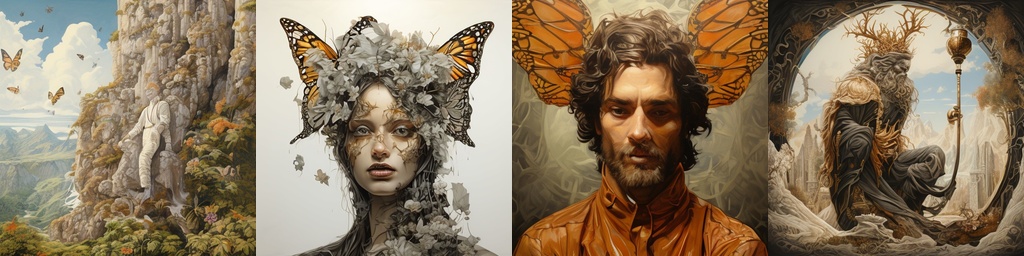} & \includegraphics[height=1.87cm]{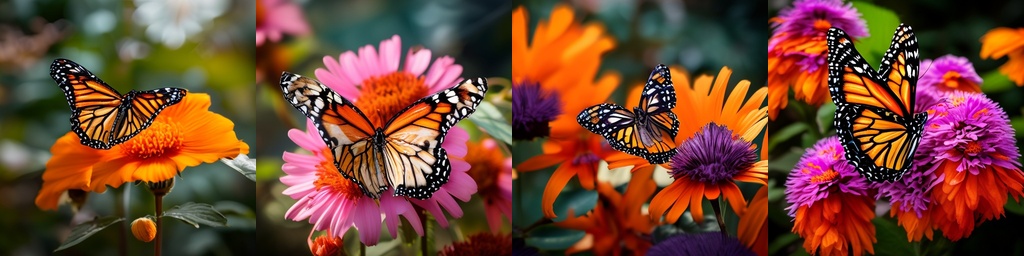} \\
\hline
sdxl-turbo & \includegraphics[height=1.87cm]{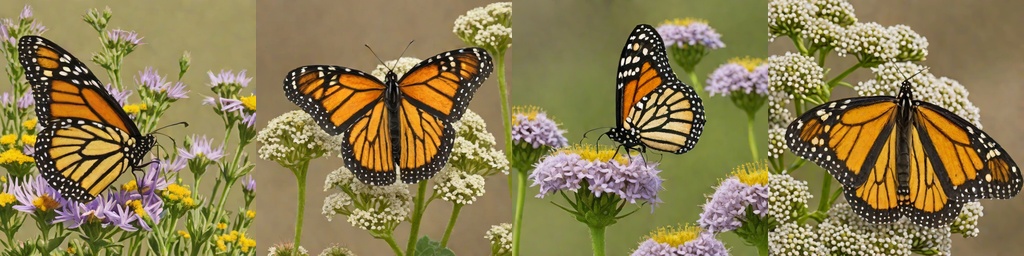} & \includegraphics[height=1.87cm]{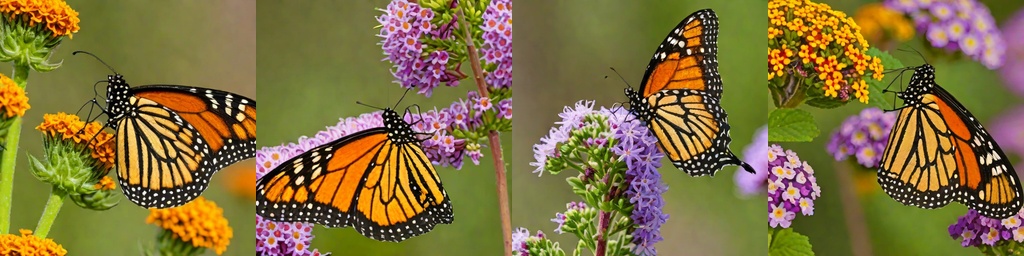} \\
\hline
sd30 & \includegraphics[height=1.87cm]{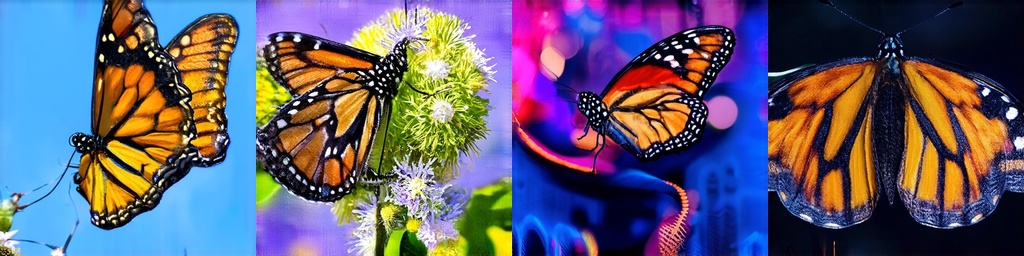} & \includegraphics[height=1.87cm]{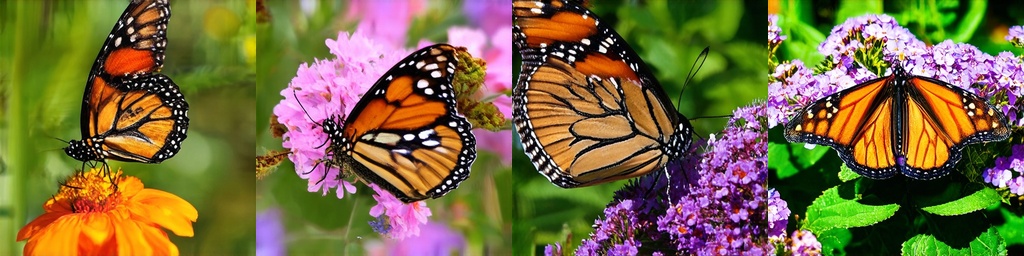} \\
\hline
flux dev & \includegraphics[height=1.87cm]{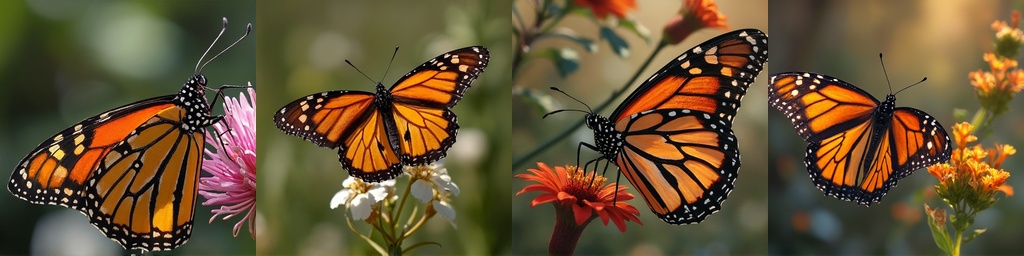} & \includegraphics[height=1.87cm]{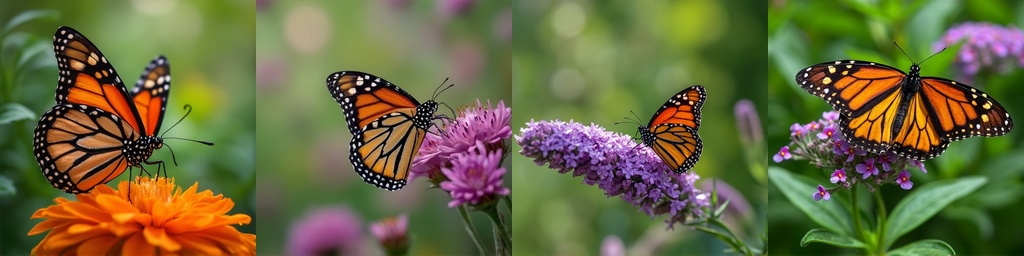} \\
\hline
flux-schnell & \includegraphics[height=1.87cm]{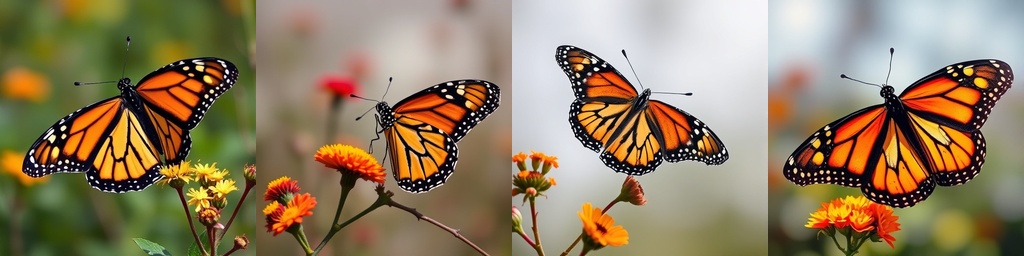} & \includegraphics[height=1.87cm]{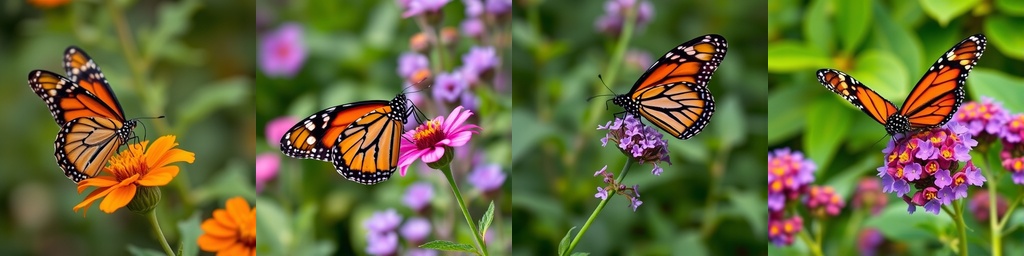} \\
\hline
sd35 & \includegraphics[height=1.87cm]{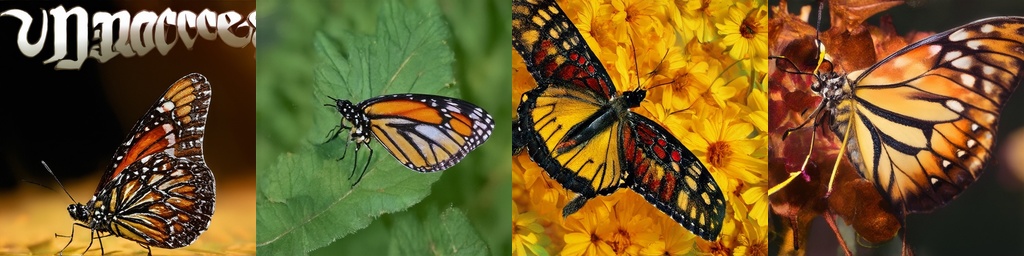} & \includegraphics[height=1.87cm]{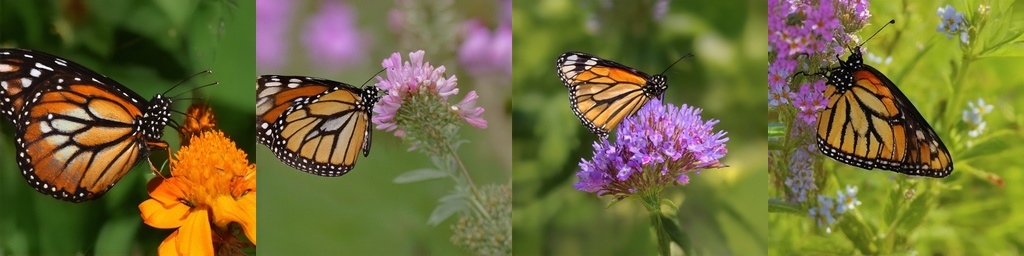} \\
\hline
sd35-large & \includegraphics[height=1.87cm]{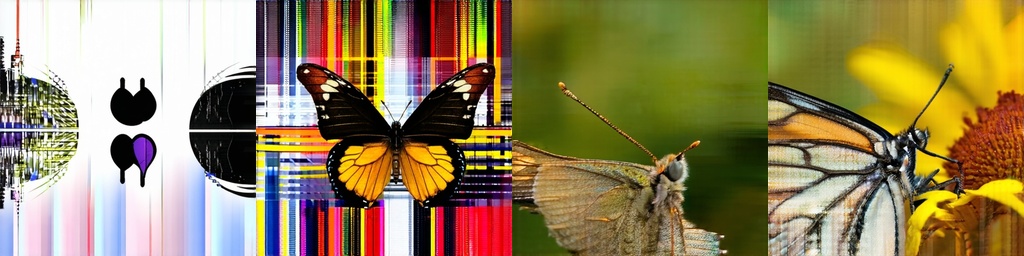} & \includegraphics[height=1.87cm]{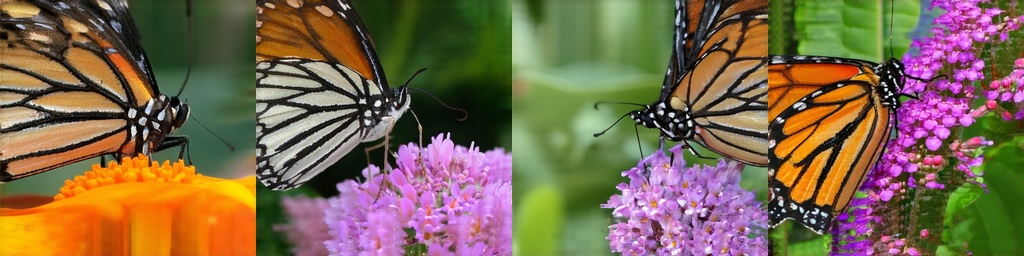} \\
\hline
sd35-turbo & \includegraphics[height=1.87cm]{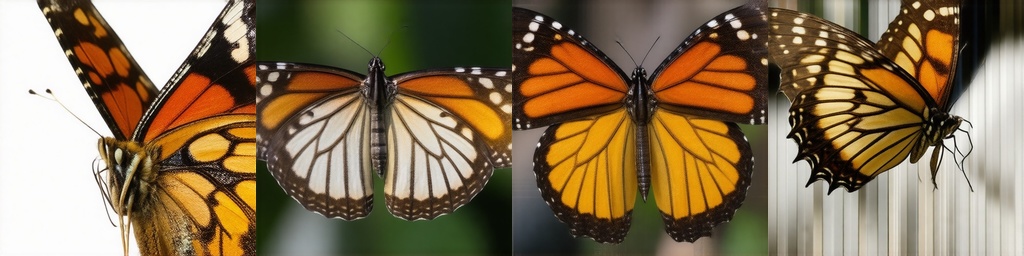} & \includegraphics[height=1.87cm]{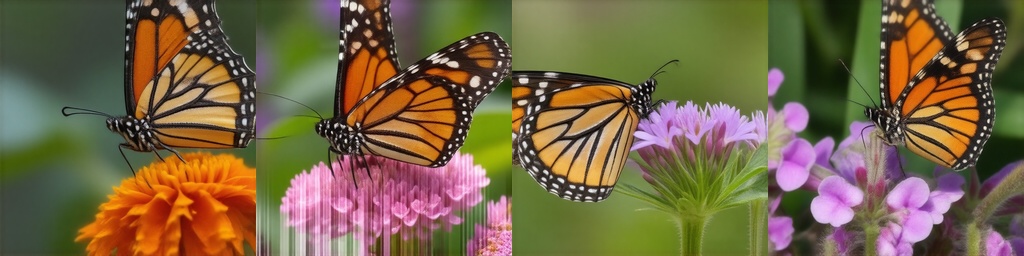} \\
\hline
sana & \includegraphics[height=1.87cm]{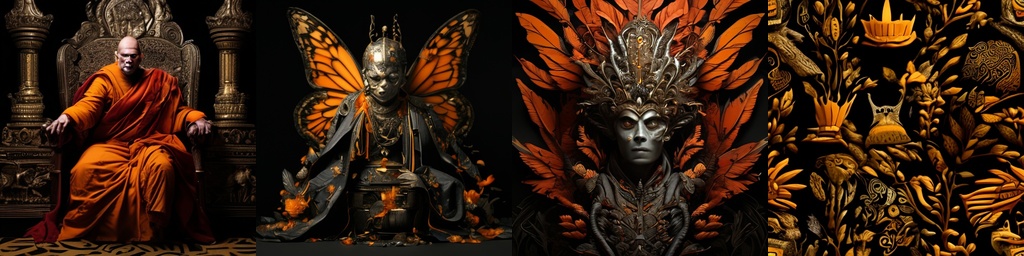} & \includegraphics[height=1.87cm]{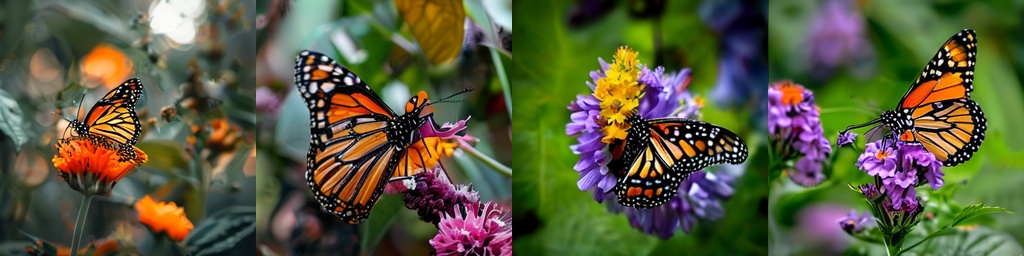} \\
\hline
qwen & \includegraphics[height=1.87cm]{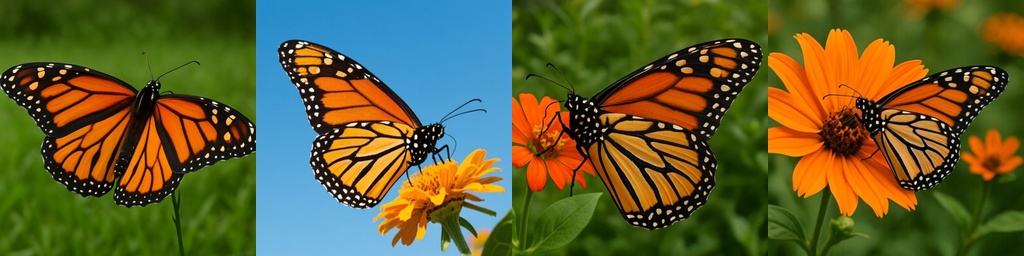} & \includegraphics[height=1.87cm]{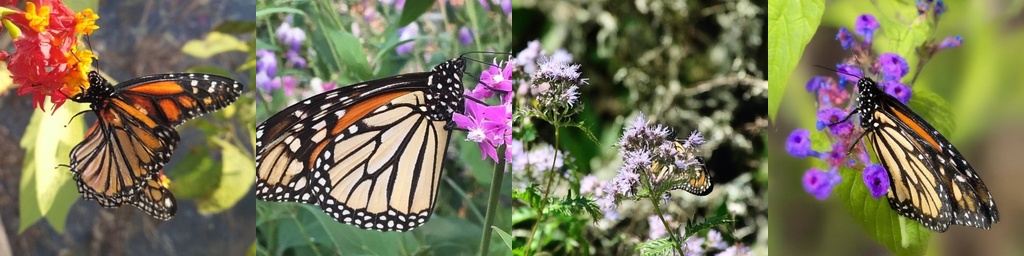} \\
\hline

\hline
\multicolumn{3}{|c|}{\textbf{koala}} \\
\hline
sd15 & \includegraphics[height=1.87cm]{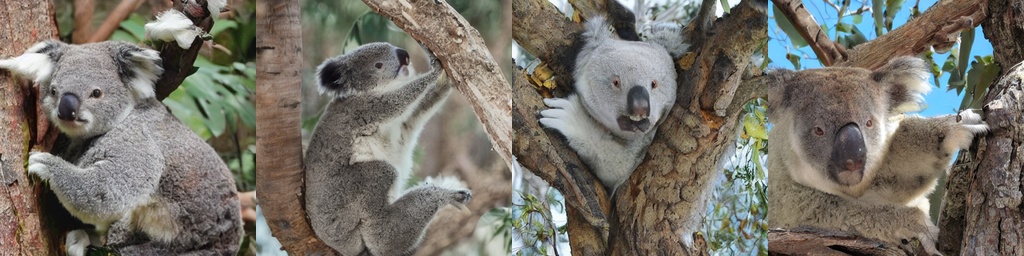} & \includegraphics[height=1.87cm]{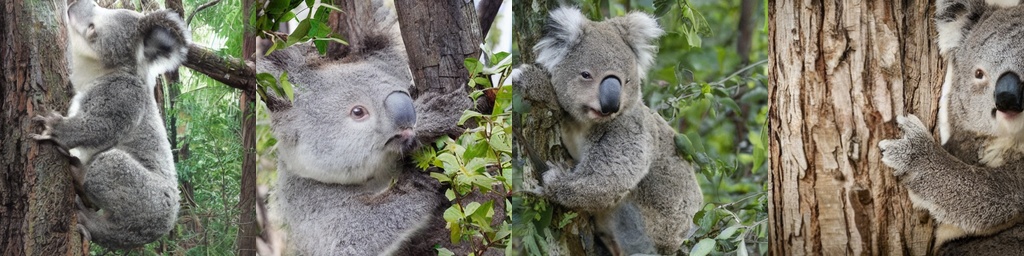} \\
\hline
sd21 & \includegraphics[height=1.87cm]{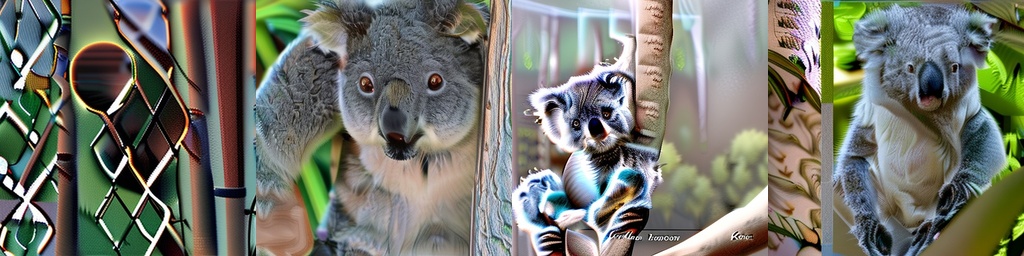} & \includegraphics[height=1.87cm]{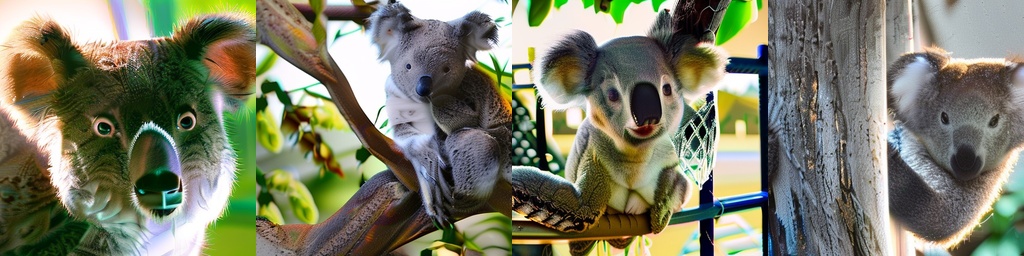} \\
\hline
sdxl & \includegraphics[height=1.87cm]{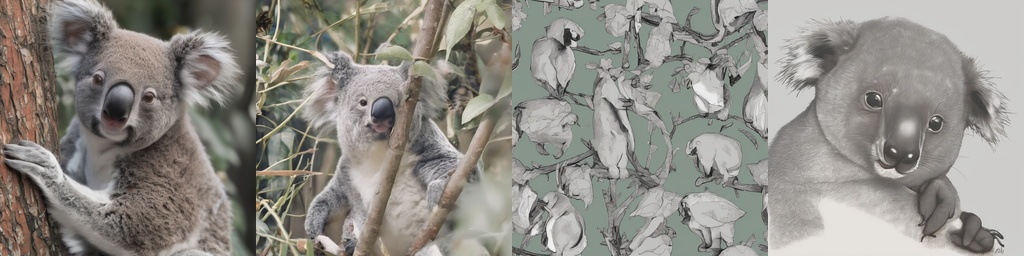} & \includegraphics[height=1.87cm]{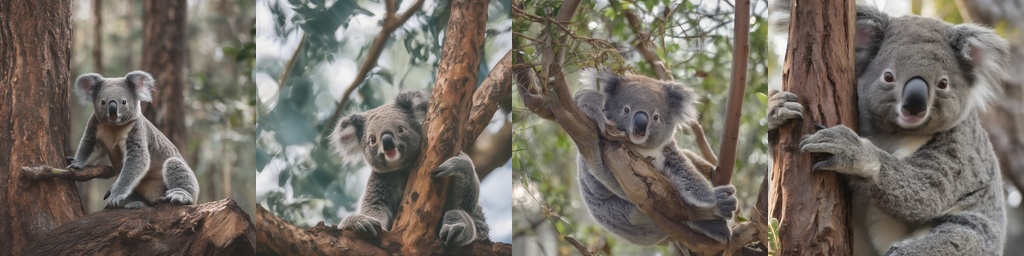} \\
\hline
pixelart & \includegraphics[height=1.87cm]{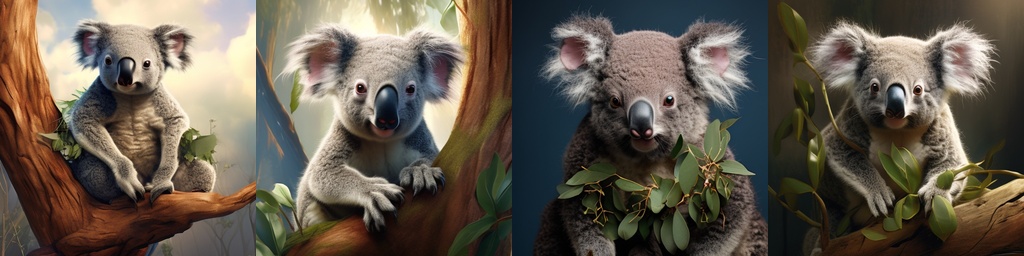} & \includegraphics[height=1.87cm]{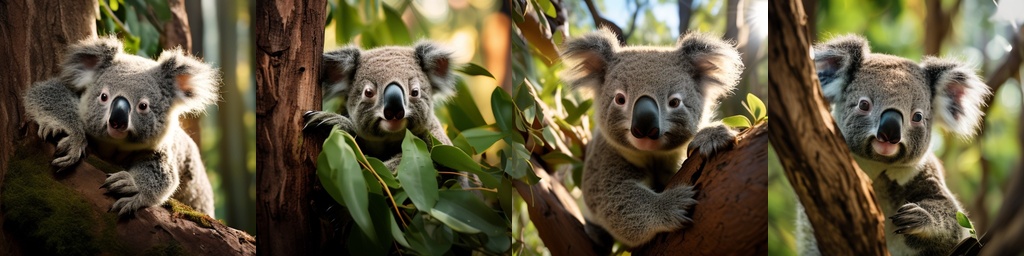} \\
\hline
sdxl-turbo & \includegraphics[height=1.87cm]{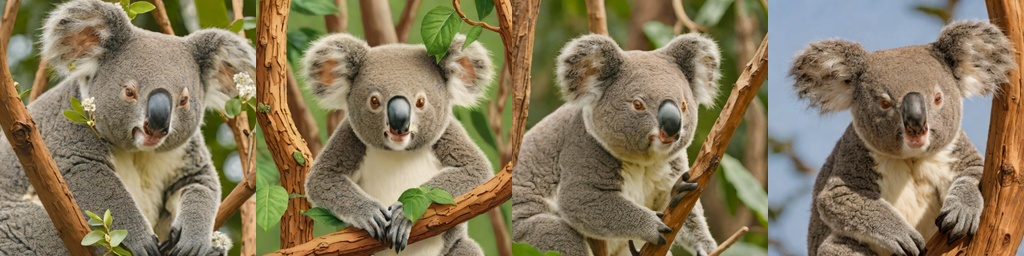} & \includegraphics[height=1.87cm]{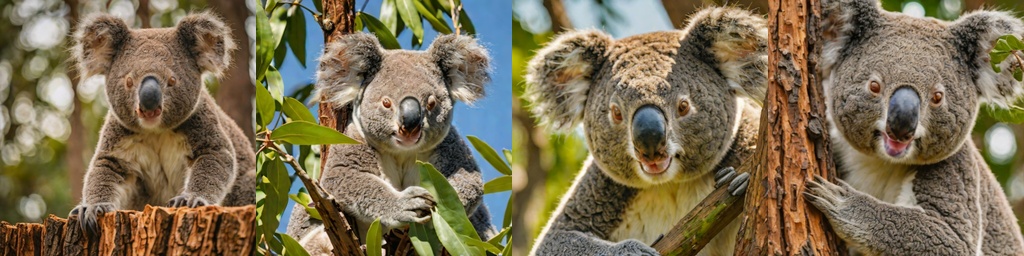} \\
\hline
sd30 & \includegraphics[height=1.87cm]{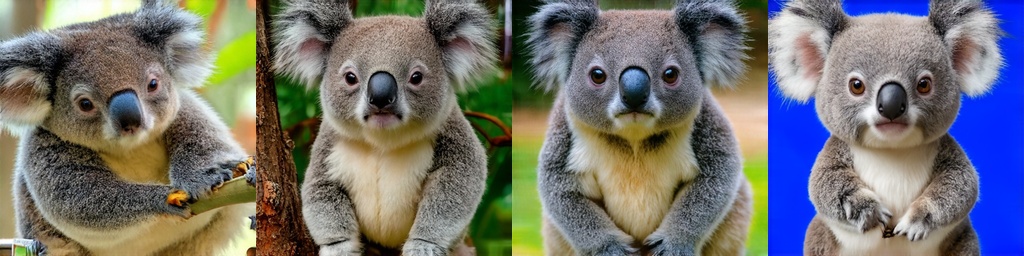} & \includegraphics[height=1.87cm]{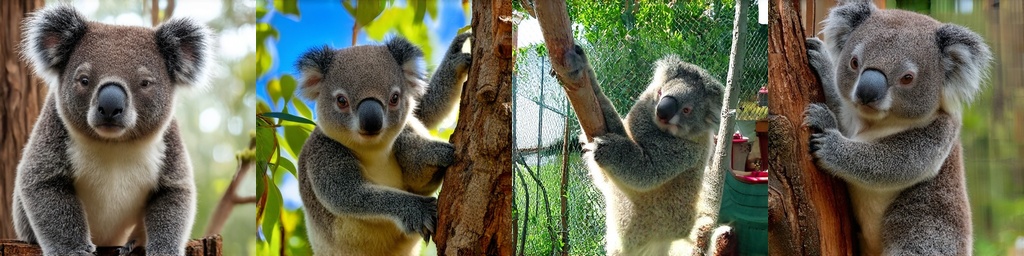} \\
\hline
flux dev & \includegraphics[height=1.87cm]{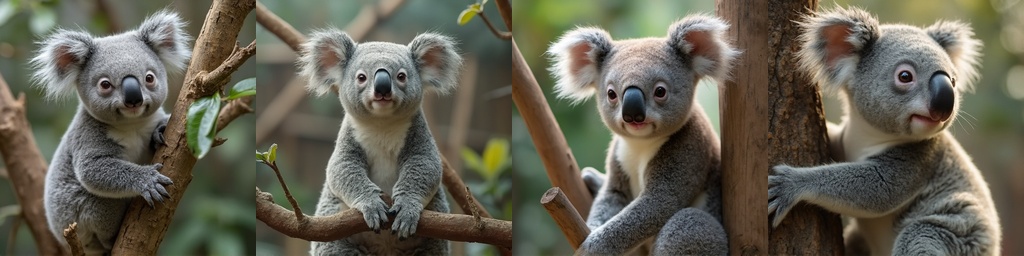} & \includegraphics[height=1.87cm]{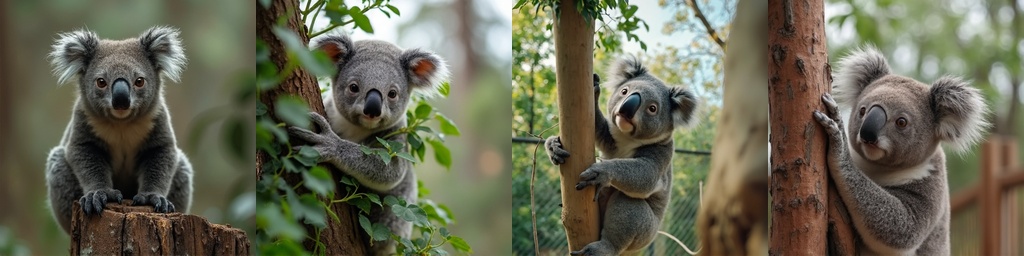} \\
\hline
flux-schnell & \includegraphics[height=1.87cm]{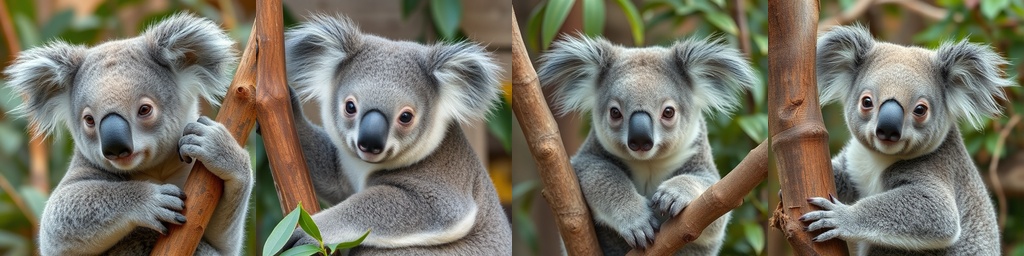} & \includegraphics[height=1.87cm]{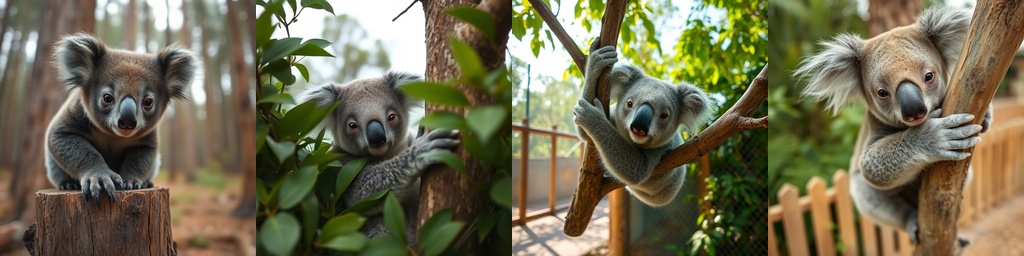} \\
\hline
sd35 & \includegraphics[height=1.87cm]{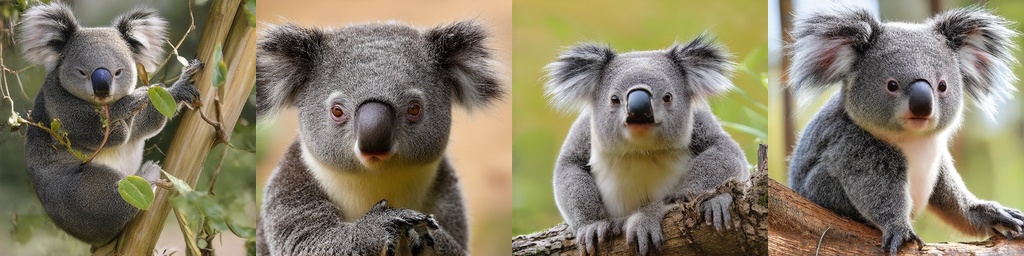} & \includegraphics[height=1.87cm]{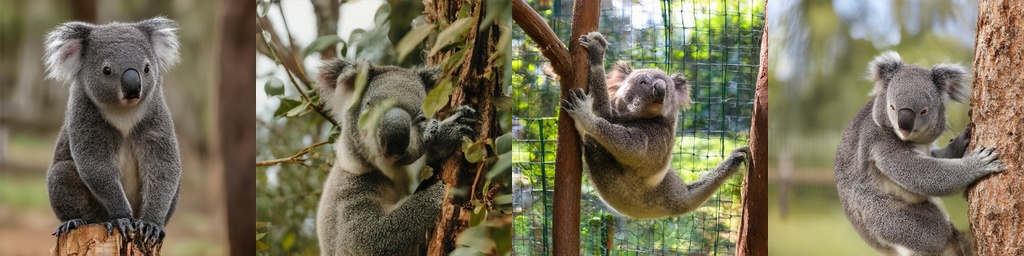} \\
\hline
sd35-large & \includegraphics[height=1.87cm]{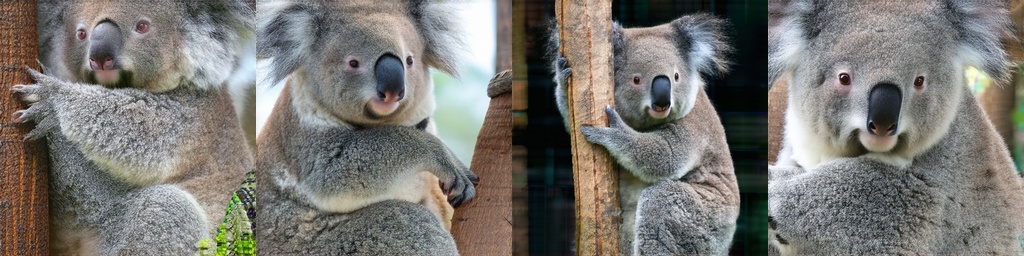} & \includegraphics[height=1.87cm]{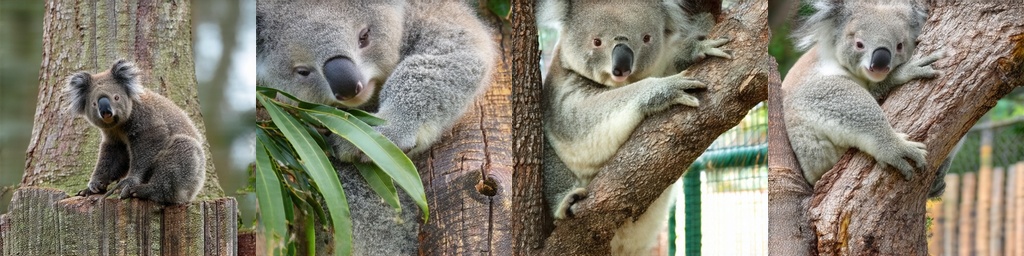} \\
\hline
sd35-turbo & \includegraphics[height=1.87cm]{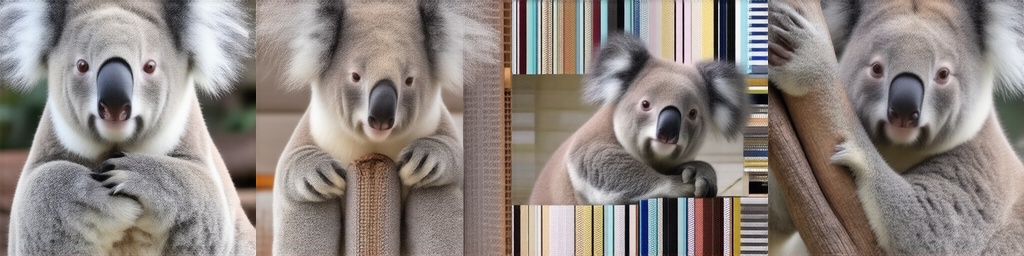} & \includegraphics[height=1.87cm]{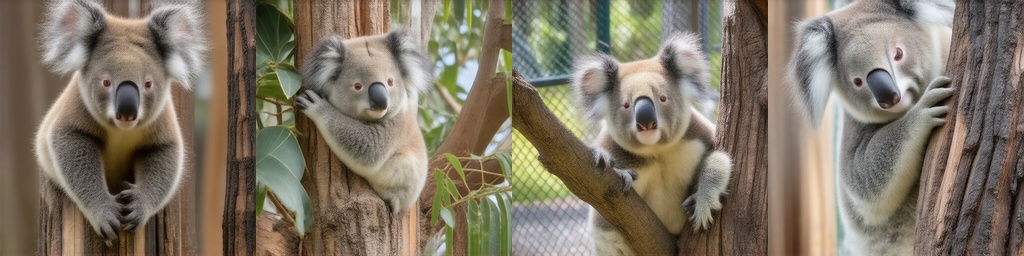} \\
\hline
sana & \includegraphics[height=1.87cm]{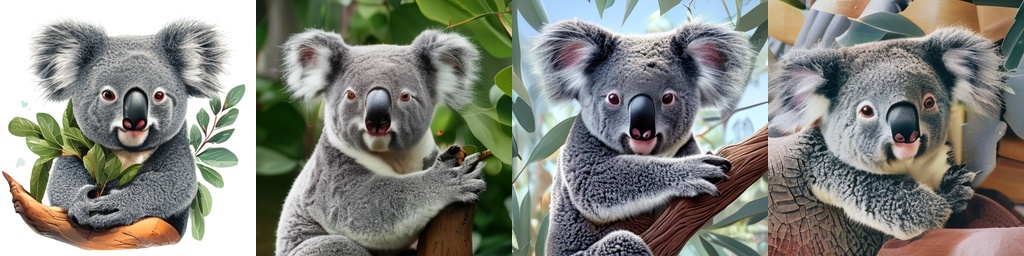} & \includegraphics[height=1.87cm]{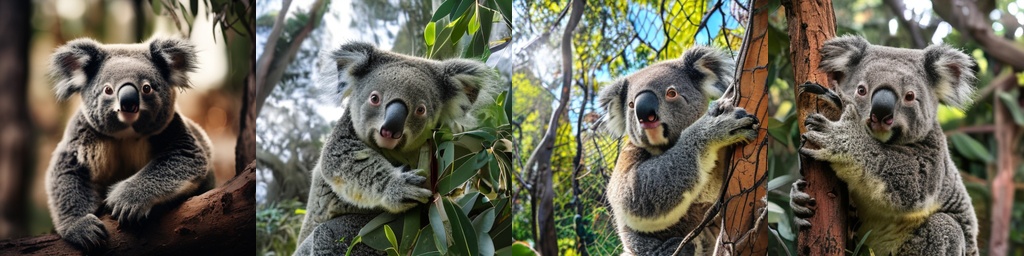} \\
\hline
qwen & \includegraphics[height=1.87cm]{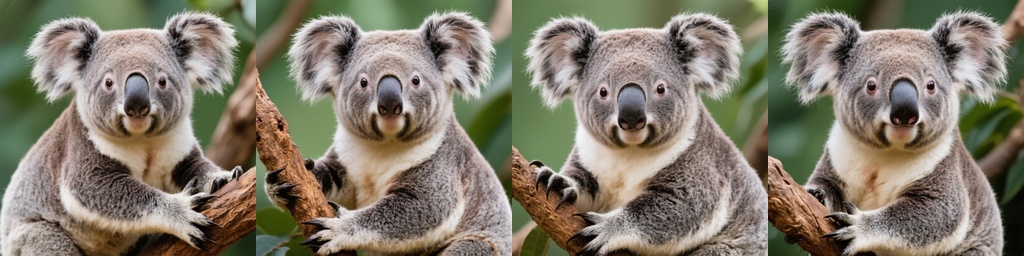} & \includegraphics[height=1.87cm]{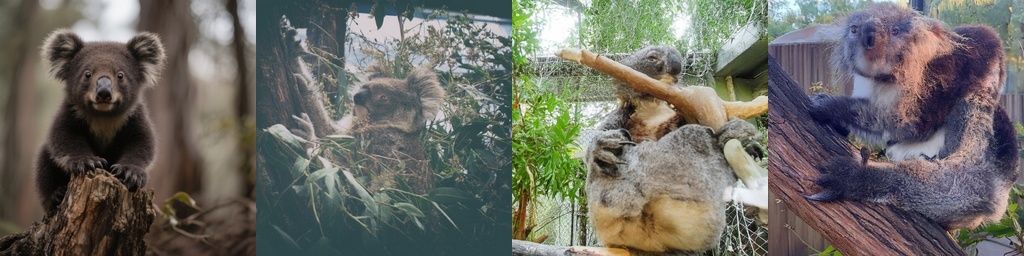} \\
\hline

\hline
\multicolumn{3}{|c|}{\textbf{broom}} \\
\hline
sd15 & \includegraphics[height=1.87cm]{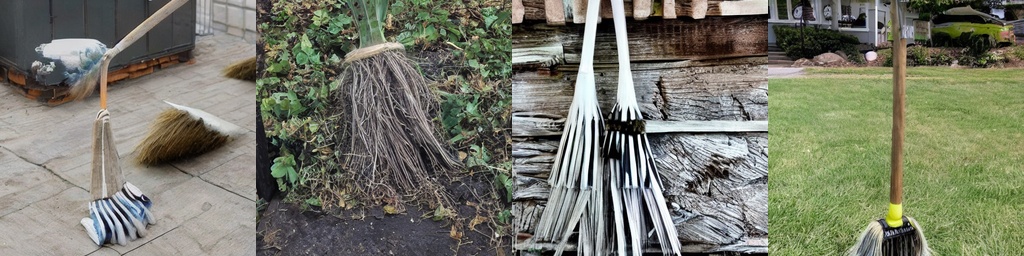} & \includegraphics[height=1.87cm]{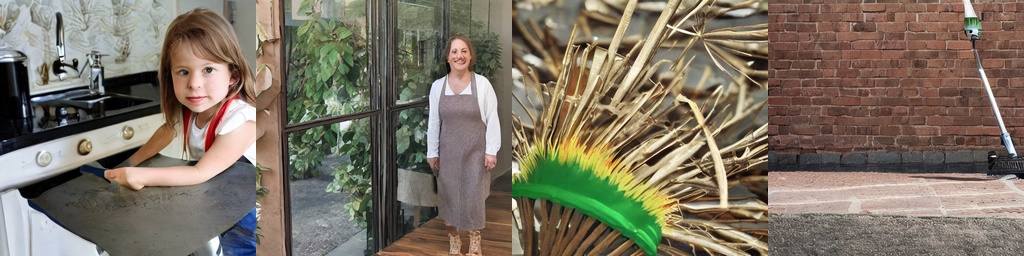} \\
\hline
sd21 & \includegraphics[height=1.87cm]{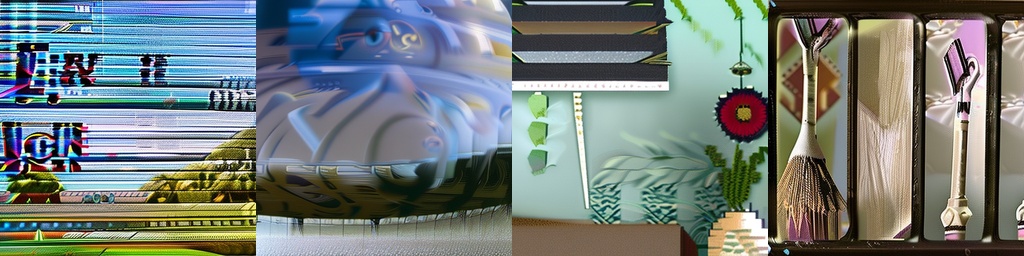} & \includegraphics[height=1.87cm]{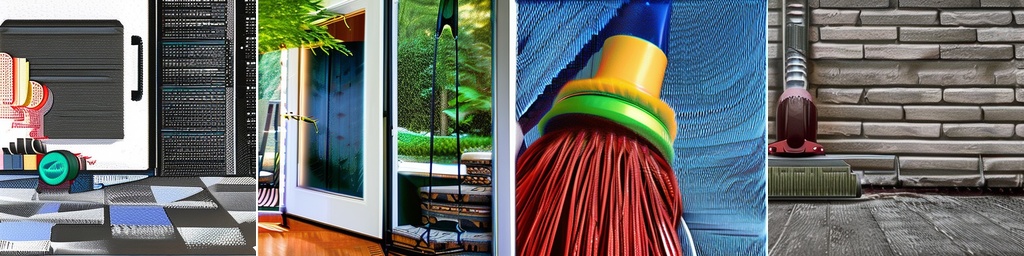} \\
\hline
sdxl & \includegraphics[height=1.87cm]{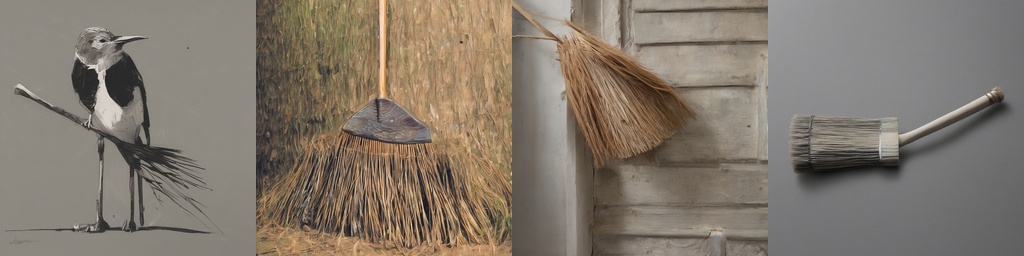} & \includegraphics[height=1.87cm]{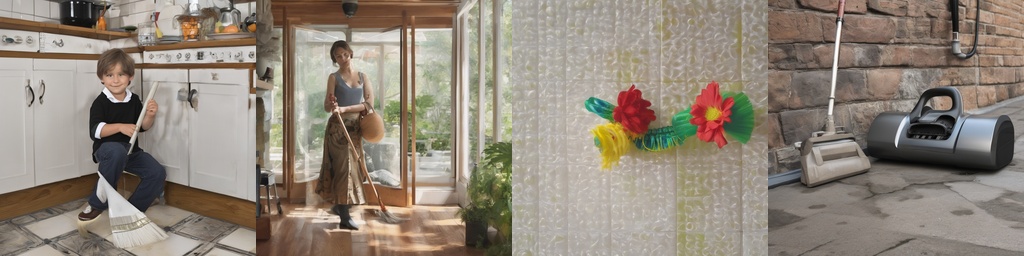} \\
\hline
pixelart & \includegraphics[height=1.87cm]{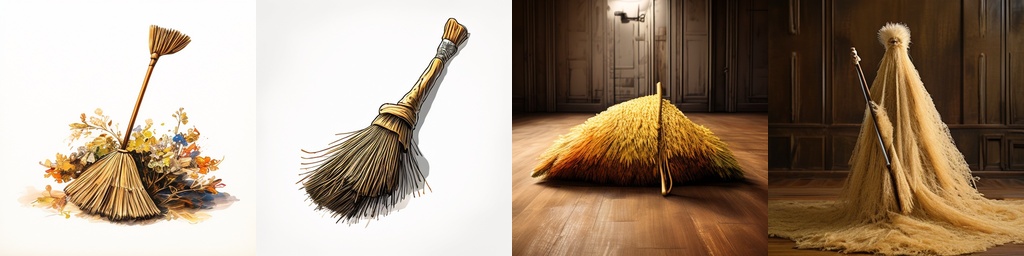} & \includegraphics[height=1.87cm]{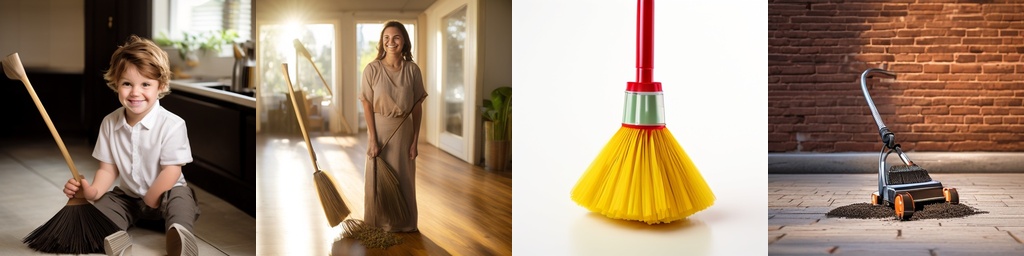} \\
\hline
sdxl-turbo & \includegraphics[height=1.87cm]{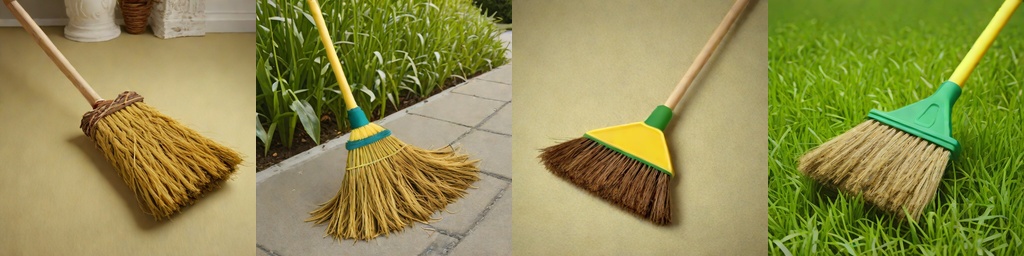} & \includegraphics[height=1.87cm]{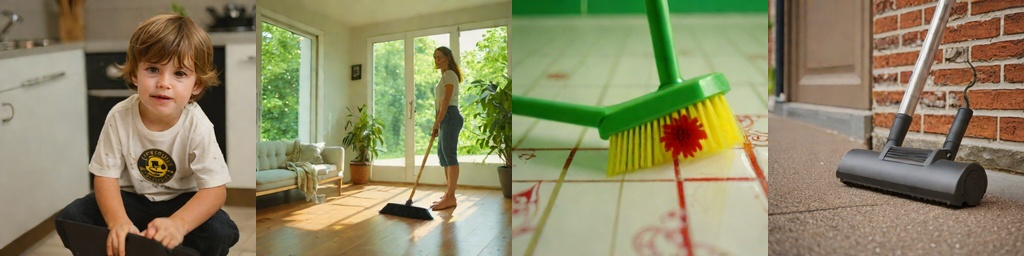} \\
\hline
sd30 & \includegraphics[height=1.87cm]{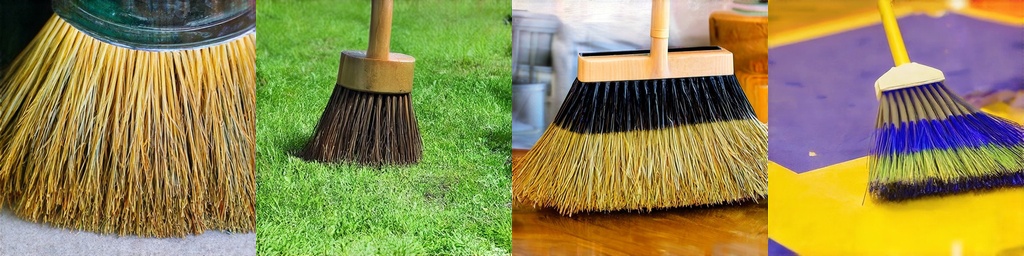} & \includegraphics[height=1.87cm]{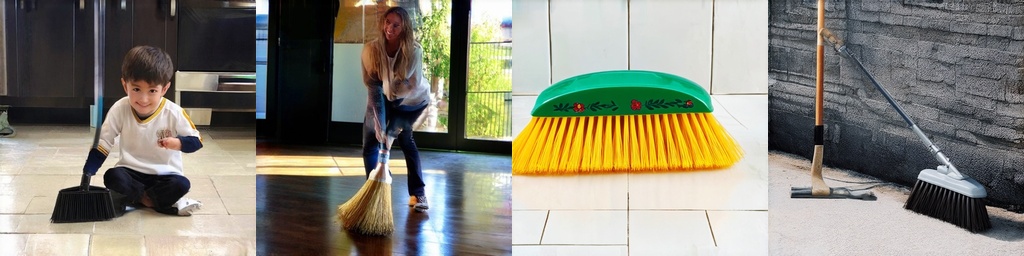} \\
\hline
flux dev & \includegraphics[height=1.87cm]{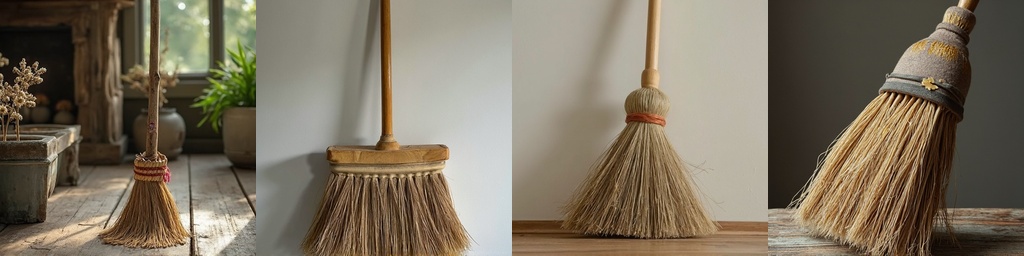} & \includegraphics[height=1.87cm]{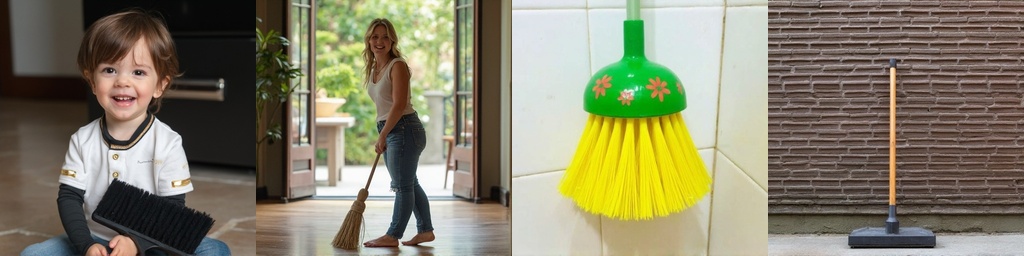} \\
\hline
flux-schnell & \includegraphics[height=1.87cm]{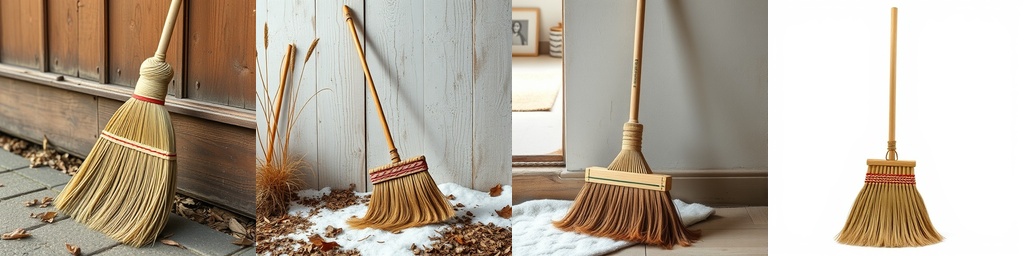} & \includegraphics[height=1.87cm]{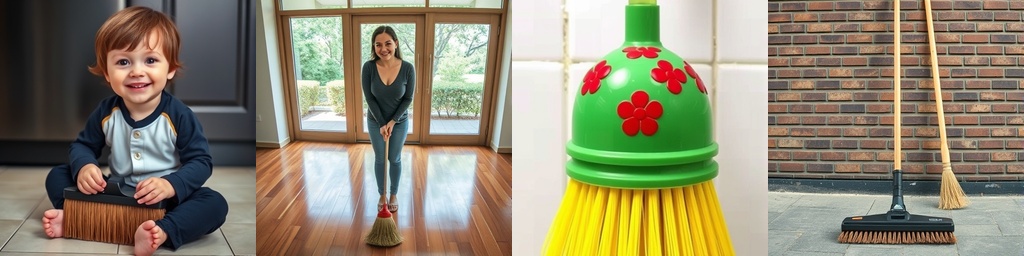} \\
\hline
sd35 & \includegraphics[height=1.87cm]{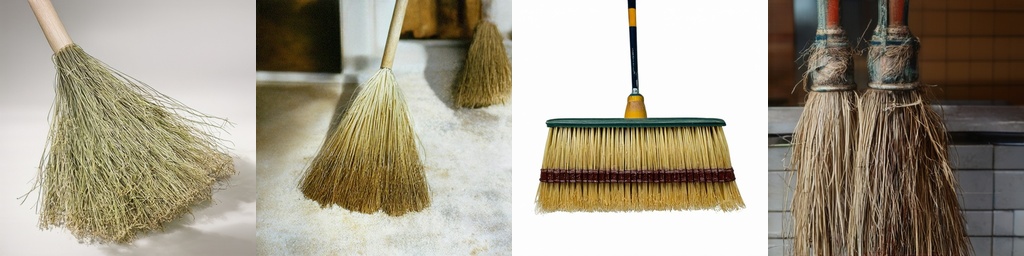} & \includegraphics[height=1.87cm]{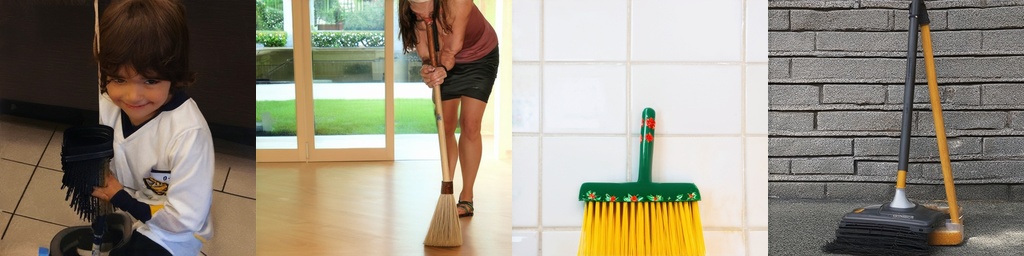} \\
\hline
sd35-large & \includegraphics[height=1.87cm]{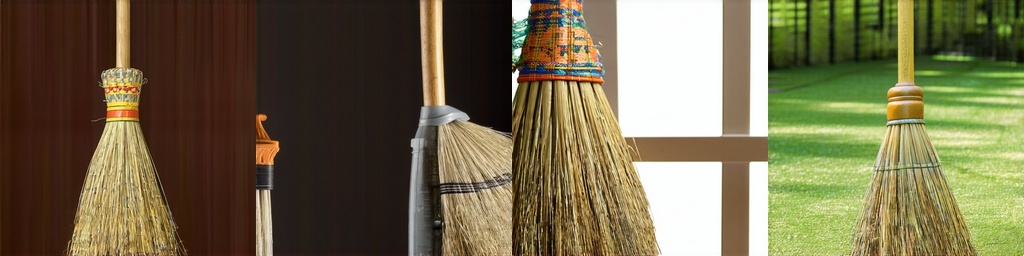} & \includegraphics[height=1.87cm]{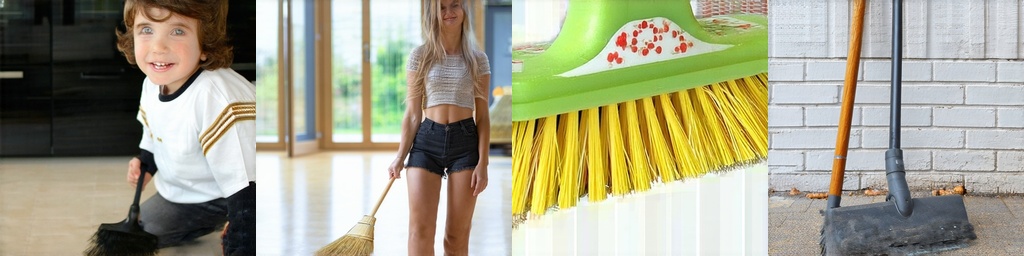} \\
\hline
sd35-turbo & \includegraphics[height=1.87cm]{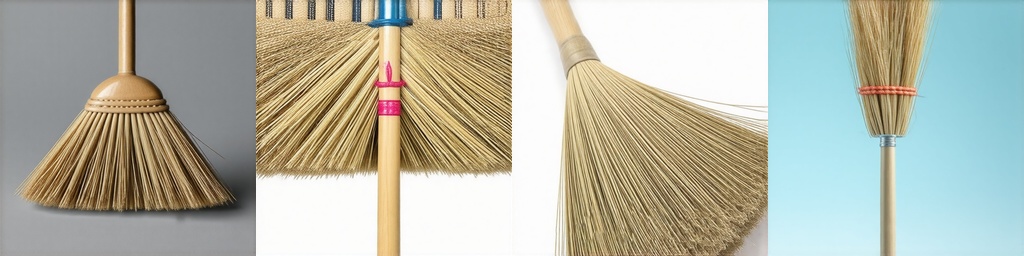} & \includegraphics[height=1.87cm]{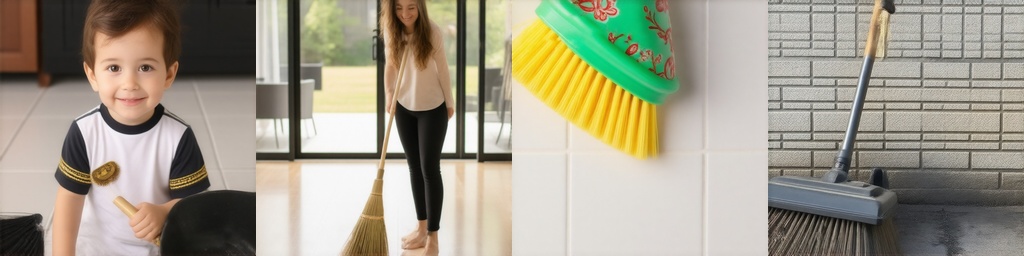} \\
\hline
sana & \includegraphics[height=1.87cm]{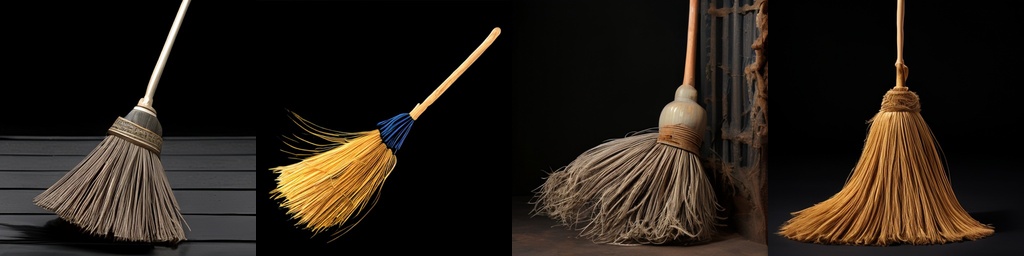} & \includegraphics[height=1.87cm]{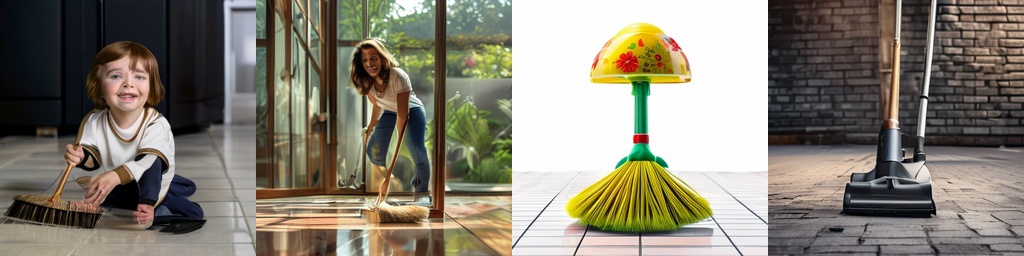} \\
\hline
qwen & \includegraphics[height=1.87cm]{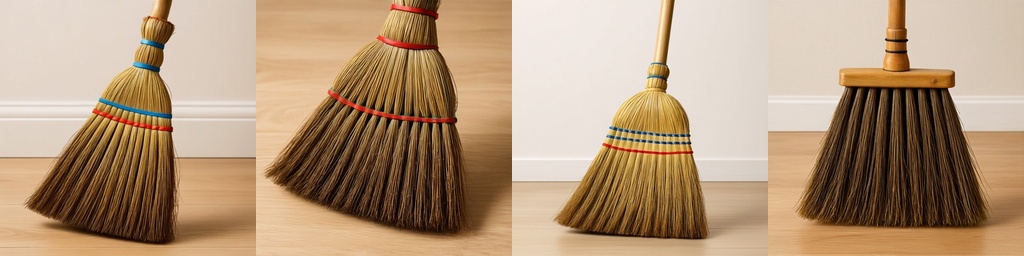} & \includegraphics[height=1.87cm]{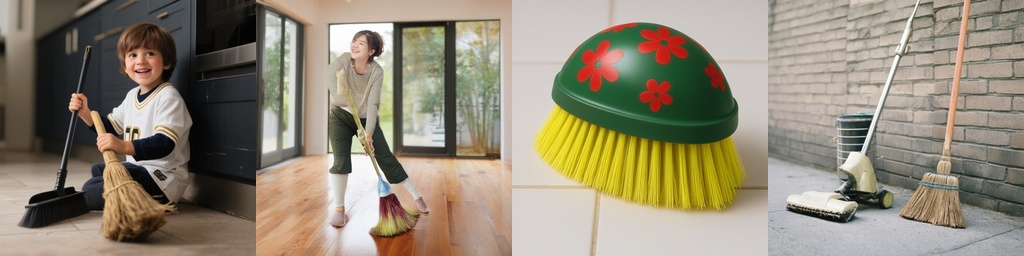} \\
\hline

\end{longtable}

\newpage
\section{Image Caption Samples}

\begin{longtable}{|m{3cm}|m{14cm}|}
\caption{The table below contains sample captions that were used as prompts in the paper. Captions were obtained through prompting the GPT4-nano model.}\\
\hline
\textbf{Image} & \textbf{Caption} \\
\hline
\endfirsthead

\multicolumn{2}{c}%
{\tablename\ \thetable\ -- \textit{Continued from previous page}} \\
\hline
\textbf{Image} & \textbf{Caption} \\
\hline
\endhead

\hline
\multicolumn{2}{r}{\textit{Continued on next page}} \\
\endfoot

\hline
\endlastfoot

\includegraphics[width=3cm]{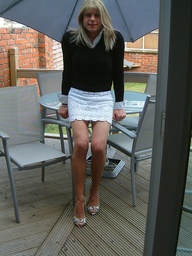} & The image features a woman standing on a wooden patio, framed by an open door in the foreground. She is holding a large umbrella against the backdrop of a brick house and a small garden area, with outdoor furniture and a grayish sky indicating an overcast day. The camera angle captures her from a straight-on perspective, focusing on her entire body and the surrounding outdoor setting. \\
\hline
\includegraphics[width=3cm]{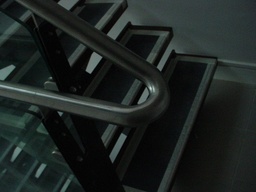} & The image captures an indoor setting with stair steps viewed from a slightly elevated angle, focusing on the black metal handrail and the dark stairs below. In the background, a glass railing and the railing's supporting structure are visible, contributing to a modern architectural aesthetic. The lighting is subdued, emphasizing the metallic and glass materials in the scene. \\
\hline
\includegraphics[width=3cm]{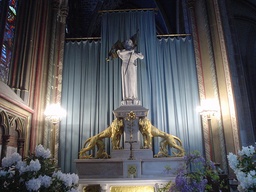} & The image features a religious display set against a richly decorated interior with vertical, multicolored striped curtains and ornate wood paneling. In the foreground, a pedestal supports a statue of Jesus Christ on the cross, flanked by two golden lion-like sculptures with outstretched paws. The camera angle is slightly upward, emphasizing the statue and the intricate details of the backdrop. \\
\hline
\includegraphics[width=3cm]{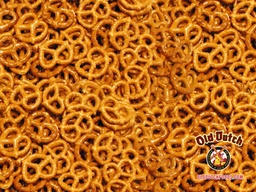} & The image displays a dense layer of golden-brown pretzels filling the entire frame, creating a textured background. The foreground is dominated by the uniform, shiny pretzels, shot from a close-up, slightly overhead camera angle that emphasizes their shape and glossiness. There are no other distinct objects or elements visible in the background. \\
\hline
\includegraphics[width=3cm]{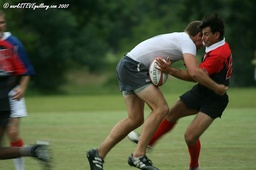} & A rugby match takes place outdoors on a lush green field, with a blurred background of trees indicating a park-like setting. In the foreground, two players are engaged in a tense tackle, with one player wearing a white shirt and gray shorts holding the rugby ball, while the other, in a red and black jersey, attempts to challenge him. The camera angle is slightly tilted and close-up, capturing the intensity of the moment and emphasizing the players' dynamic movements. \\
\hline
\includegraphics[width=3cm]{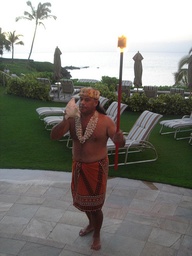} & The image features a lush, well-maintained garden area overlooking a body of water, with a few tall palm trees in the background. In the foreground, a traditionally dressed performer with a lei around his neck holds a lit torch or flame, standing on a paved surface. The camera angle captures him at eye level, emphasizing his cultural attire and the scenic outdoor setting behind him. \\
\hline
\includegraphics[width=3cm]{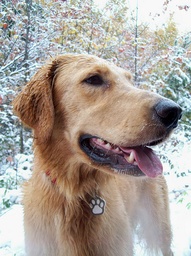} & The image features a close-up side profile of a golden retriever with a friendly expression, its mouth slightly open and tongue visible. The dog is wearing a collar with a paw print tag and is captured from a slightly lower angle, highlighting its head and upper body. In the background, there is a snowy landscape with trees and falling snowflakes, creating a wintery atmosphere. \\
\hline
\includegraphics[width=3cm]{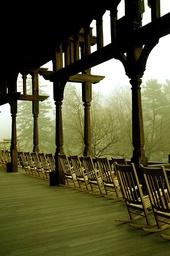} & The image depicts a covered porch or veranda with a row of wooden rocking chairs arranged along the railing, facing outward. The background features a misty, foggy landscape with leafless trees, creating a serene and slightly mysterious ambiance. The camera angle is level with the chairs, capturing the perspective of someone standing on the porch looking toward the landscape. \\
\hline
\includegraphics[width=3cm]{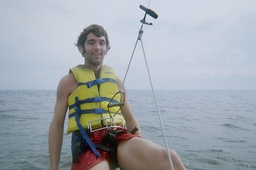} & A young man is sitting on a boat or dock near the water, wearing a bright yellow life jacket. He is smiling and looking toward the camera, with an ocean or large body of water and an overcast sky in the background. The image is taken from a slightly low angle, capturing his upper body and part of his legs, with the water stretching out behind him. \\
\hline
\includegraphics[width=3cm]{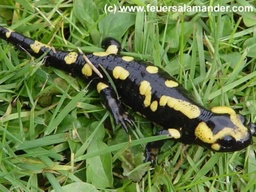} & The image features a close-up view of a fire salamander with striking black and yellow coloration, positioned amidst lush green grass. The camera angle captures the salamander from a top-down perspective, highlighting its elongated body, glossy skin, and distinctive markings. The background consists of dense, vibrant grass blades, providing a natural and contrasting setting for the amphibian. \\
\hline
\end{longtable}

\newpage

\section{Imagenet-1k Subset}
\begin{longtable}{|l|p{12cm}|}

\caption{We visualize image samples from the subset of the imagenet-1k (200 classes, 500 IPC) that we use as real data in our experiments}\\

\hline
\textbf{Class Name} & \textbf{Sample Images} \\
\hline
\endfirsthead

\multicolumn{2}{c}%
{\tablename\ \thetable\ -- \textit{Continued from previous page}} \\
\hline
\textbf{Class Name} & \textbf{Sample Images} \\
\hline
\endhead

\hline
\multicolumn{2}{r}{\textit{Continued on next page}} \\
\endfoot

\hline
\endlastfoot

goldfish & \includegraphics[width=12cm]{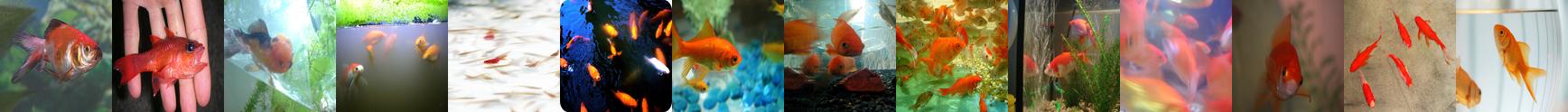} \\
\hline
european fire salamander & \includegraphics[width=12cm]{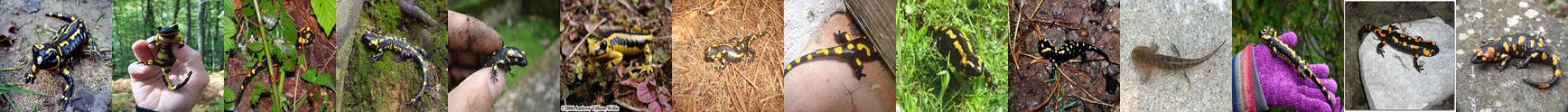} \\
\hline
bullfrog & \includegraphics[width=12cm]{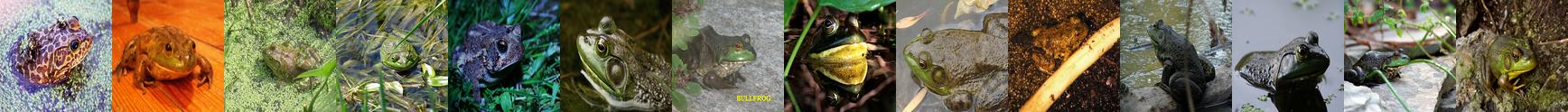} \\
\hline
tailed frog & \includegraphics[width=12cm]{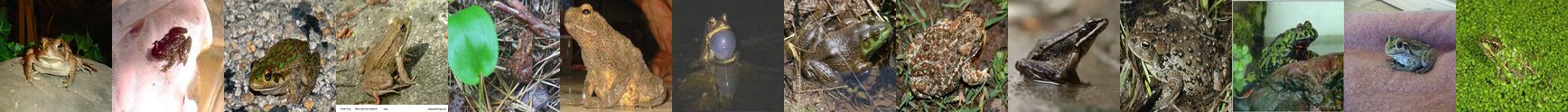} \\
\hline
american alligator & \includegraphics[width=12cm]{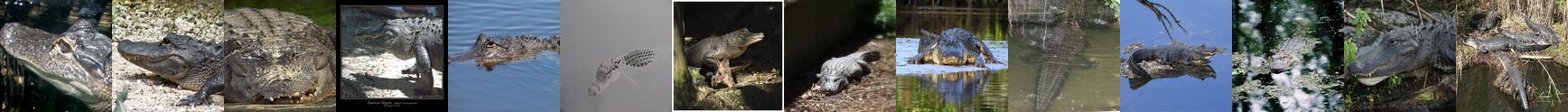} \\
\hline
boa constrictor & \includegraphics[width=12cm]{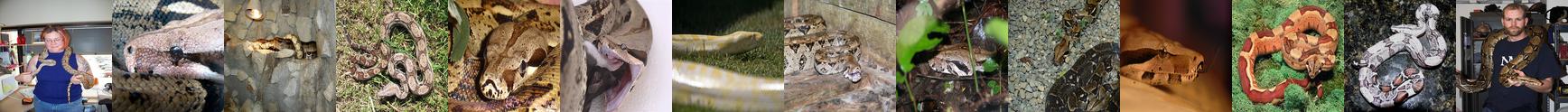} \\
\hline
trilobite & \includegraphics[width=12cm]{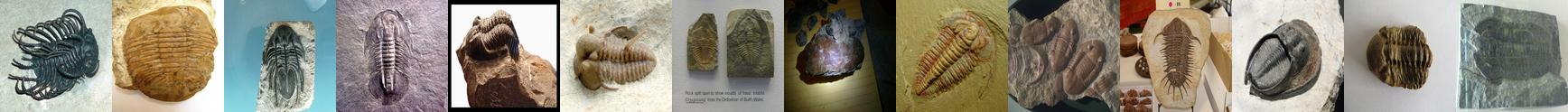} \\
\hline
scorpion & \includegraphics[width=12cm]{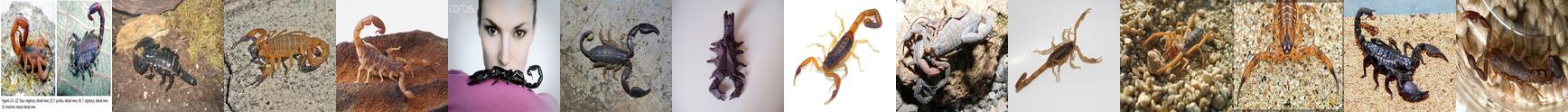} \\
\hline
black widow & \includegraphics[width=12cm]{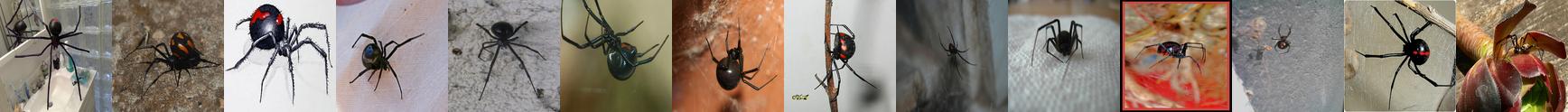} \\
\hline
tarantula & \includegraphics[width=12cm]{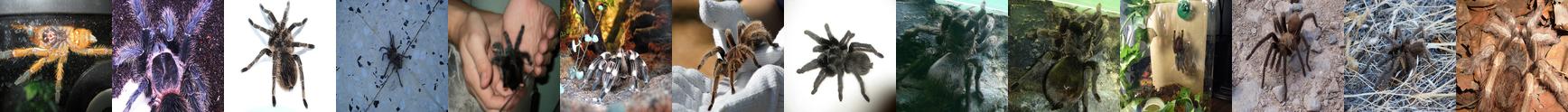} \\
\hline
centipede & \includegraphics[width=12cm]{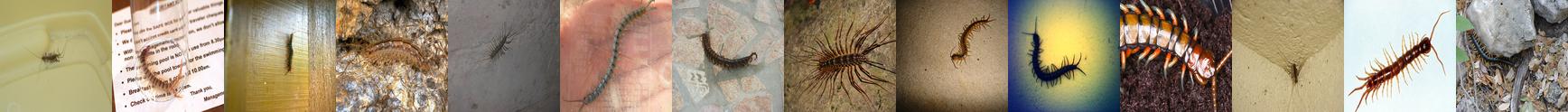} \\
\hline
goose & \includegraphics[width=12cm]{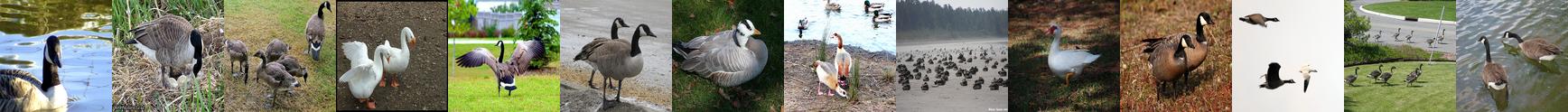} \\
\hline
koala & \includegraphics[width=12cm]{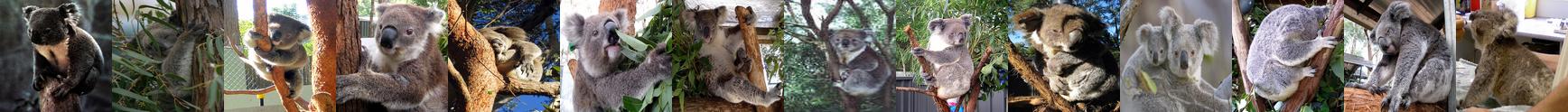} \\
\hline
jellyfish & \includegraphics[width=12cm]{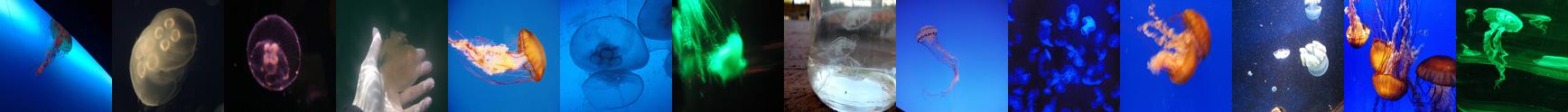} \\
\hline
brain coral & \includegraphics[width=12cm]{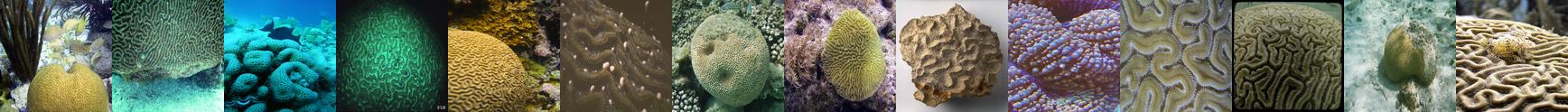} \\
\hline
snail & \includegraphics[width=12cm]{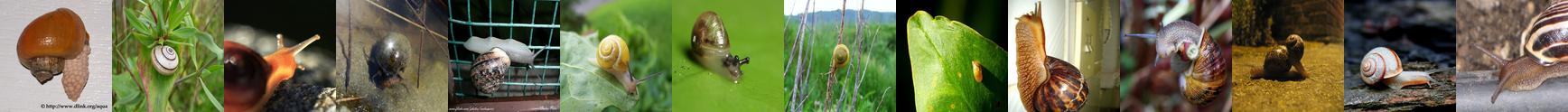} \\
\hline
slug & \includegraphics[width=12cm]{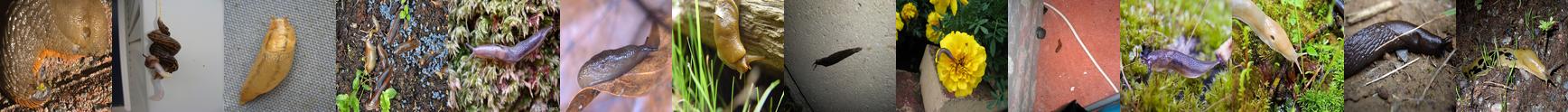} \\
\hline
sea slug & \includegraphics[width=12cm]{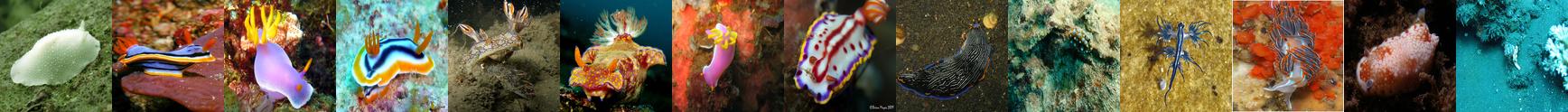} \\
\hline
american lobster & \includegraphics[width=12cm]{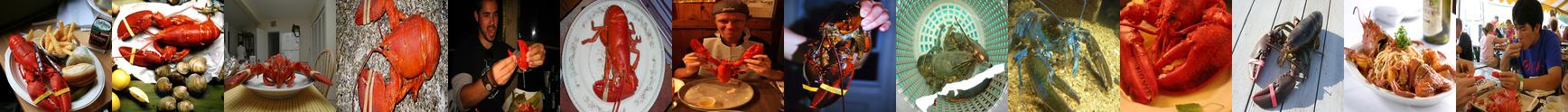} \\
\hline
spiny lobster & \includegraphics[width=12cm]{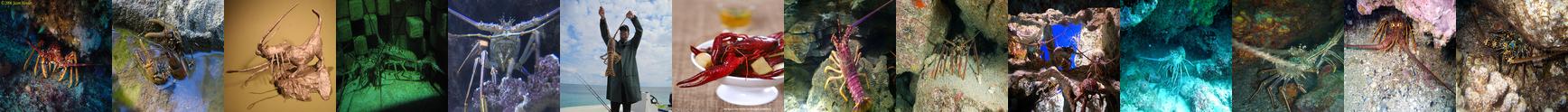} \\
\hline
black stork & \includegraphics[width=12cm]{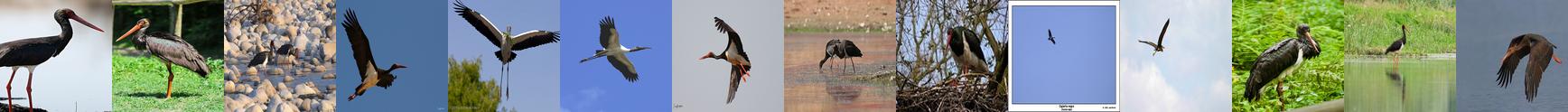} \\
\hline
king penguin & \includegraphics[width=12cm]{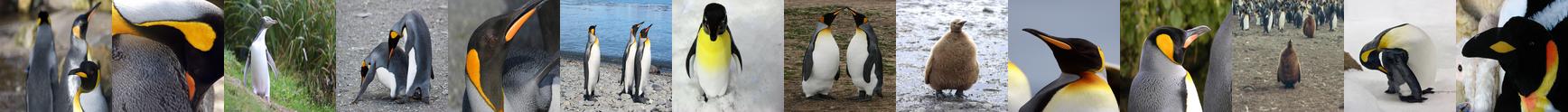} \\
\hline
albatross & \includegraphics[width=12cm]{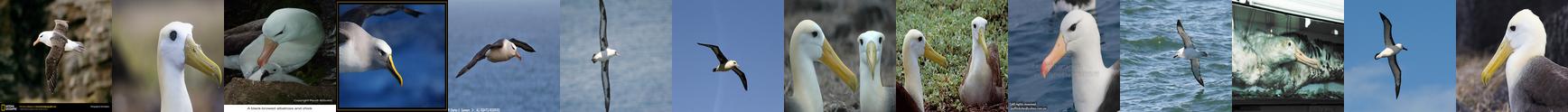} \\
\hline
dugong & \includegraphics[width=12cm]{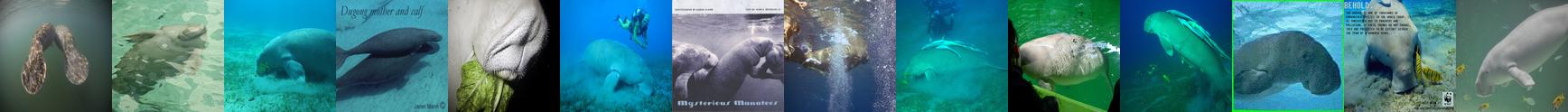} \\
\hline
chihuahua & \includegraphics[width=12cm]{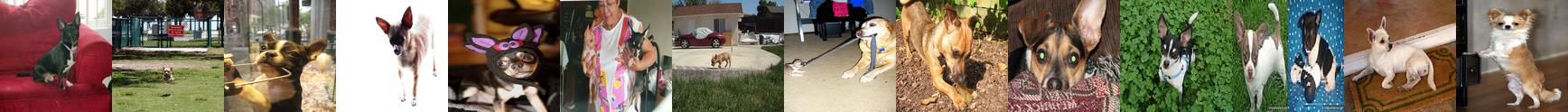} \\
\hline
yorkshire terrier & \includegraphics[width=12cm]{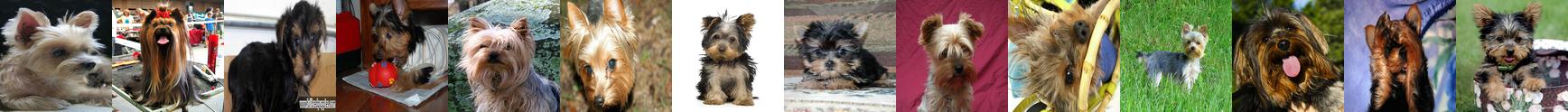} \\
\hline
golden retriever & \includegraphics[width=12cm]{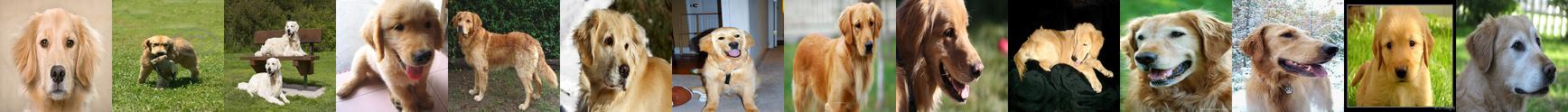} \\
\hline
labrador retriever & \includegraphics[width=12cm]{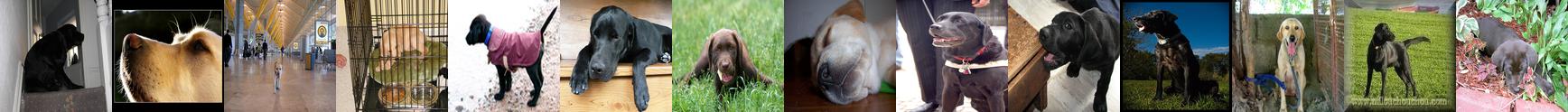} \\
\hline
german shepherd & \includegraphics[width=12cm]{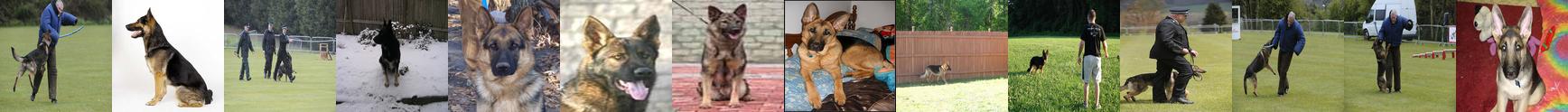} \\
\hline
standard poodle & \includegraphics[width=12cm]{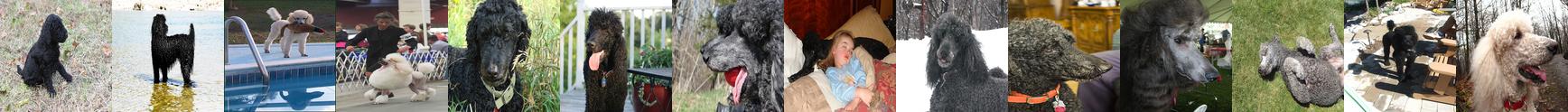} \\
\hline
tabby cat & \includegraphics[width=12cm]{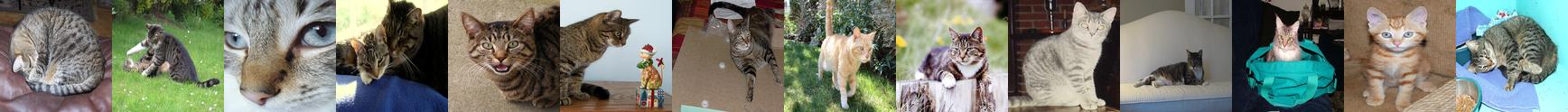} \\
\hline
persian cat & \includegraphics[width=12cm]{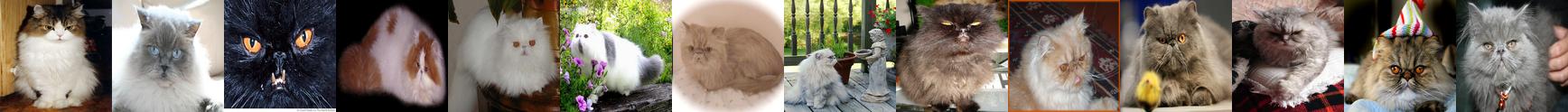} \\
\hline
egyptian cat & \includegraphics[width=12cm]{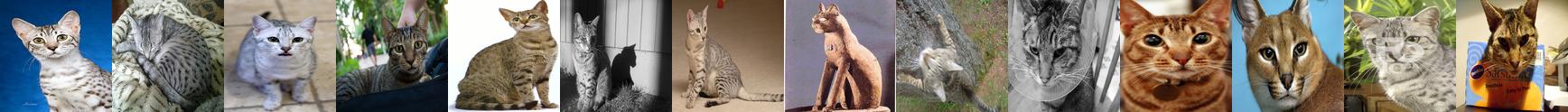} \\
\hline
cougar & \includegraphics[width=12cm]{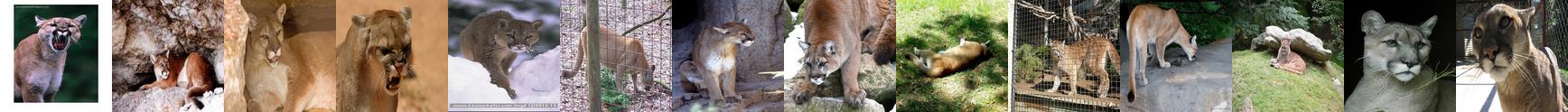} \\
\hline
lion & \includegraphics[width=12cm]{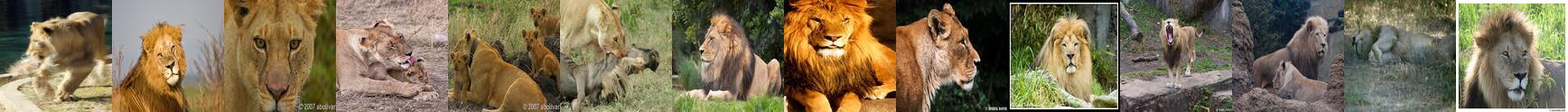} \\
\hline
brown bear & \includegraphics[width=12cm]{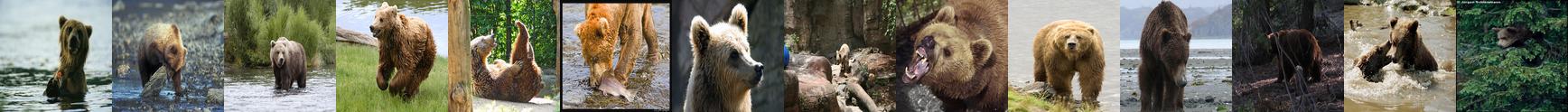} \\
\hline
ladybug & \includegraphics[width=12cm]{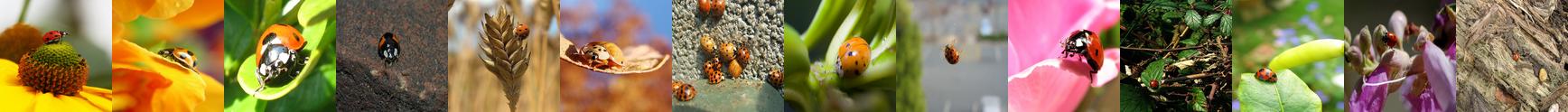} \\
\hline
fly & \includegraphics[width=12cm]{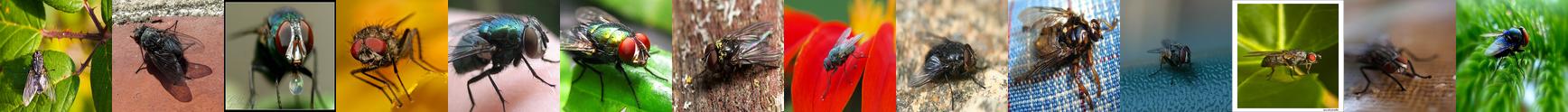} \\
\hline
bee & \includegraphics[width=12cm]{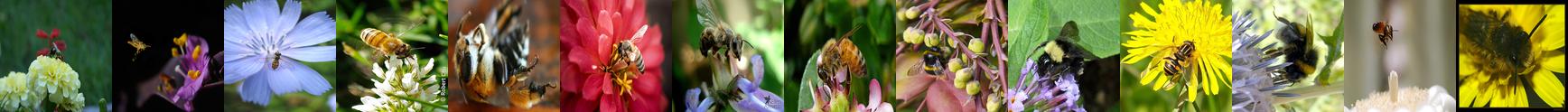} \\
\hline
grasshoppe & \includegraphics[width=12cm]{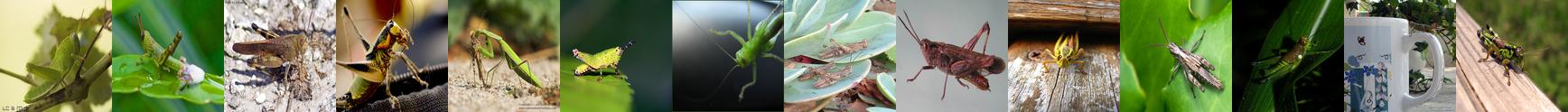} \\
\hline
walking stick & \includegraphics[width=12cm]{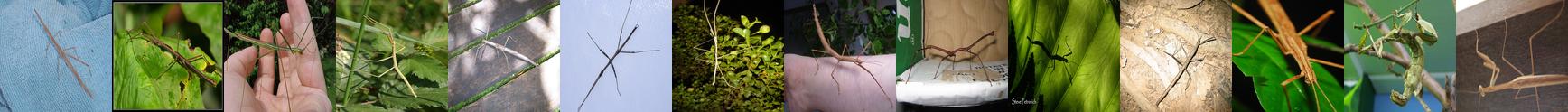} \\
\hline
cockroach & \includegraphics[width=12cm]{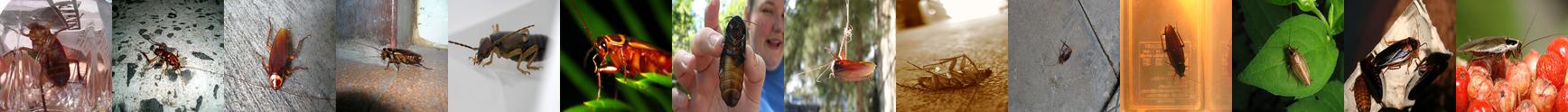} \\
\hline
mantis & \includegraphics[width=12cm]{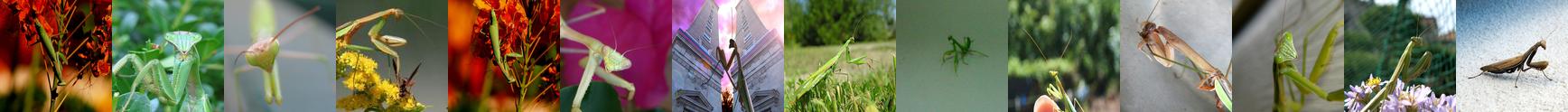} \\
\hline
dragonfly & \includegraphics[width=12cm]{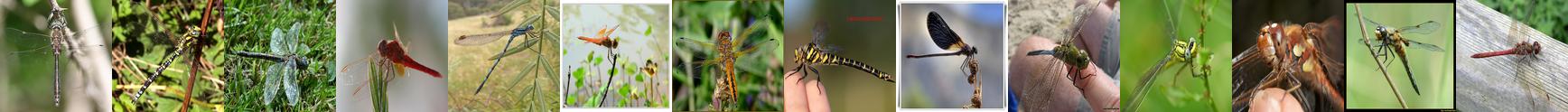} \\
\hline
monarch & \includegraphics[width=12cm]{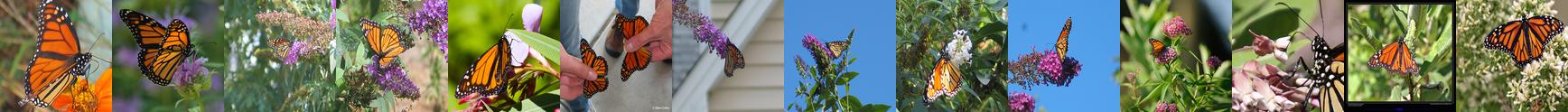} \\
\hline
sulphur butterfly & \includegraphics[width=12cm]{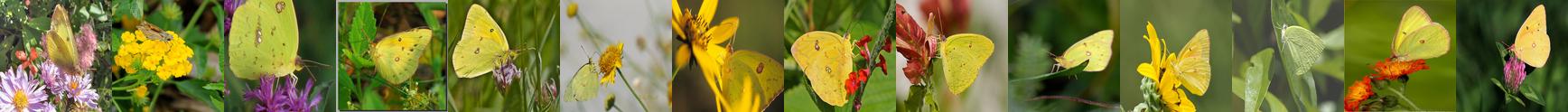} \\
\hline
sea cucumber & \includegraphics[width=12cm]{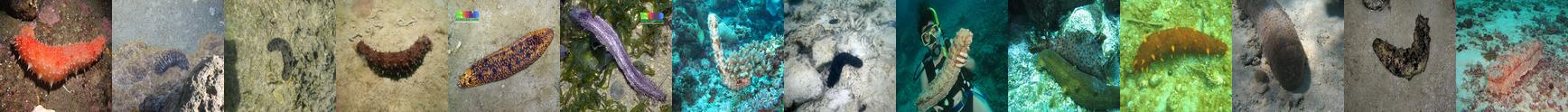} \\
\hline
guinea pig & \includegraphics[width=12cm]{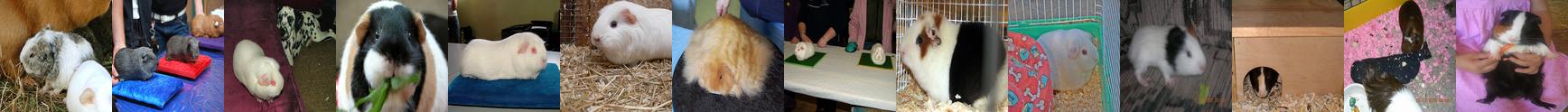} \\
\hline
hog & \includegraphics[width=12cm]{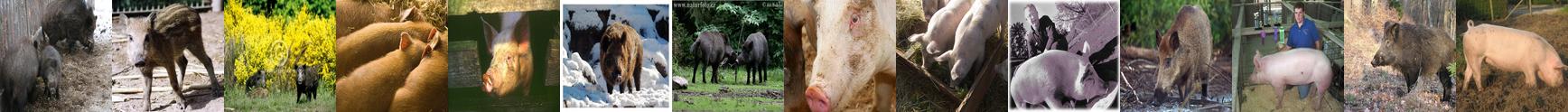} \\
\hline
ox & \includegraphics[width=12cm]{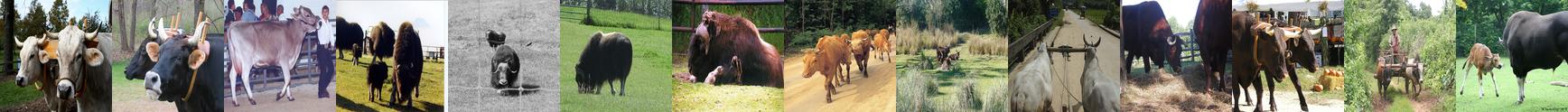} \\
\hline
bison & \includegraphics[width=12cm]{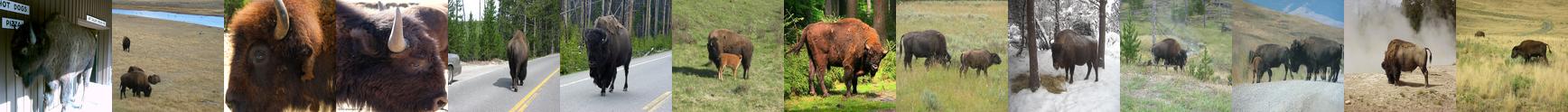} \\
\hline
bighorn & \includegraphics[width=12cm]{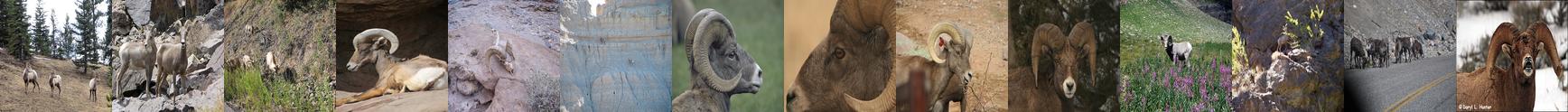} \\
\hline
gazelle & \includegraphics[width=12cm]{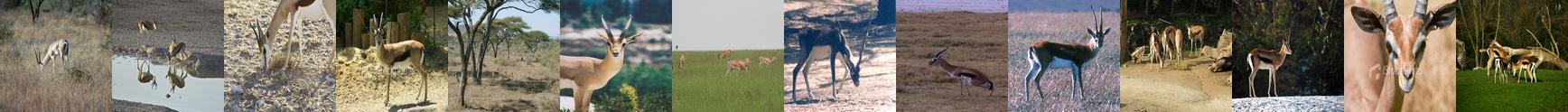} \\
\hline
arabian camel & \includegraphics[width=12cm]{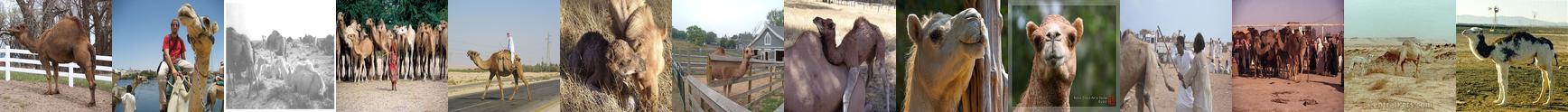} \\
\hline
orangutan & \includegraphics[width=12cm]{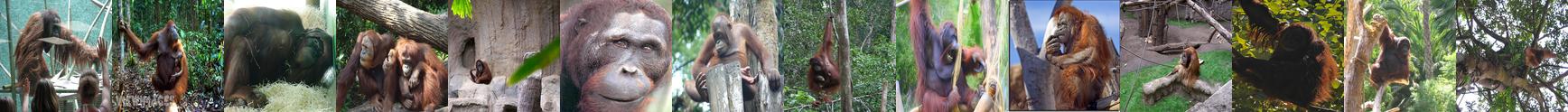} \\
\hline
chimpanzee & \includegraphics[width=12cm]{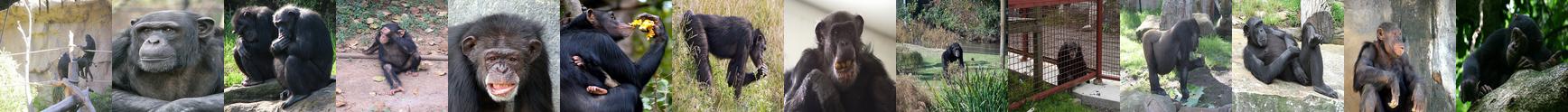} \\
\hline
baboon & \includegraphics[width=12cm]{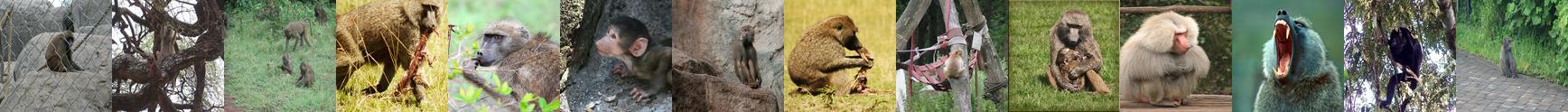} \\
\hline
african elephant & \includegraphics[width=12cm]{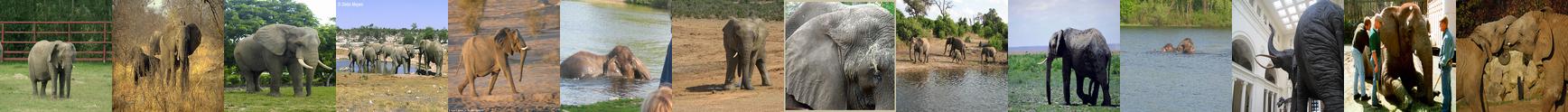} \\
\hline
panda & \includegraphics[width=12cm]{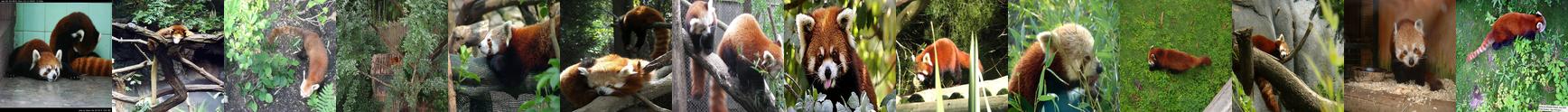} \\
\hline
abacus & \includegraphics[width=12cm]{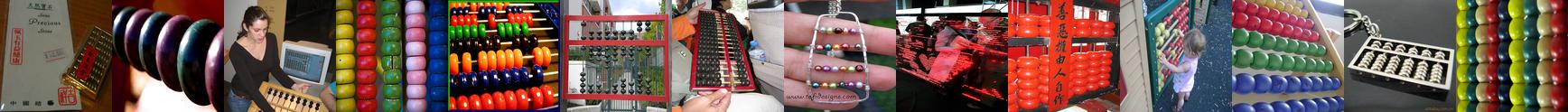} \\
\hline
academic gown & \includegraphics[width=12cm]{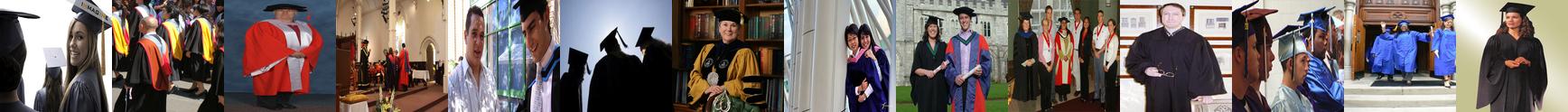} \\
\hline
altar & \includegraphics[width=12cm]{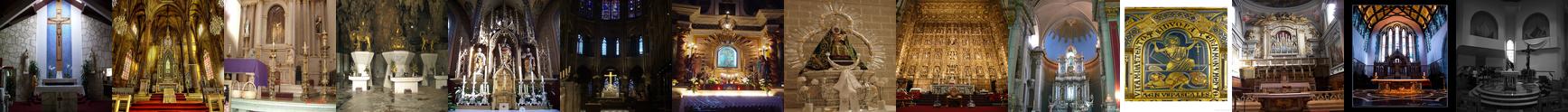} \\
\hline
apron & \includegraphics[width=12cm]{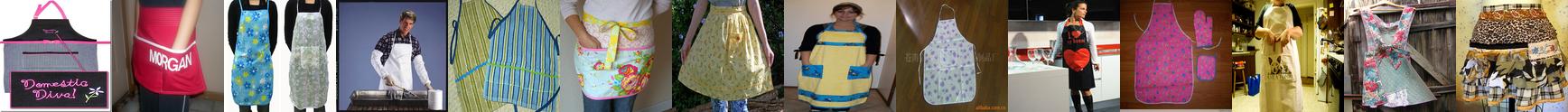} \\
\hline
backpack & \includegraphics[width=12cm]{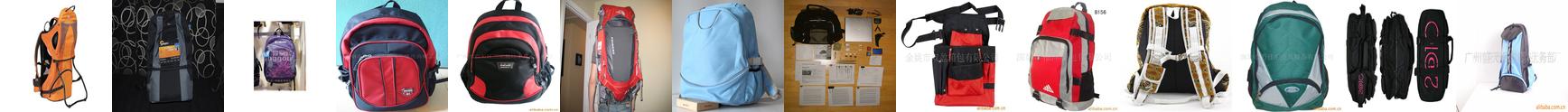} \\
\hline
bannister & \includegraphics[width=12cm]{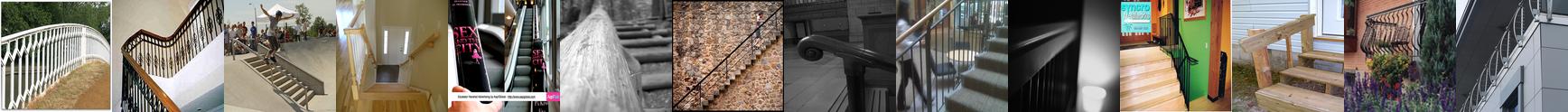} \\
\hline
barbershop & \includegraphics[width=12cm]{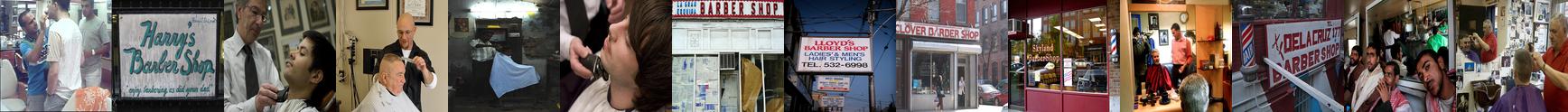} \\
\hline
barn & \includegraphics[width=12cm]{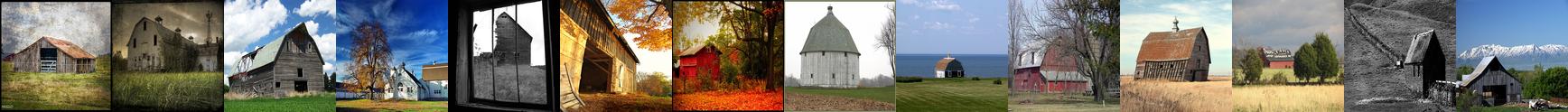} \\
\hline
barrel & \includegraphics[width=12cm]{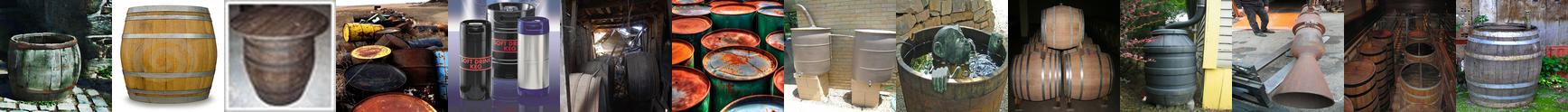} \\
\hline
basketball & \includegraphics[width=12cm]{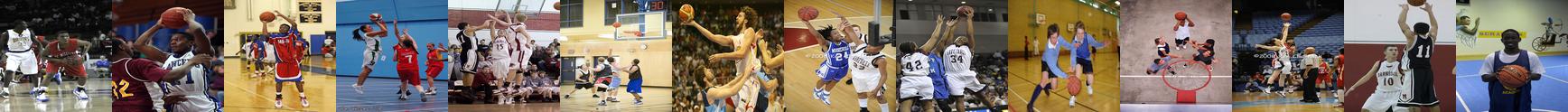} \\
\hline
bathtub & \includegraphics[width=12cm]{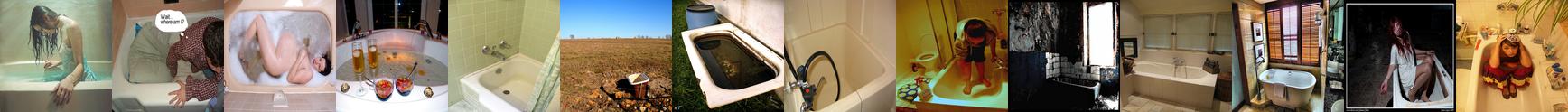} \\
\hline
beach wagon & \includegraphics[width=12cm]{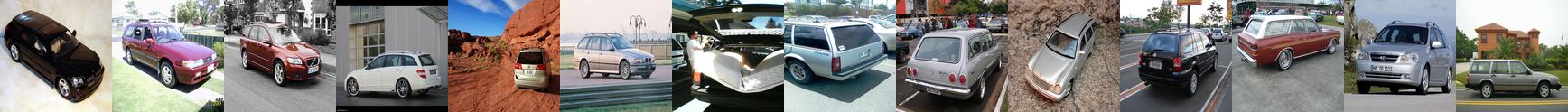} \\
\hline
beacon & \includegraphics[width=12cm]{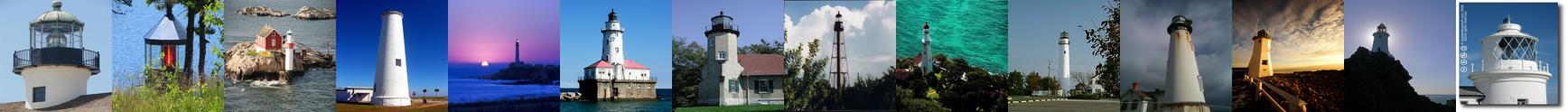} \\
\hline
beaker & \includegraphics[width=12cm]{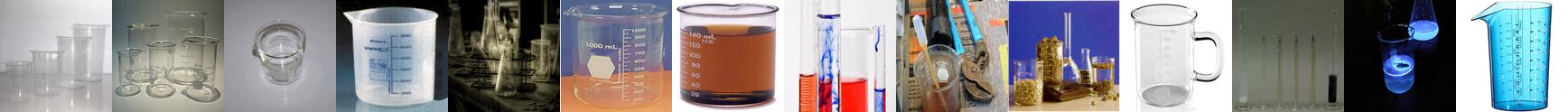} \\
\hline
beer bottle & \includegraphics[width=12cm]{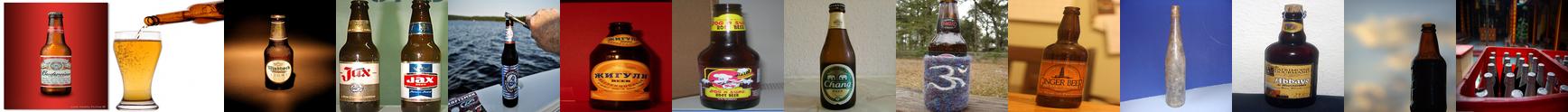} \\
\hline
bikini & \includegraphics[width=12cm]{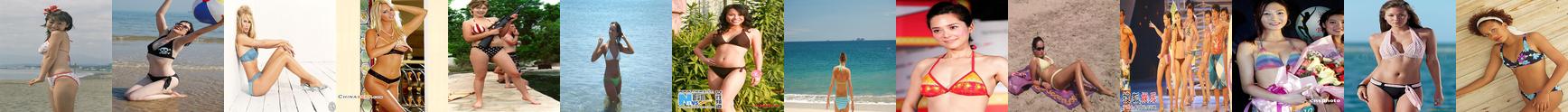} \\
\hline
binoculars & \includegraphics[width=12cm]{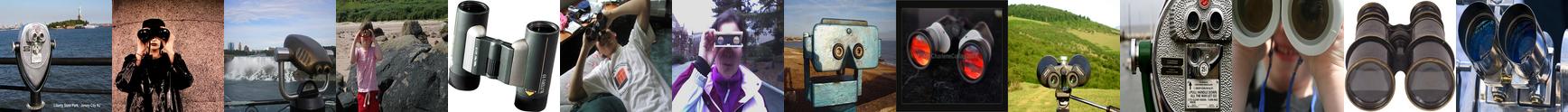} \\
\hline
birdhouse & \includegraphics[width=12cm]{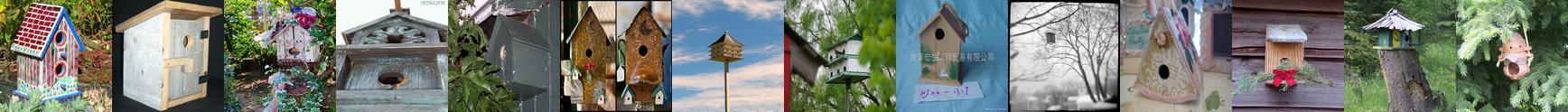} \\
\hline
bow tie & \includegraphics[width=12cm]{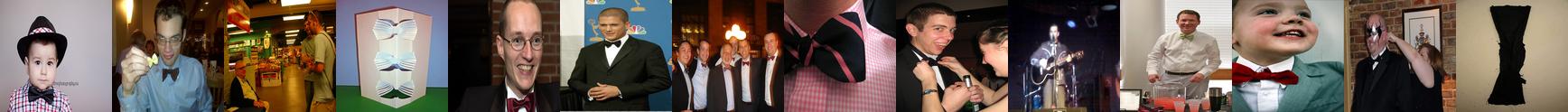} \\
\hline
brass & \includegraphics[width=12cm]{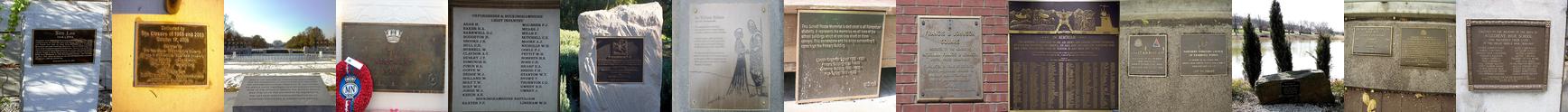} \\
\hline
broom & \includegraphics[width=12cm]{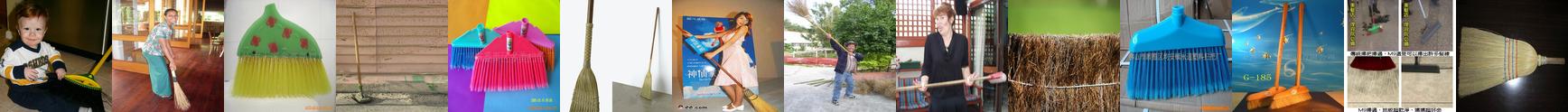} \\
\hline
bucket & \includegraphics[width=12cm]{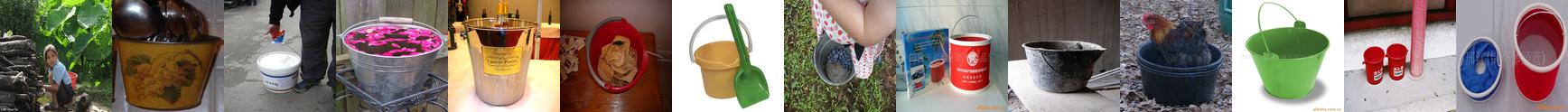} \\
\hline
bullet train & \includegraphics[width=12cm]{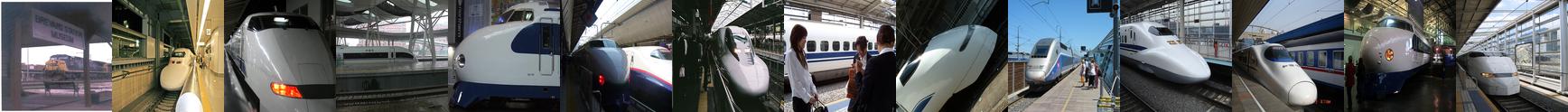} \\
\hline
butcher shop & \includegraphics[width=12cm]{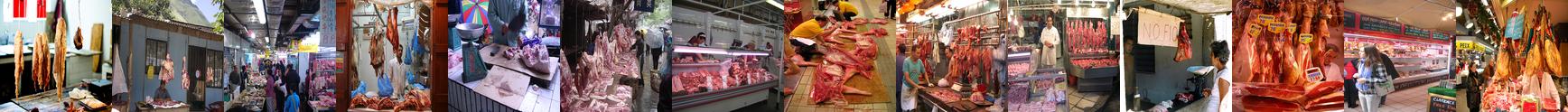} \\
\hline
candle & \includegraphics[width=12cm]{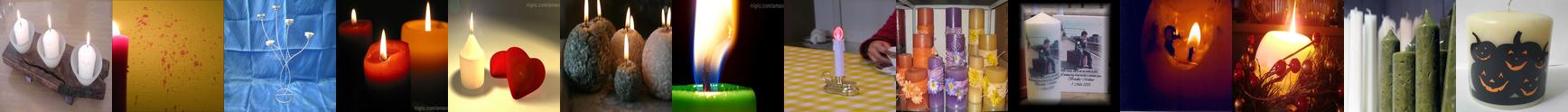} \\
\hline
cannon & \includegraphics[width=12cm]{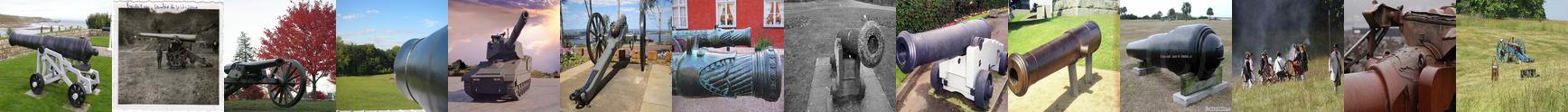} \\
\hline
cardigan & \includegraphics[width=12cm]{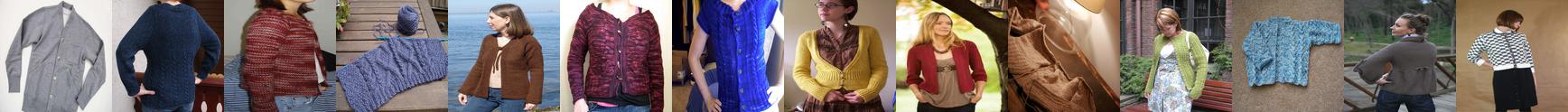} \\
\hline
cash machine & \includegraphics[width=12cm]{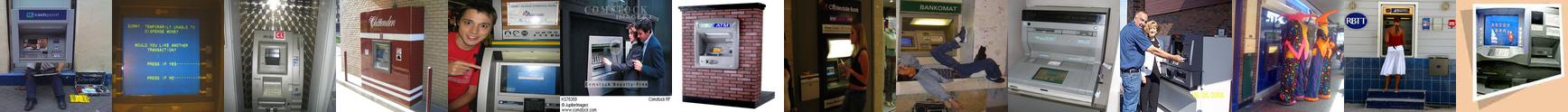} \\
\hline
CD player & \includegraphics[width=12cm]{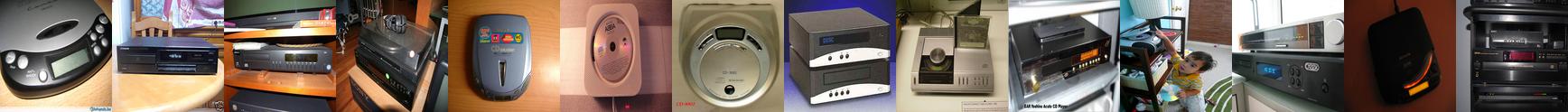} \\
\hline
chain & \includegraphics[width=12cm]{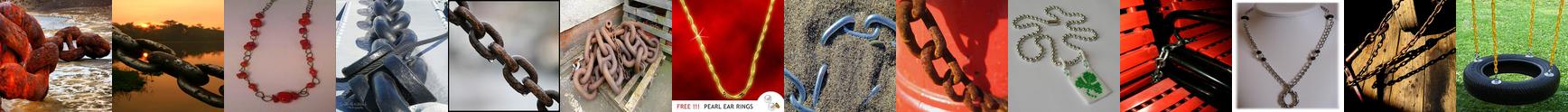} \\
\hline
chest & \includegraphics[width=12cm]{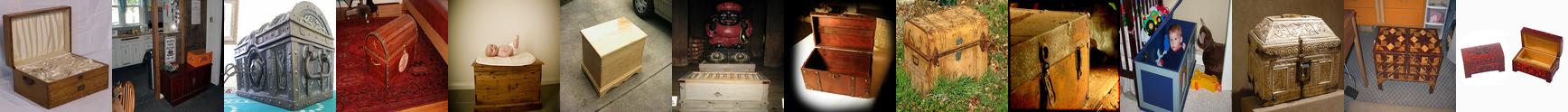} \\
\hline
christmas stocking & \includegraphics[width=12cm]{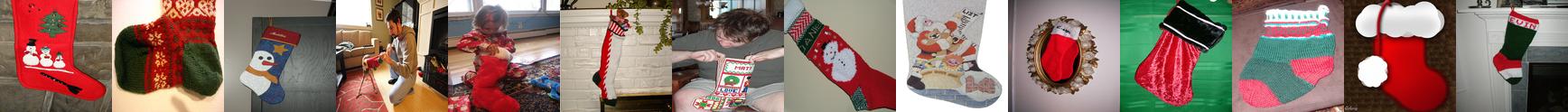} \\
\hline
cliff dwelling & \includegraphics[width=12cm]{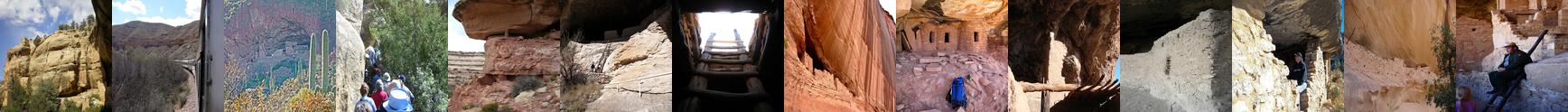} \\
\hline
computer keyboard & \includegraphics[width=12cm]{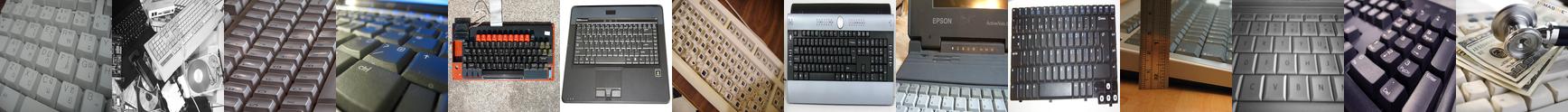} \\
\hline
confectionery & \includegraphics[width=12cm]{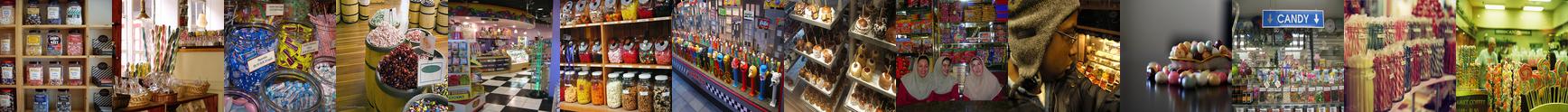} \\
\hline
convertible & \includegraphics[width=12cm]{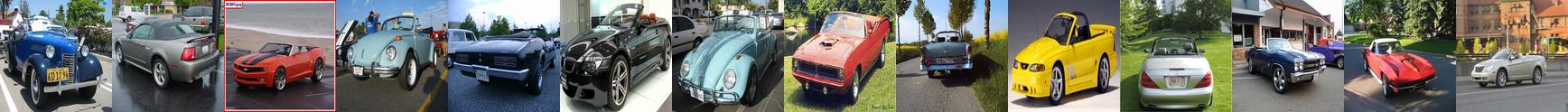} \\
\hline
crane & \includegraphics[width=12cm]{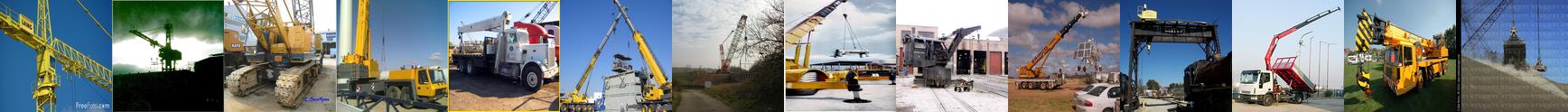} \\
\hline
dam & \includegraphics[width=12cm]{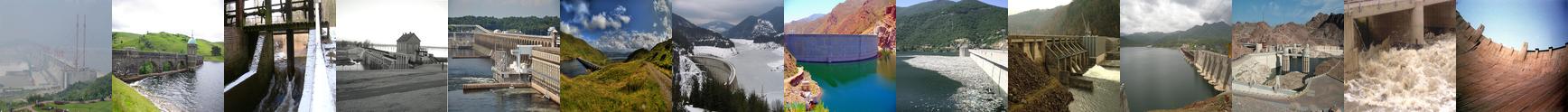} \\
\hline
desk & \includegraphics[width=12cm]{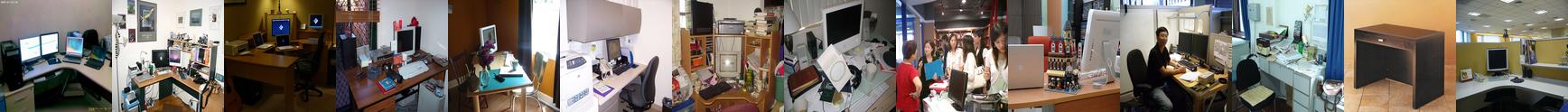} \\
\hline
dining table & \includegraphics[width=12cm]{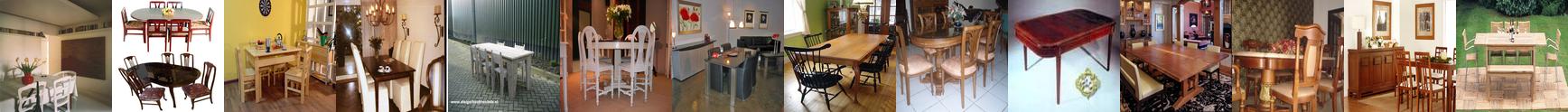} \\
\hline
drumstick & \includegraphics[width=12cm]{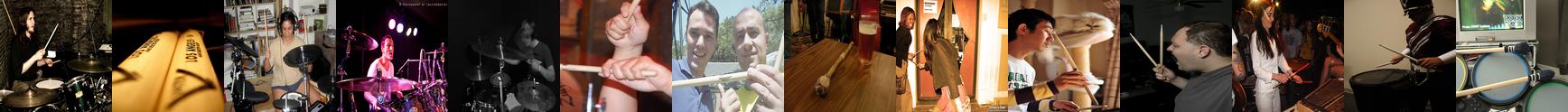} \\
\hline
dumbbell & \includegraphics[width=12cm]{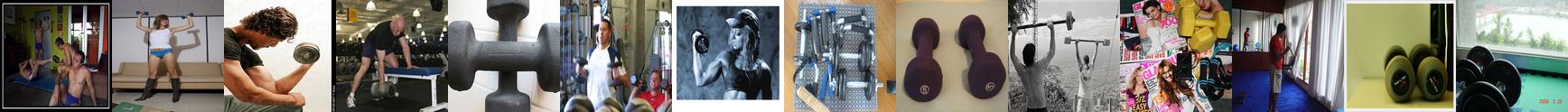} \\
\hline
flagpole & \includegraphics[width=12cm]{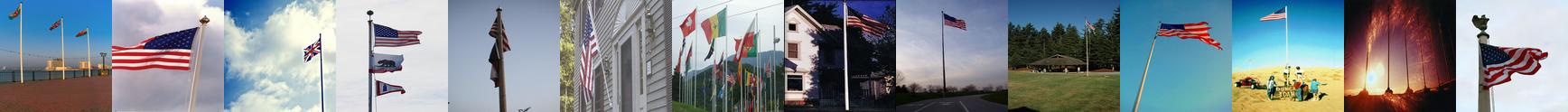} \\
\hline
fountain & \includegraphics[width=12cm]{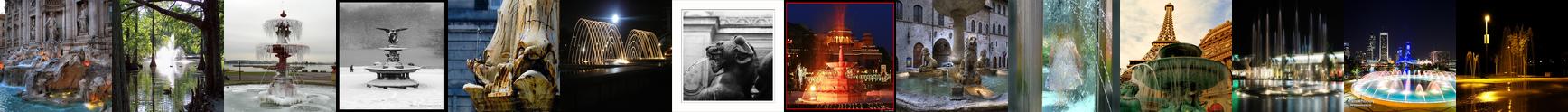} \\
\hline
freight car & \includegraphics[width=12cm]{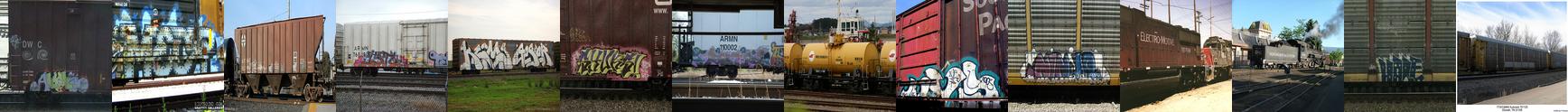} \\
\hline
frying pan & \includegraphics[width=12cm]{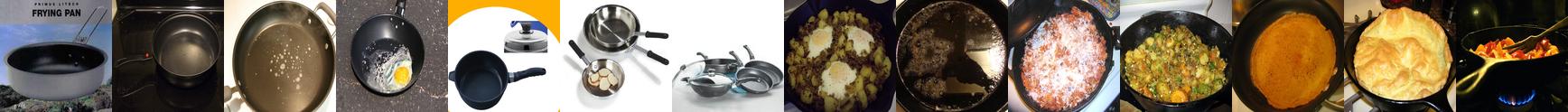} \\
\hline
fur coat & \includegraphics[width=12cm]{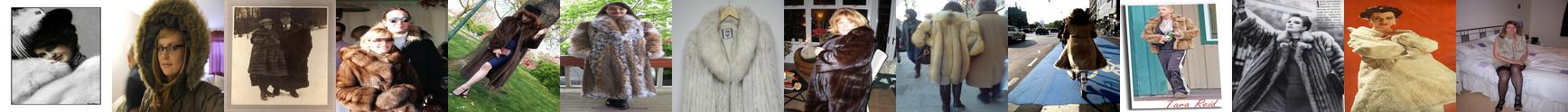} \\
\hline
gasmask & \includegraphics[width=12cm]{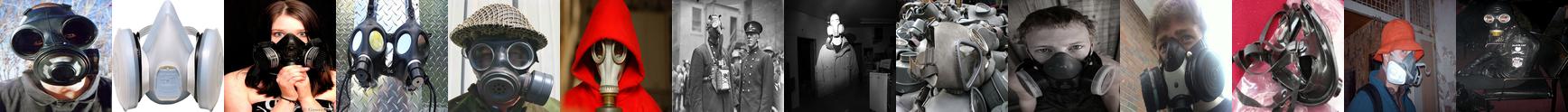} \\
\hline
go-kart & \includegraphics[width=12cm]{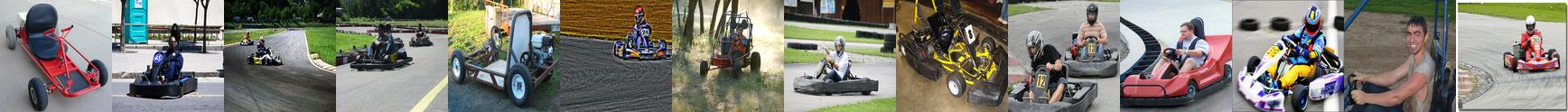} \\
\hline
gondola & \includegraphics[width=12cm]{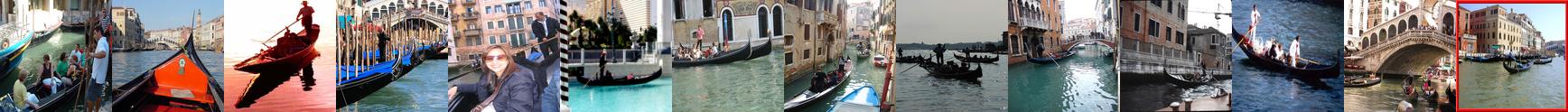} \\
\hline
hourglass & \includegraphics[width=12cm]{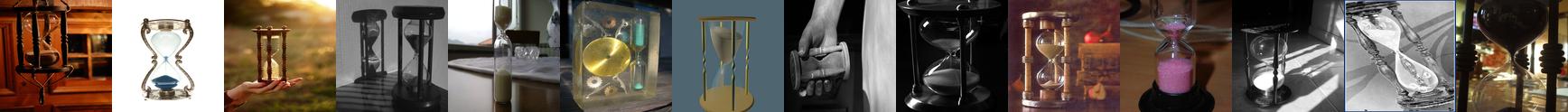} \\
\hline
iPod & \includegraphics[width=12cm]{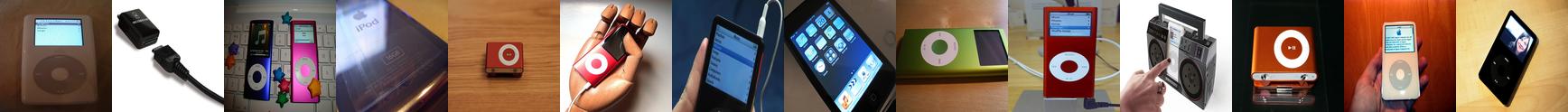} \\
\hline
jinrikisha & \includegraphics[width=12cm]{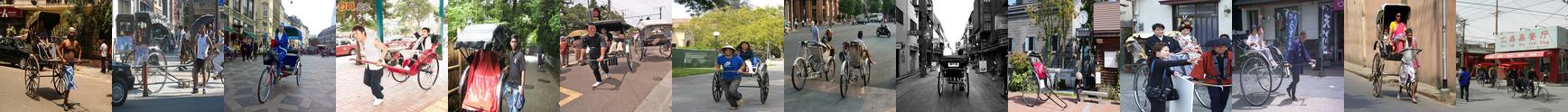} \\
\hline
kimono & \includegraphics[width=12cm]{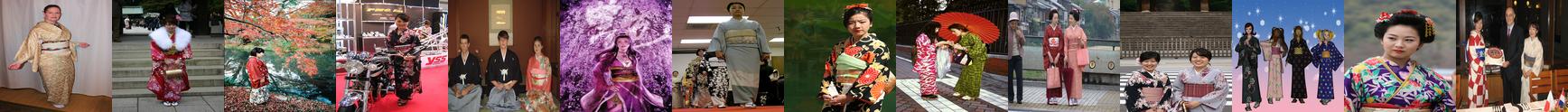} \\
\hline
lampshade & \includegraphics[width=12cm]{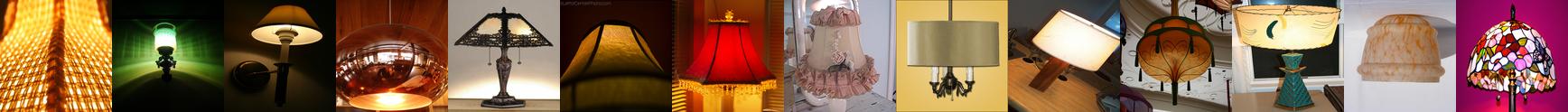} \\
\hline
lawn mower & \includegraphics[width=12cm]{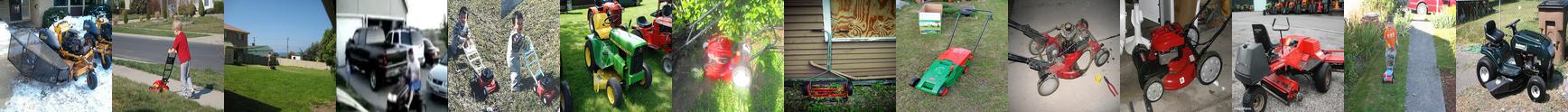} \\
\hline
lifeboat & \includegraphics[width=12cm]{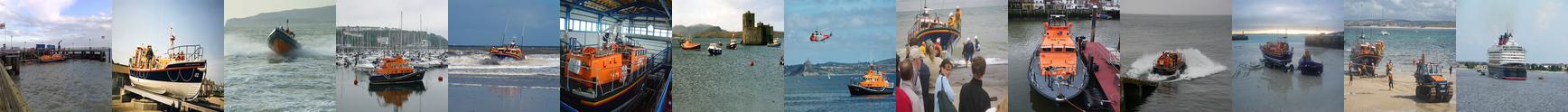} \\
\hline
limousine & \includegraphics[width=12cm]{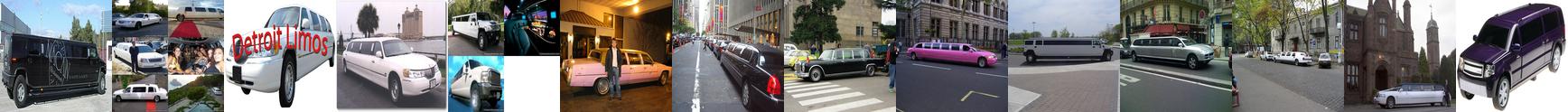} \\
\hline
magnetic compass & \includegraphics[width=12cm]{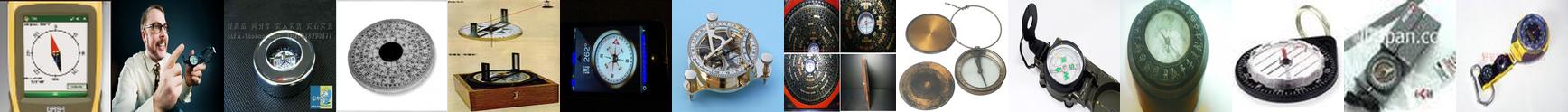} \\
\hline
maypole & \includegraphics[width=12cm]{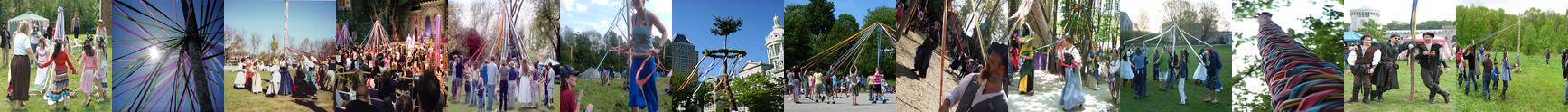} \\
\hline
military uniform & \includegraphics[width=12cm]{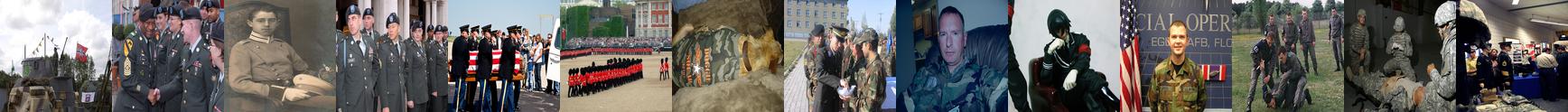} \\
\hline
miniskirt & \includegraphics[width=12cm]{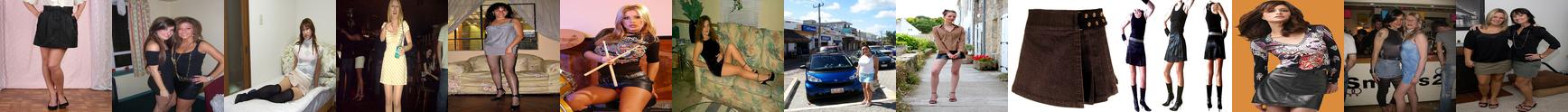} \\
\hline
moving van & \includegraphics[width=12cm]{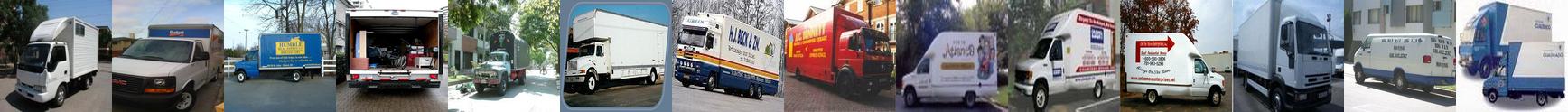} \\
\hline
nail & \includegraphics[width=12cm]{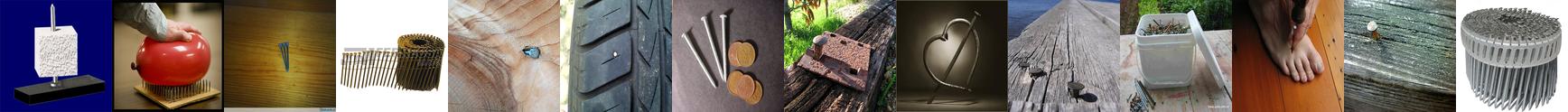} \\
\hline
neck brace & \includegraphics[width=12cm]{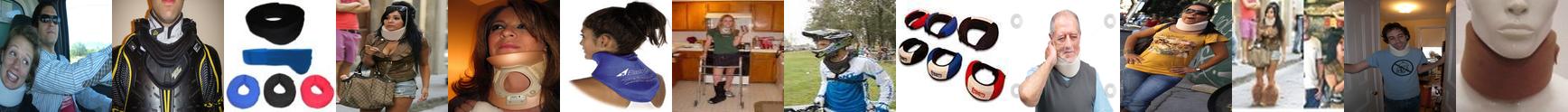} \\
\hline
obelisk & \includegraphics[width=12cm]{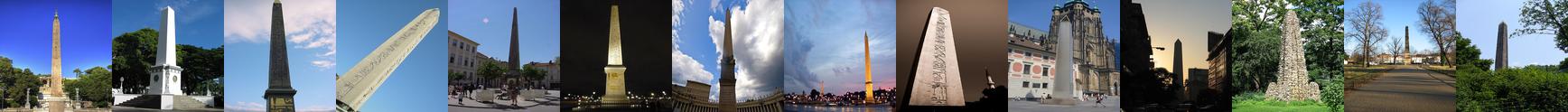} \\
\hline
oboe & \includegraphics[width=12cm]{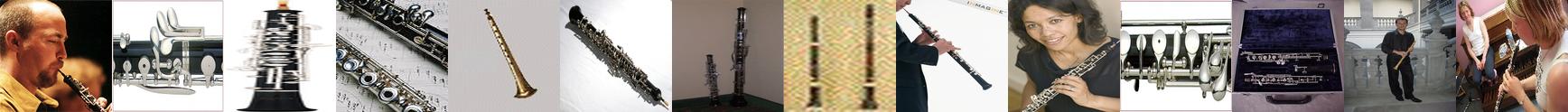} \\
\hline
pipe organ & \includegraphics[width=12cm]{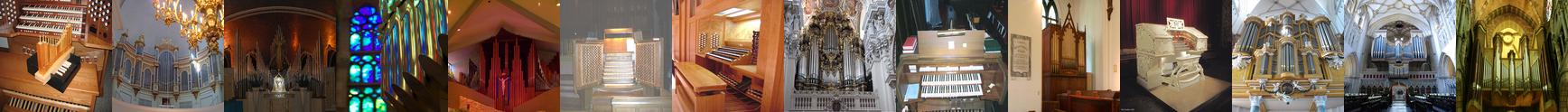} \\
\hline
parking meter & \includegraphics[width=12cm]{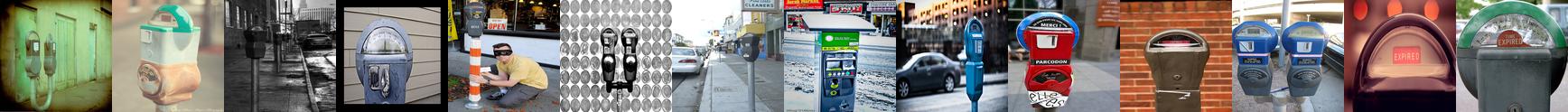} \\
\hline
pay-phone & \includegraphics[width=12cm]{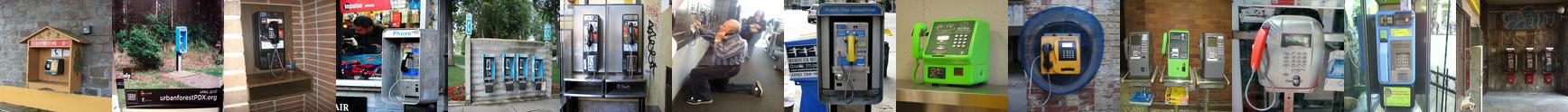} \\
\hline
picket fence & \includegraphics[width=12cm]{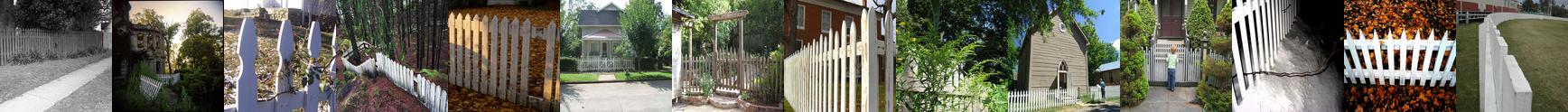} \\
\hline
pill bottle & \includegraphics[width=12cm]{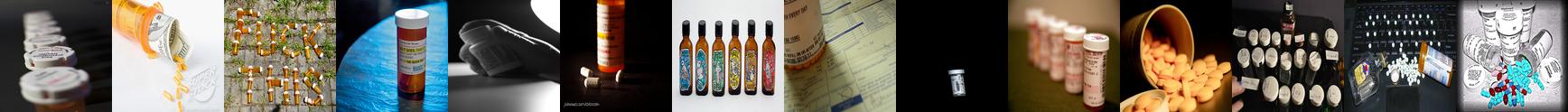} \\
\hline
plunger & \includegraphics[width=12cm]{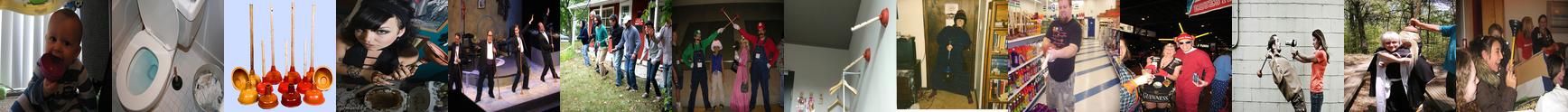} \\
\hline
pole & \includegraphics[width=12cm]{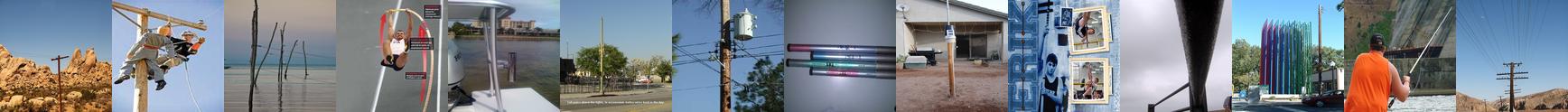} \\
\hline
police van & \includegraphics[width=12cm]{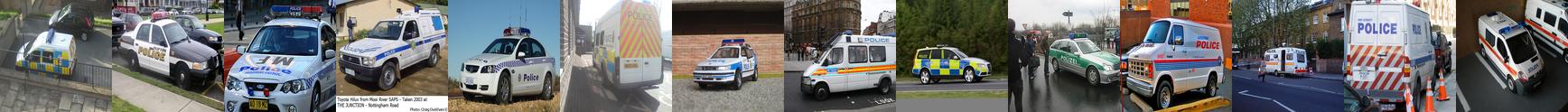} \\
\hline
poncho & \includegraphics[width=12cm]{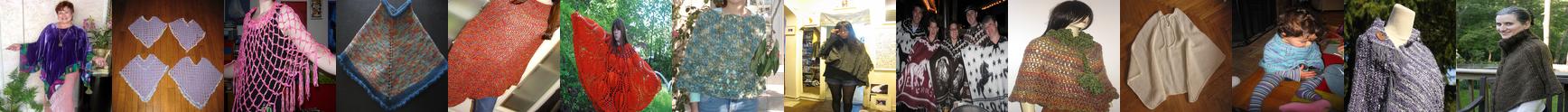} \\
\hline
pop bottle & \includegraphics[width=12cm]{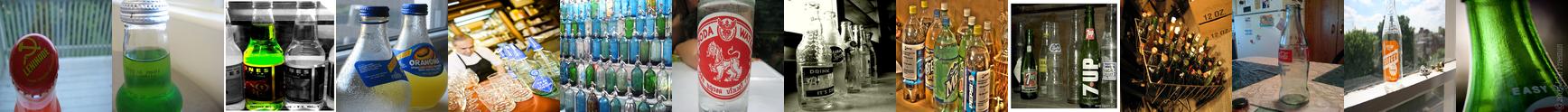} \\
\hline
potter's wheel & \includegraphics[width=12cm]{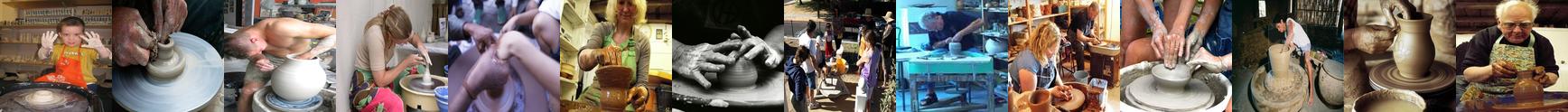} \\
\hline
projectile & \includegraphics[width=12cm]{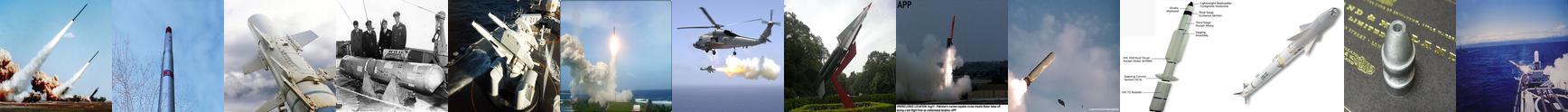} \\
\hline
punching bag & \includegraphics[width=12cm]{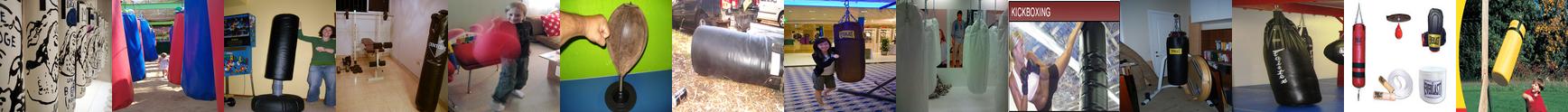} \\
\hline
reel & \includegraphics[width=12cm]{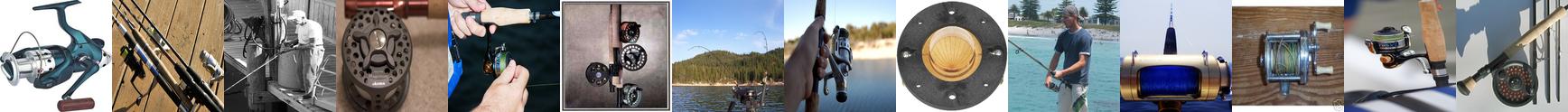} \\
\hline
refrigerator & \includegraphics[width=12cm]{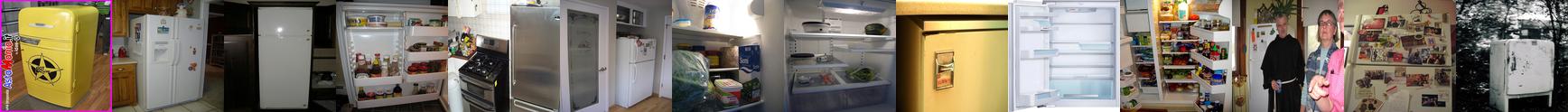} \\
\hline
remote control & \includegraphics[width=12cm]{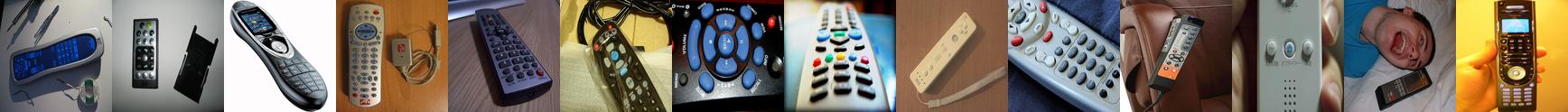} \\
\hline
rocking chair & \includegraphics[width=12cm]{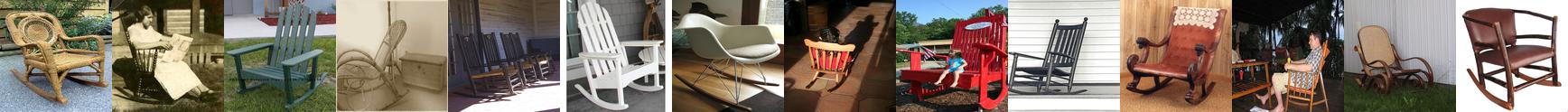} \\
\hline
rugby ball & \includegraphics[width=12cm]{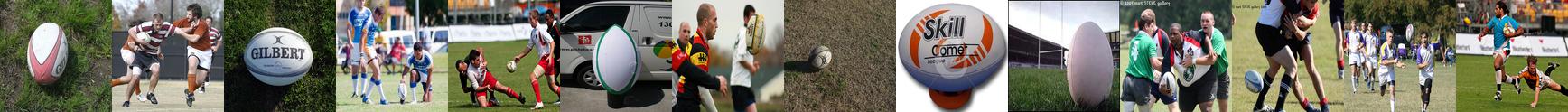} \\
\hline
sandal & \includegraphics[width=12cm]{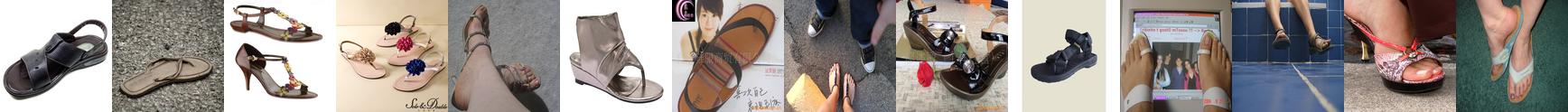} \\
\hline
school bus & \includegraphics[width=12cm]{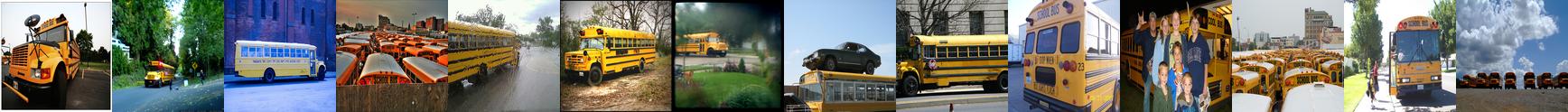} \\
\hline
scoreboard & \includegraphics[width=12cm]{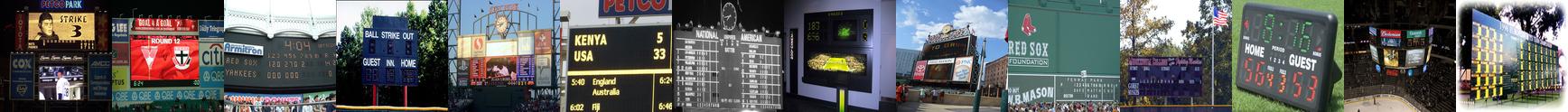} \\
\hline
sewing machine & \includegraphics[width=12cm]{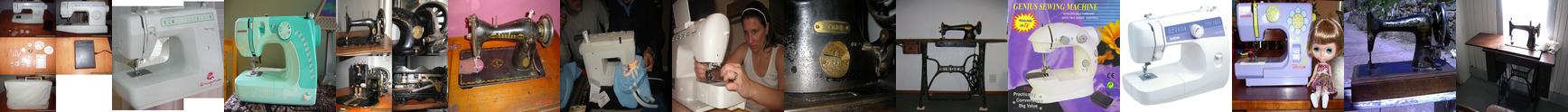} \\
\hline
snorkel & \includegraphics[width=12cm]{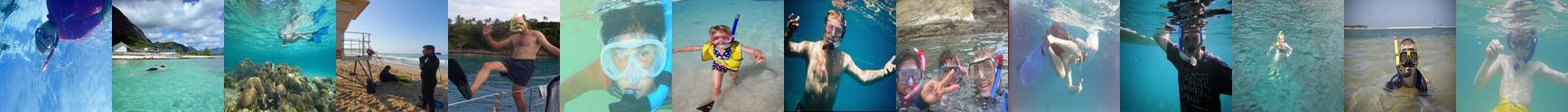} \\
\hline
sock & \includegraphics[width=12cm]{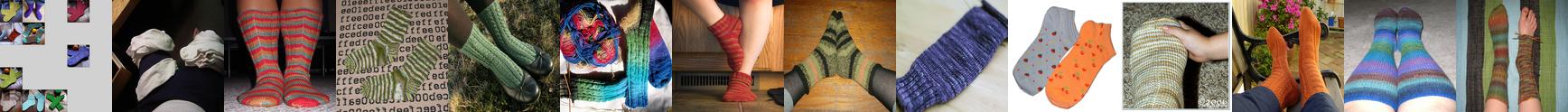} \\
\hline
sombrero & \includegraphics[width=12cm]{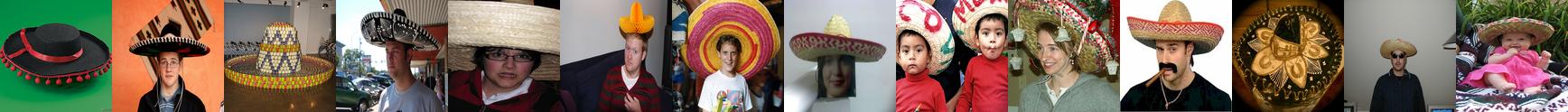} \\
\hline
space heater & \includegraphics[width=12cm]{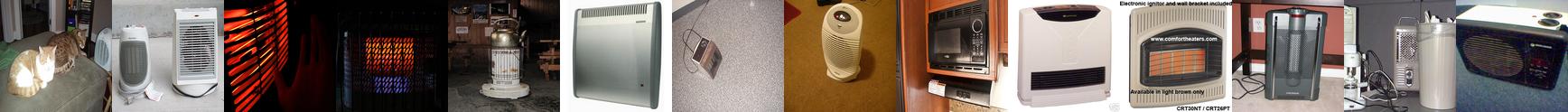} \\
\hline
spider web & \includegraphics[width=12cm]{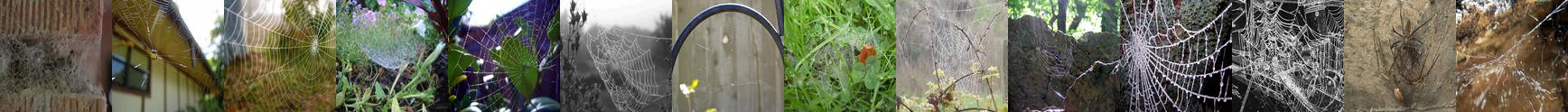} \\
\hline
sports car & \includegraphics[width=12cm]{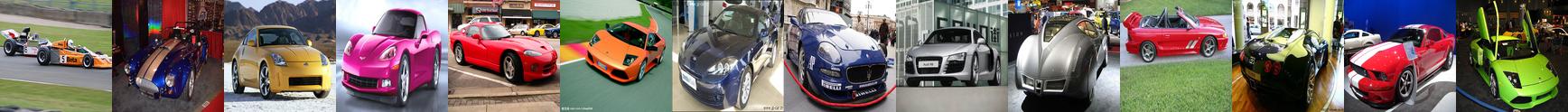} \\
\hline
steel arch bridge & \includegraphics[width=12cm]{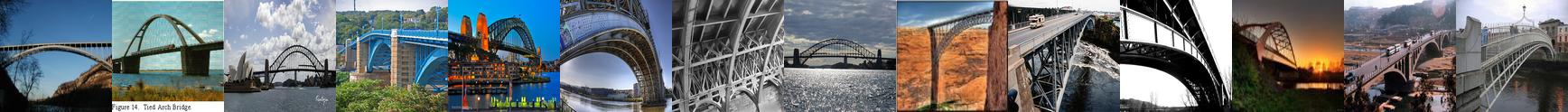} \\
\hline
stopwatch & \includegraphics[width=12cm]{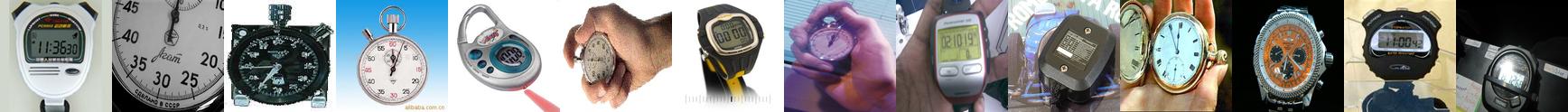} \\
\hline
sunglasses & \includegraphics[width=12cm]{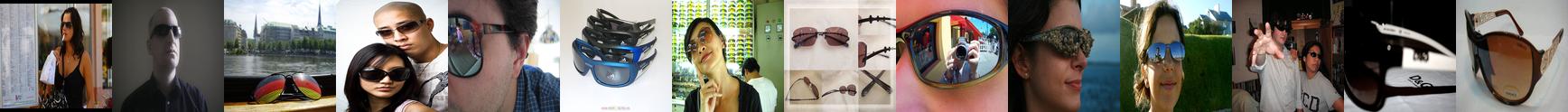} \\
\hline
suspension bridge & \includegraphics[width=12cm]{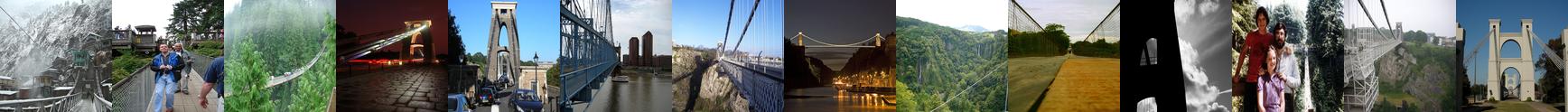} \\
\hline
swimming trunks & \includegraphics[width=12cm]{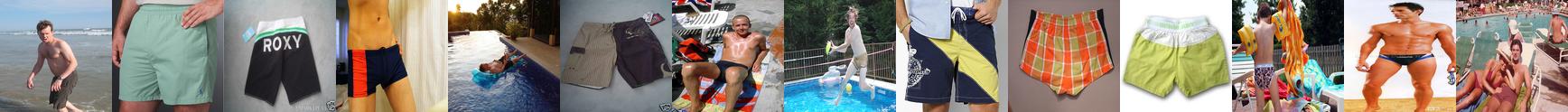} \\
\hline
syringe & \includegraphics[width=12cm]{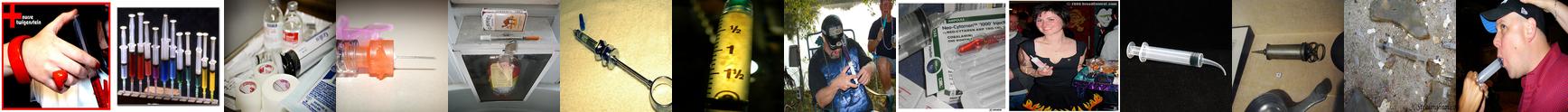} \\
\hline
teapot & \includegraphics[width=12cm]{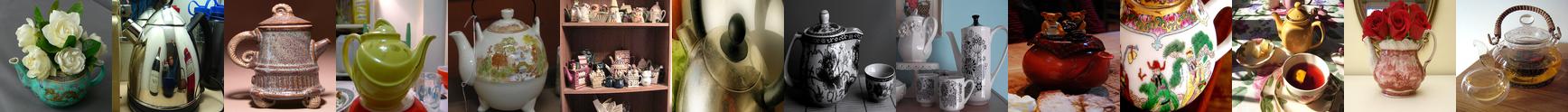} \\
\hline
teddy bear & \includegraphics[width=12cm]{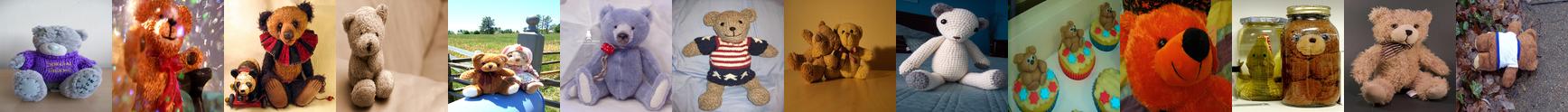} \\
\hline
thatched roof & \includegraphics[width=12cm]{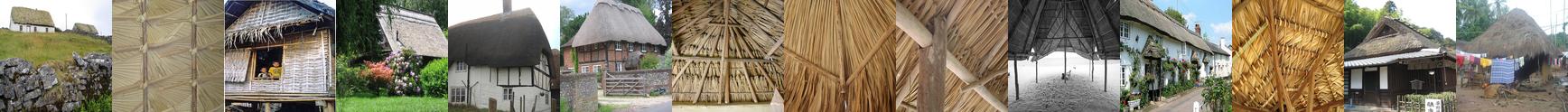} \\
\hline
torch & \includegraphics[width=12cm]{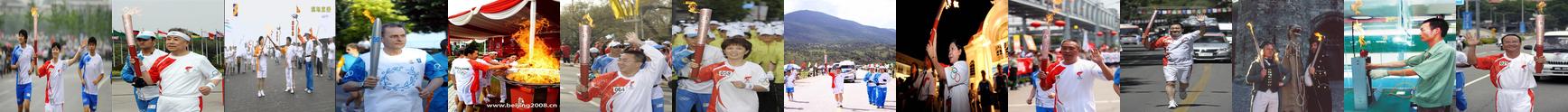} \\
\hline
tractor & \includegraphics[width=12cm]{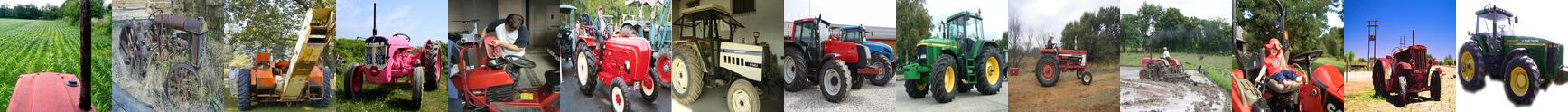} \\
\hline
triumphal arch & \includegraphics[width=12cm]{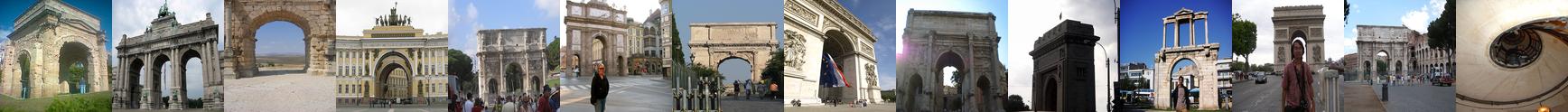} \\
\hline
trolleybus & \includegraphics[width=12cm]{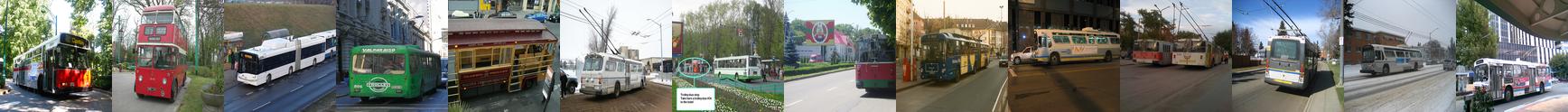} \\
\hline
turnstile & \includegraphics[width=12cm]{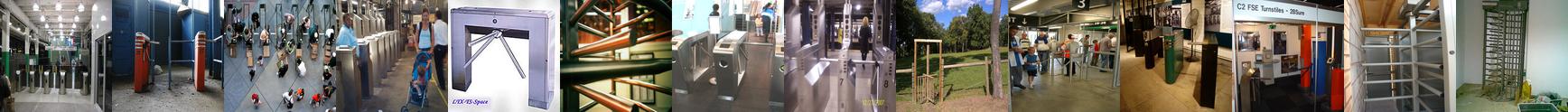} \\
\hline
umbrella & \includegraphics[width=12cm]{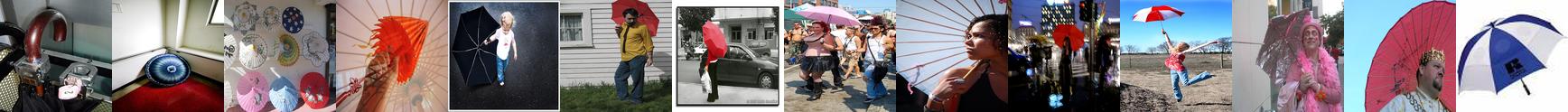} \\
\hline
vestment & \includegraphics[width=12cm]{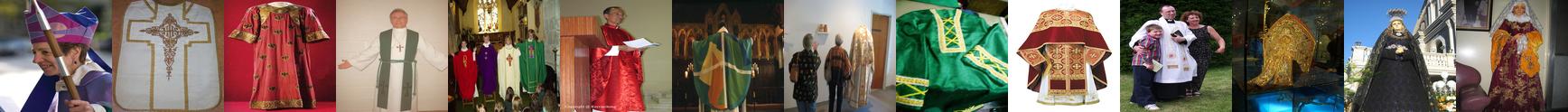} \\
\hline
viaduct & \includegraphics[width=12cm]{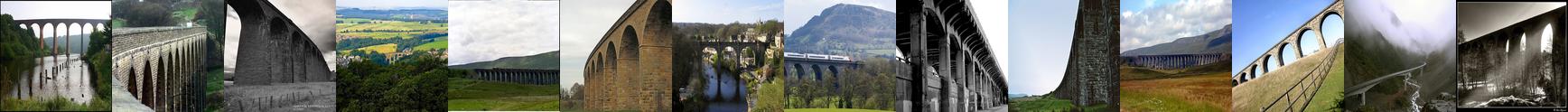} \\
\hline
volleyball & \includegraphics[width=12cm]{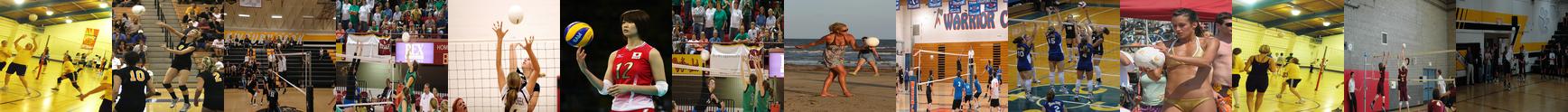} \\
\hline
water jug & \includegraphics[width=12cm]{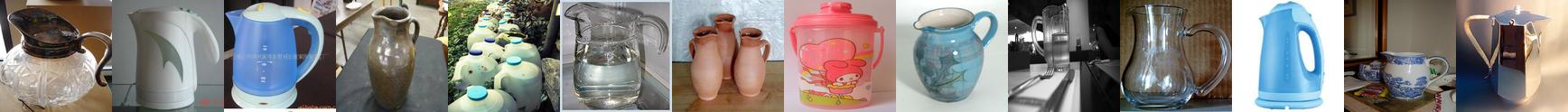} \\
\hline
water tower & \includegraphics[width=12cm]{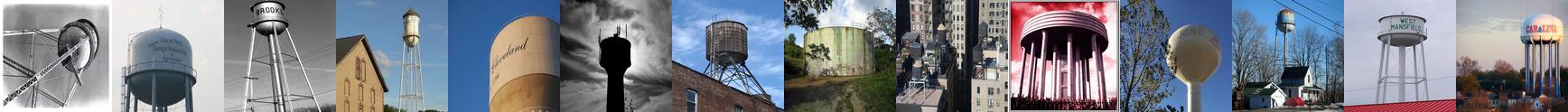} \\
\hline
wok & \includegraphics[width=12cm]{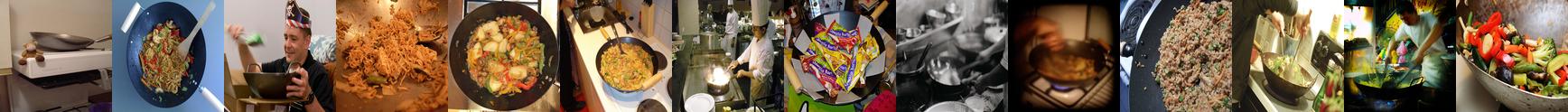} \\
\hline
wooden spoon & \includegraphics[width=12cm]{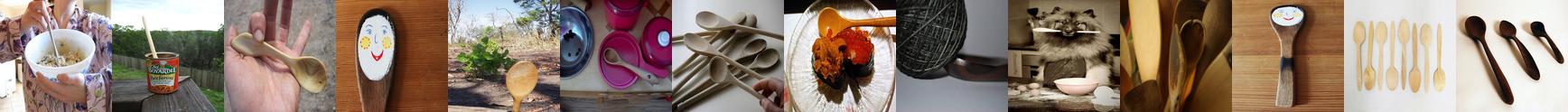} \\
\hline
comic book & \includegraphics[width=12cm]{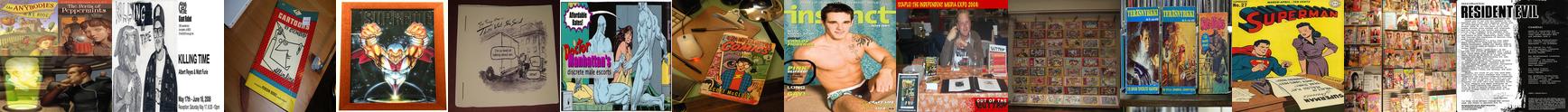} \\
\hline
plate & \includegraphics[width=12cm]{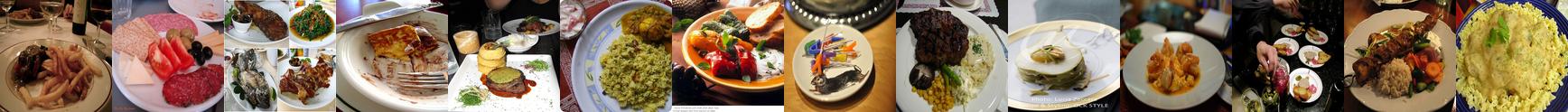} \\
\hline
guacamole & \includegraphics[width=12cm]{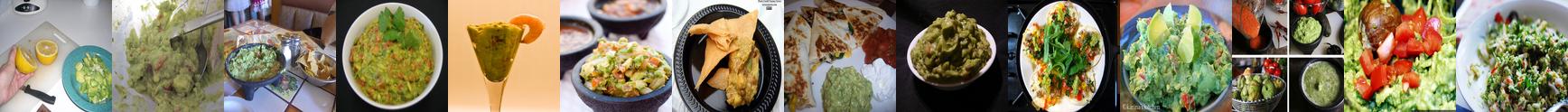} \\
\hline
ice cream & \includegraphics[width=12cm]{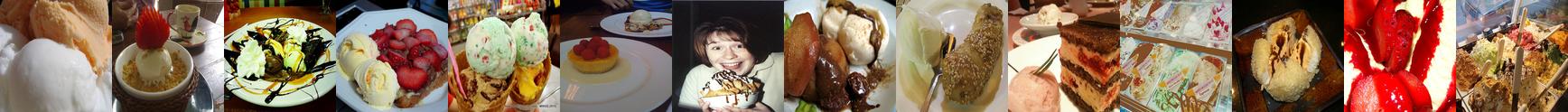} \\
\hline
lollipop & \includegraphics[width=12cm]{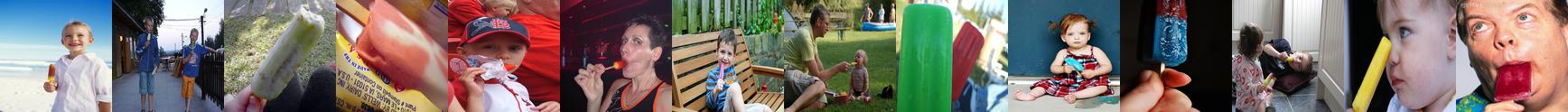} \\
\hline
pretzel & \includegraphics[width=12cm]{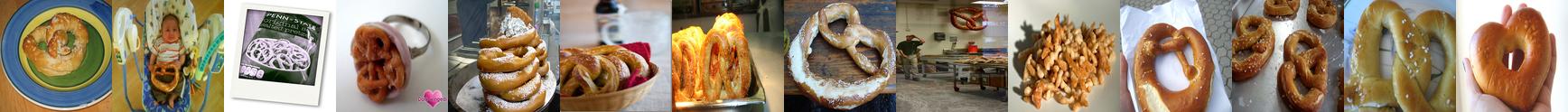} \\
\hline
mashed potato & \includegraphics[width=12cm]{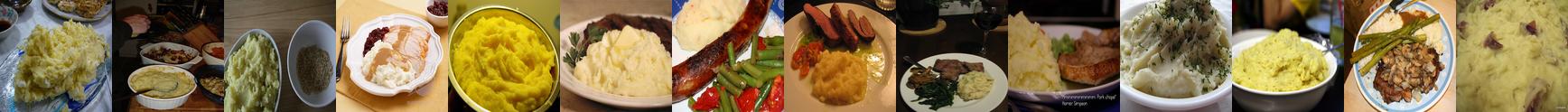} \\
\hline
cauliflower & \includegraphics[width=12cm]{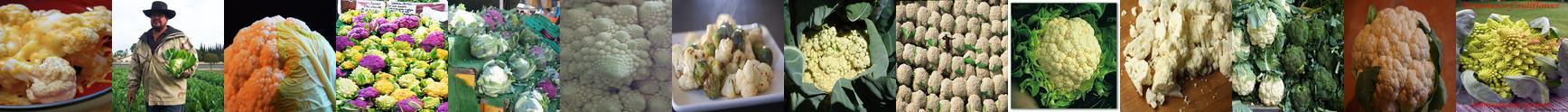} \\
\hline
bell pepper & \includegraphics[width=12cm]{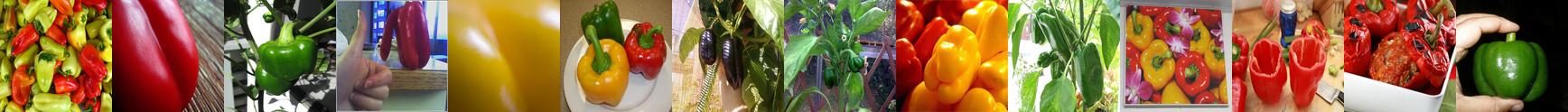} \\
\hline
mushroom & \includegraphics[width=12cm]{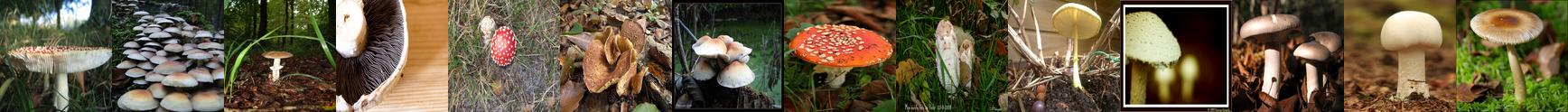} \\
\hline
orange & \includegraphics[width=12cm]{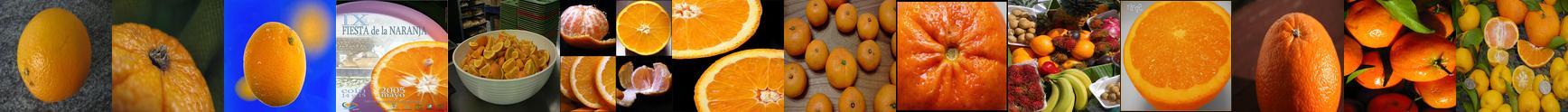} \\
\hline
lemon & \includegraphics[width=12cm]{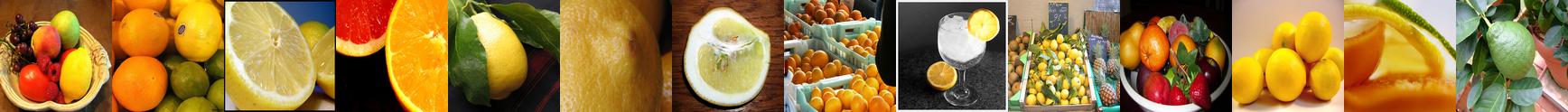} \\
\hline
banana & \includegraphics[width=12cm]{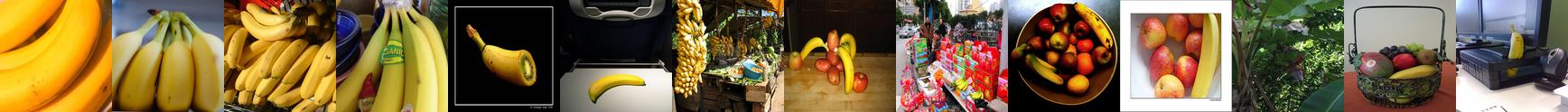} \\
\hline
pomegranate & \includegraphics[width=12cm]{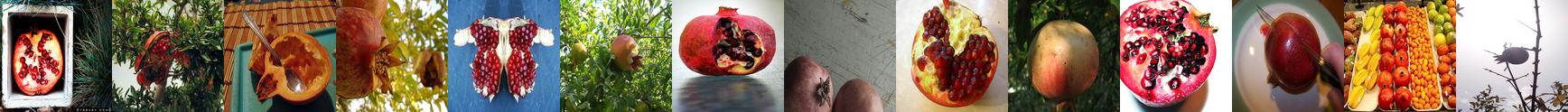} \\
\hline
meat loaf & \includegraphics[width=12cm]{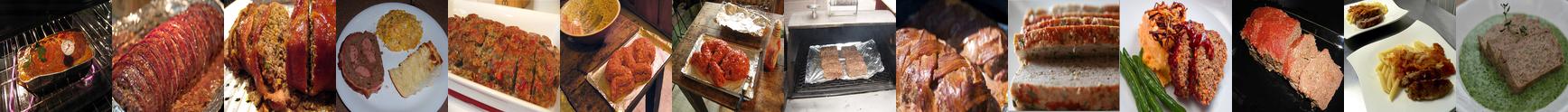} \\
\hline
pizza & \includegraphics[width=12cm]{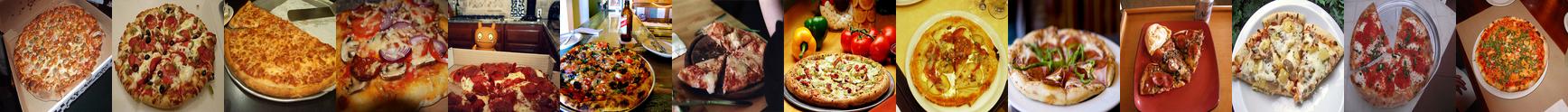} \\
\hline
potpie & \includegraphics[width=12cm]{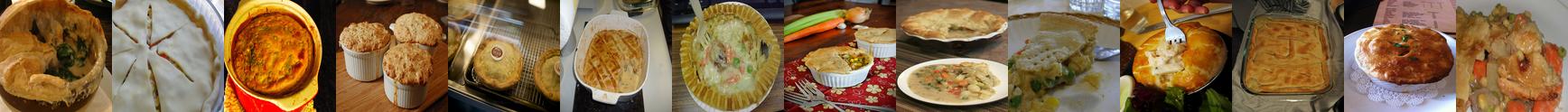} \\
\hline
espresso & \includegraphics[width=12cm]{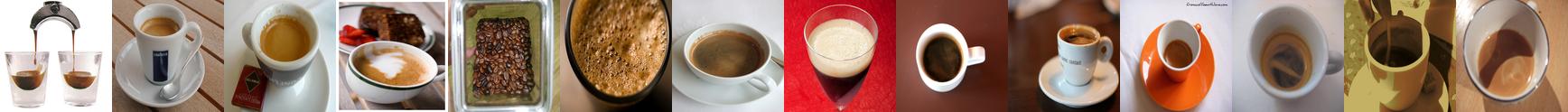} \\
\hline
alp & \includegraphics[width=12cm]{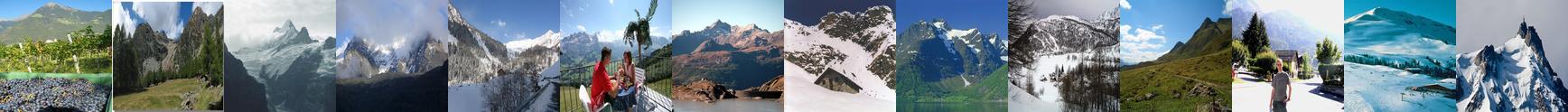} \\
\hline
cliff & \includegraphics[width=12cm]{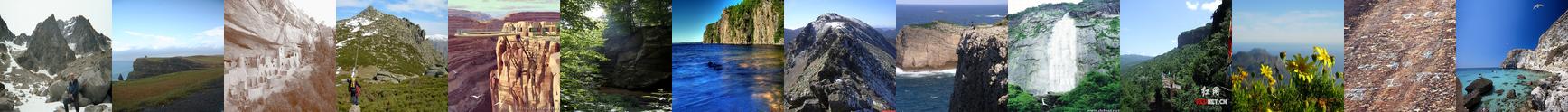} \\
\hline
coral reef & \includegraphics[width=12cm]{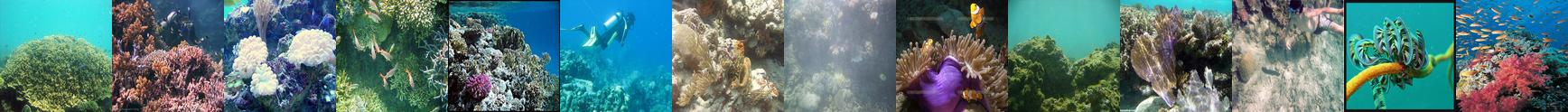} \\
\hline
lakeside & \includegraphics[width=12cm]{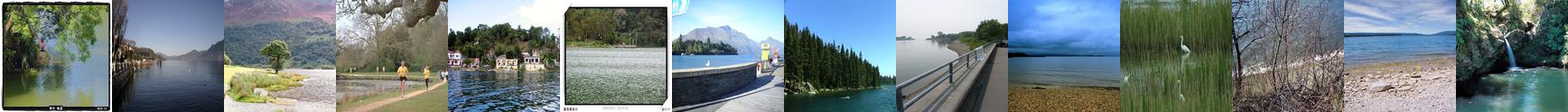} \\
\hline
seashore & \includegraphics[width=12cm]{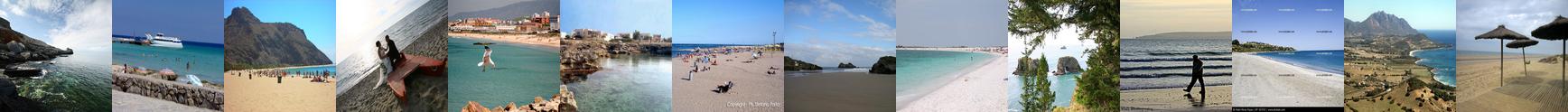} \\
\hline
acorn & \includegraphics[width=12cm]{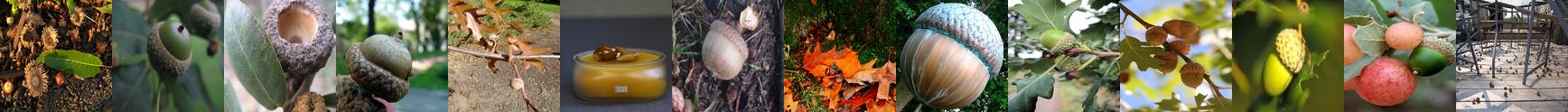} \\
\hline
\end{longtable}

\end{document}